\documentclass{article}

\usepackage{graphicx}
\usepackage{booktabs} %
\usepackage{scalerel}
\usepackage{float}

\usepackage{subcaption}
\usepackage{caption}

\usepackage[toc,page]{appendix}

\usepackage{url}            %
\usepackage{amsfonts}       %
\usepackage{nicefrac}       %
\usepackage{microtype}      %
\usepackage{cancel}

\usepackage[dvipsnames]{xcolor}
\usepackage{amsmath, latexsym}
\usepackage[most]{tcolorbox}
\usepackage{placeins}
\usepackage{enumitem}
\usepackage{bold-extra}
\usepackage[T1]{fontenc}
\usepackage{standalone}
\usepackage{tikz}
\usepackage{tikzscale}
\usepackage{pgfplots}
\usepackage{pgf}
\usetikzlibrary{calc}
\usetikzlibrary{positioning}
\usetikzlibrary{angles,quotes}
\usetikzlibrary{backgrounds}
\usetikzlibrary{external}
\usetikzlibrary{fit}
\usetikzlibrary{arrows}
\usetikzlibrary{arrows.meta}
\usetikzlibrary{shapes.symbols}
\usetikzlibrary{shadings}
\usetikzlibrary{shapes}
\usetikzlibrary{fadings}
\usetikzlibrary{bayesnet}
\usetikzlibrary{matrix}
\usetikzlibrary{decorations.pathmorphing, patterns}
\pgfplotsset{compat=1.15}

\definecolor{C1}{HTML}{1F77B4}
\definecolor{C2}{HTML}{FF7F0E}
\definecolor{C3}{HTML}{2CA02C}
\definecolor{C4}{HTML}{D62728}
\definecolor{C5}{HTML}{9467BD}
\colorlet{C1light}{C1!70!white}
\colorlet{C2light}{C2!70!white}
\colorlet{C3light}{C3!70!white}
\colorlet{C4light}{C4!70!white}
\colorlet{C5light}{C5!70!white}
\colorlet{C1vlight}{C1!20!white}
\colorlet{C2vlight}{C2!20!white}
\colorlet{C3vlight}{C3!20!white}
\colorlet{C4vlight}{C4!20!white}
\colorlet{C5vlight}{C5!20!white}
\colorlet{linkcolor}{violet}
\colorlet{citecolor}{RedOrange}  %
\colorlet{urlcolor}{Aquamarine}
\definecolor{ultrapink}{rgb}{1.0, 0.44, 1.0}
\RequirePackage{contour}
\RequirePackage{xspace}

\tikzset{
    /pgf/decoration/amplitude = 0.1em,
    /pgf/decoration/segment length = 0.5em}
\usepackage{cuted}
\usepackage{flushend}

\newsavebox{\varmatrixbox}

\ExplSyntaxOn
\keys_define:nn {martin/varmatrix}
 {
  sep   .dim_set:N = \l_martin_varmatrix_sep_dim,
  delim .tl_set:N  = \l_martin_varmatrix_delim_tl,
  size  .tl_set:N  = \l_martin_varmatrix_size_tl,
  sep   .initial:n = 0.7\arraycolsep,
  size  .initial:n = \small,
 }

\NewDocumentEnvironment{varmatrix}{O{}}
 {
  \keys_set:nn {martin/varmatrix} { #1 }
  \begin{lrbox}{\varmatrixbox}
  \l_martin_varmatrix_size_tl
  \setlength{\arraycolsep}{\l_martin_varmatrix_sep_dim}
  $\begin{\l_martin_varmatrix_delim_tl matrix}
 }
 {
  \end{\l_martin_varmatrix_delim_tl matrix}$
  \end{lrbox}
  \vcenter{\box\varmatrixbox}
 }
\ExplSyntaxOff

\usepackage{hyperref}
\usepackage[customcolors,shade]{hf-tikz}

\newsavebox\qrrTikzmarkBox
\RenewDocumentCommand{\tikzmarkin}{O{} m D(){0.1,-0.15} D(){-0.1,0.18} u{\tikzmarkend}}{%
\ifx\\#5\\
    \dp\qrrTikzmarkBox=0pt\relax
    \ht\qrrTikzmarkBox=0pt\relax
\else
    \sbox\qrrTikzmarkBox{$\displaystyle#5$}
\fi
\tikz[remember picture,overlay]
\draw[line width=1pt,rectangle,fill=\fcol,#1,draw=\bcol]
(pic cs:#2) ++([yshift=-\the\dp\qrrTikzmarkBox]#3) rectangle ([yshift=\the\ht\qrrTikzmarkBox]#4) node [anchor=text] (#2) {}
;
#5
\tikzmarkend
}

\usepackage{nccmath}
\usepackage[accepted]{icml2023}
\usepackage{multirow}
\usepackage{stfloats} 
\usepackage{xargs} 
\usepackage{float}
\usepackage{mdframed}
\newmdenv[
  topline=false,
  bottomline=false,
  rightline=false,
  skipabove=\topsep,
  ]{siderules}

\newcommand{\sgparam}{\tau}
\newcommand{\slparam}{\eta}

\newcommand{\gparam}{\boldsymbol{\tau}}
\newcommand{\lparam}{\boldsymbol{\eta}}

\newcommand{\transport}{T}
\newcommand{\RExp}{\mathrm{RExp}}
\newcommand{\MExp}{\mathrm{Expm}}
\newcommand{\LieBracket}[2]{\left<{#1}, {#2}\right>}

\newcommand{\stepsize}{\beta}

\newcommand{\vlat}{\vz}
\newcommand{\lat}{z}

\newcommand\cut[1]{}

\newcommand{\squishlist}{
   \begin{list}{$\bullet$}
    { \setlength{\itemsep}{0pt}      \setlength{\parsep}{3pt}
      \setlength{\topsep}{3pt}       \setlength{\partopsep}{0pt}
      \setlength{\leftmargin}{1.5em} \setlength{\labelwidth}{1em}
      \setlength{\labelsep}{0.5em} } }

\newcommand{\squishlisttwo}{
   \begin{list}{$\bullet$}
    { \setlength{\itemsep}{0pt}    \setlength{\parsep}{0pt}
      \setlength{\topsep}{0pt}     \setlength{\partopsep}{0pt}
      \setlength{\leftmargin}{2em} \setlength{\labelwidth}{1.5em}
      \setlength{\labelsep}{0.5em} } }

\newcommand{\squishend}{
    \end{list}  }

{}
{}
{}
{}
{}

\newcommand{\half}{\mbox{$\frac{1}{2}$}}

\newcommand{\real}{\mbox{$\mathbb{R}$}}

\newcommand{\Unmyexpect}[1]{\mathbb{E}_{\scaleto{#1\mathstrut}{6pt}}}

\newcommand{\gauss}{\mbox{${\cal N}$}}

\newcommand{\myvec}[1]{\mbox{$\mathbf{#1}$}}
\newcommand{\myvecsym}[1]{\mbox{$\boldsymbol{#1}$}}

\newcommand{\vmu}{\mbox{$\myvecsym{\mu}$}}
\newcommand{\vlambda}{\mbox{$\myvecsym{\lambda}$}}

\newcommand{\vphi}{\mbox{$\myvecsym{\phi}$}}

\newcommand{\vpsi}{\myvecsym{\psi}}
\newcommand{\vnu}{\mbox{$\myvecsym{\nu}$}}

\newcommand{\vSigma}{\mbox{$\myvecsym{\Sigma}$}}

\newcommand{\vomega}{\mbox{$\myvecsym{\omega}$}}

\newcommand{\vxi}{\mbox{$\myvecsym{\xi}$}}

\newcommand{\vb}{\mbox{$\myvec{b}$}}
\newcommand{\vc}{\mbox{$\myvec{c}$}}

\newcommand{\ve}{\mbox{$\myvec{e}$}}

\newcommand{\vf}{\mbox{$\myvec{f}$}}
\newcommand{\vg}{\mbox{$\myvec{g}$}}

\newcommand{\vm}{\mbox{$\myvec{m}$}}

\newcommand{\vr}{\mbox{$\myvec{r}$}}

\newcommand{\vv}{\mbox{$\myvec{v}$}}
\newcommand{\vw}{\mbox{$\myvec{w}$}}

\newcommand{\vx}{\mbox{$\myvec{x}$}}

\newcommand{\vy}{\mbox{$\myvec{y}$}}

\newcommand{\vz}{\mbox{$\myvec{z}$}}
\newcommand{\vA}{\mbox{$\myvec{A}$}}
\newcommand{\vB}{\mbox{$\myvec{B}$}}
\newcommand{\vC}{\mbox{$\myvec{C}$}}
\newcommand{\vD}{\mbox{$\myvec{D}$}}
\newcommand{\vE}{\mbox{$\myvec{E}$}}
\newcommand{\vF}{\mbox{$\myvec{F}$}}
\newcommand{\vG}{\mbox{$\myvec{G}$}}
\newcommand{\vH}{\mbox{$\myvec{H}$}}
\newcommand{\vI}{\mbox{$\myvec{I}$}}
\newcommand{\vJ}{\mbox{$\myvec{J}$}}
\newcommand{\vK}{\mbox{$\myvec{K}$}}
\newcommand{\vL}{\mbox{$\myvec{L}$}}
\newcommand{\vM}{\mbox{$\myvec{M}$}}
\newcommand{\vN}{\mbox{$\myvec{N}$}}

\newcommand{\vS}{\mbox{$\myvec{S}$}}
\newcommand{\vT}{\mbox{$\myvec{T}$}}
\newcommand{\vU}{\mbox{$\myvec{U}$}}
\newcommand{\vV}{\mbox{$\myvec{V}$}}
\newcommand{\vW}{\mbox{$\myvec{W}$}}
\newcommand{\vX}{\mbox{$\myvec{X}$}}

\newcommand{\be}{\begin{equation}}
\newcommand{\ee}{\end{equation}}
\newcommand{\bea}{\begin{eqnarray}}
\newcommand{\eea}{\end{eqnarray}}
\newcommand{\beaa}{\begin{eqnarray*}}
\newcommand{\eeaa}{\end{eqnarray*}}

\usepackage{comment}

\begin{document}

\newcommand{\ourtitle}{
  Simplifying Momentum-based Positive-definite Submanifold Optimization\\ with Applications to Deep Learning   \vspace{-0.2cm}
}
\icmltitlerunning{Simplifying Momentum-based Positive-definite Submanifold Optimization with Applications to DL}

\twocolumn[

\icmltitle{ \ourtitle }

\begin{icmlauthorlist}
\icmlauthor{Wu Lin}{ubc}
\icmlauthor{Valentin Duruisseaux}{ucsd}
\icmlauthor{Melvin Leok}{ucsd}
\icmlauthor{Frank Nielsen}{sony}
\icmlauthor{Mohammad Emtiyaz Khan}{riken}
\icmlauthor{Mark Schmidt}{ubc,amii}
\end{icmlauthorlist}

\icmlaffiliation{ubc}{University of British Columbia, Vancouver, Canada}
\icmlaffiliation{riken}{RIKEN Center for Advanced Intelligence Project, Tokyo, Japan}
\icmlaffiliation{sony}{Sony Computer Science Laboratories Inc., Tokyo, Japan}
\icmlaffiliation{amii}{CIFAR AI Chair, Alberta Machine Intelligence Institute, Alberta, Canada}
\icmlaffiliation{ucsd}{University of California San Diego, San Diego,   USA}

\icmlcorrespondingauthor{Wu Lin}{yorker.lin@gmail.com \vspace{-0.2cm}}

\icmlkeywords{Natural Gradient Descent, Information Geometry, Variational Inference, Optimization, Search, Deep Learning}

\vskip 0.3in
]

\printAffiliationsAndNotice{}  %

\begin{abstract}
Riemannian submanifold optimization with momentum is computationally challenging because, to ensure that the iterates remain on the submanifold, we often need to solve difficult differential equations. 
Here, we simplify such difficulties for a class of sparse or structured symmetric positive-definite matrices with the affine-invariant metric.
We do so by proposing a generalized version of the Riemannian normal coordinates that dynamically orthonormalizes the metric and locally converts the problem into an unconstrained problem in the Euclidean space.
We use our approach to simplify existing approaches for structured covariances and develop matrix-inverse-free $2^\text{nd}$-order optimizers for deep learning with low precision by using only matrix multiplications.
\vspace{-0.2cm}

\end{abstract}

\vspace{-0.6cm}
\section{Introduction}
\vspace{-0.1cm}
\vspace{-0.1cm}
Estimation of symmetric positive definite (SPD) matrices is important in machine learning (ML) and related fields. For example, many optimization methods require it to estimate preconditioning matrices. Approximate inference methods also estimate SPD matrices to obtain Gaussian posterior approximations. Other applications include dictionary learning~\citep{cherian2016riemannian}, trace regression~\citep{slawski2015regularization} for kernel matrices, metric learning~\citep{guillaumin2009you}, log-det maximization~\citep{wang2010solving}, Gaussian mixtures~\citep{hosseini2015matrix},
and Gaussian graphical models~\citep{makam2021symmetries}.

Because the set of SPD matrices forms a Riemannian manifold, one can use Riemannian gradient methods for SPD estimation, but this can be computationally infeasible in high-dimensions. This is because the methods often require full-rank matrix decomposition (see Table~\ref{tab:spd_table}).
Computations can be reduced by using sparse matrices induced by a submanifold. However, this complicates manifold operations needed for such submanifold optimization. For example, the Riemannian gradient computation often involves metric  inversion (see Table~\ref{tab:spd_table}).  
Other operations needed for Riemannian momentum, such as the Riemannian exponential map and the parallel transport map, also require solving an intractable ordinary differential equation (ODE).

Another idea to develop practical Riemannian methods is to use moving coordinates where a local coordinate is generated, used, and discarded at each iteration. 
Such approaches can efficiently handle manifold constraints (see for example the proposed structured natural-gradient descent (NGD) method by \citet{lin2021tractable}).
However, it is nontrivial to include metric-aware momentum using moving coordinates in a computationally efficient way.
In this paper, we aim to simplify the addition of momentum to such methods and develop efficient momentum-based updates on submanifolds.
We propose special local coordinates for a class of SPD submanifolds (Eq.~\eqref{eq:submanifolds}) with the affine-invariant metric.~Our approach avoids the use of global coordinates as well as the computation of Riemannian exponential and transport maps.~Instead, we exploit Lie-algebra structures in the local coordinates to obtain 
efficient structure-preserving updates on submanifolds induced by Lie subgroups.
Under our local coordinates, all constraints disappear and the metric at evaluation points becomes the standard Euclidean metric. This \emph{metric-preserving}  
trivialization on a submanifold enables an efficient metric-inverse-free Riemannian momentum update by essentially performing, in the local coordinates, a momentum-based Euclidean gradient descent (GD) update.

We extend the structured NGD approach in several ways: (i) we establish its connection to Riemannian methods (Sec.~\ref{sec:momentum}); (ii) we demystify the construction of coordinates (Sec.~\ref{sec:demystifying}); (iii) we introduce new coordinates for efficient momentum computation (Sec.~\ref{sec:momentum}); and (iv) we expand its scope to structured SPD matrices where Gaussian and Bayesian assumptions are not needed (Sec.~\ref{sec:expanding}).
By exploiting the submanifold structure of preconditioners, we use our method to develop new inverse-free structured optimizers 
for deep learning (DL) in low precision settings (Sec.~\ref{sec:opt_dl}).

\section{Manifold Optimization and its Challenges}
\vspace{-0.2cm}
Consider a complete manifold  $\mathcal{M}$ with a Riemannian metric $\vF$ represented by a single global coordinate $\gparam$.
Under this coordinate system, Riemannian gradient descent~\citep{absil2009optimization,bonnabel2013stochastic} (RGD) is defined as

 \vspace{-0.3cm}
\resizebox{0.9\linewidth}{!}{
  \begin{minipage}[t]{\linewidth}
\begin{align}
\text{RGD}:\,\,&
  \gparam \leftarrow \RExp(\gparam, -\stepsize \vF^{-1}(\gparam) \vg(\gparam) ) ,
\label{eq:rgd}
\end{align} \end{minipage} 
}
\vspace{-0.12cm}

where $\vv(\gparam):=\vF^{-1}(\gparam) \vg(\gparam)$ is a Riemannian gradient known as a type $(1,0)$-tensor,
$\vg(\gparam)$ is a Euclidean gradient known as a type $(0,1)$-tensor, $\vF(\gparam)$ is the metric known as a type $(0,2)$-tensor, $\stepsize$ is a stepsize, and
$\RExp(\gparam,\vv)$ is the Riemannian exponential map defined by solving a nonlinear (geodesic) ODE (see Appx.~\ref{apd:expmap}). %
The nonlinearity of the ODE makes it difficult to obtain a closed-form expression for the solution.

To incorporate momentum in RGD, a Riemannian parallel transport map $\hat{\transport}_{\sgparam^{\text{(cur)}}\rightarrow \sgparam^{\text{(new)}}}(\vnu^{\text{(cur)}})$ is introduced in many works to move the Riemannian vector (known as a type $(1,0)$-tensor)  $\vnu^{\text{(cur)}}$  computed at point $\gparam^{\text{(cur)}}$ to point $\gparam^{\text{(new)}}$ on the manifold. For example, \citet{alimisis2020continuous} propose the following update using the Riemannian transport map  (see Appx.~\ref{apd:trans_rg}) and Riemannian momentum $\vnu^{\text{(cur)}}$    with momentum weight $\alpha$: 

\vspace{-0.31cm}
\hspace{3mm} \resizebox{0.96\linewidth}{!}{ 
  \begin{minipage}[t]{\linewidth} 
\begin{align}   
\text{Momentum}:\,\,&
\vnu^{\text{(cur)}} \leftarrow \alpha \vz^{\text{(cur)}} + \stepsize  \vF^{-1}(\gparam^{\text{(cur)}})  \vg(\gparam^{\text{(cur)}}) ,\nonumber \\
\text{RGD}:\,\,& \gparam^{\text{(new)}} \leftarrow \RExp (\gparam^{\text{(cur)}}, -\vnu^{\text{(cur)}} ), \nonumber \\
\text{Transport}:\,\,& \vz^{\text{(new)}} \leftarrow \hat{\transport}_{\sgparam^{\text{(cur)}}\rightarrow \sgparam^{\text{(new)}}}(\vnu^{\text{(cur)}}) .
\label{eq:rgd_mom}
\end{align}\end{minipage}
}

Unlike existing works, we suggest using Euclidean momentum and a Euclidean parallel transport map  (see Appx.~\ref{apd:trans_eg}) $\transport_{\sgparam^{\text{(cur)}}\rightarrow \sgparam^{\text{(new)}}}(\vm^{\text{(cur)}})$ to move the momentum (known as a type $(0,1)$-tensor).
The use of this Euclidean map is essential for efficient approximations of the transport,
as will be discussed in Sec.~\ref{sec:momentum}. 
Through this Euclidean map, we obtain an equivalent update of 
Eq.~\eqref{eq:rgd_mom}
(shown in Appx.~\ref{apd:equ_alimi}) via Euclidean momentum $\vm^{\text{(cur)}}$:

\vspace{-0.31cm}
\hspace{3mm} \resizebox{0.96\linewidth}{!}{
  \begin{minipage}[t]{\linewidth}
\begin{align}  
\text{Momentum}:\,\,&
\vm^{\text{(cur)}} \leftarrow \alpha \vw^{\text{(cur)}} +   \stepsize \vg(\gparam^{\text{(cur)}}), \nonumber \\
\text{RGD}:\,\,&   \gparam^{\text{(new)}}    \leftarrow \RExp(\gparam^{\text{(cur)}}, -\vF^{-1}(\gparam^{\text{(cur)}})\vm^{\text{(cur)}} ) , \nonumber \\
\text{Transport}:\,\,& \vw^{\text{(new)}} \leftarrow  {\transport}_{\sgparam^{\text{(cur)}}\rightarrow \sgparam^{\text{(new)}}}( \vm^{\text{(cur)}}). %
\label{eq:rgd_mom_eg}
\end{align} \end{minipage} 
}

Given a Euclidean gradient $\vg$ evaluated at $\gparam^{\text{(cur)}},$ the Riemannian transport map  and the Euclidean transport map  are related as follows,
\resizebox{0.95\linewidth}{!}{
  \begin{minipage}[t]{1.1\linewidth}
\begin{align}
\hat{\transport}_{\sgparam^{\text{(cur)}} \rightarrow \sgparam^{\text{(new)}}}(\vF^{-1}(\gparam^{\text{(cur)}})\vg) =  \vF^{-1}({\color{red} \gparam^{\text{(new)}}}) {\transport}_{\sgparam^{\text{(cur)}}\rightarrow\sgparam^{\text{(new)}}}(\vg).
     \label{eq:equ_trans_maps}
\end{align} \end{minipage} 
} 
We can use this relationship to show the equivalence of Eq.~\ref{eq:rgd_mom} and Eq.~\ref{eq:rgd_mom_eg}.

Both the transport maps are defined by solving transport ODEs (defined in Appx.~\ref{apd:trans_rg}-\ref{apd:trans_eg}).
However, solving any of the ODEs can be computationally intensive since it is a linear system of differential equations and the solution often involves matrix decomposition. 

\vspace{-0.2cm}
\subsection{Challenges on SPD Manifold and Submanifolds}
\vspace{-0.2cm}

Consider the $k$-by-$k$ SPD manifold  $\mathcal{M}=\{ \gparam \in \real^{k \times k} |\gparam \succ 0 \}$. The affine-invariant metric $\vF$ for the SPD manifold (see Theorem 2.10 of \citet{minh2017covariances}) is defined as twice the Fisher-Rao metric (see Appx.~\ref{apd:fim}) for the $k$-dimensional Gaussian distribution $\gauss(\mathbf{0}, \gparam)$ with zero mean and covariance $\gparam$, which is the 2nd derivative of a matrix,

\vspace{-0.25cm}
\resizebox{0.99\linewidth}{!}{
  \begin{minipage}[t]{\linewidth}
\begin{align}
\vF(\gparam) = -2   \Unmyexpect{\gauss(\mathbf{0}, \gparam)}\big[ \nabla_\sgparam^2 \log \gauss(\mathbf{0}, \gparam)\big].
\label{eq:affinemetric}
\end{align} \end{minipage}
}
\vspace{-0.02cm}

The Fisher-Rao metric known as the Fisher information matrix is a well-known and useful metric for many applications in ML. Moreover, the affine-invariant metric is more suitable and useful for SPD matrices compared to the standard Euclidean metric defined in either the global coordinate $\gparam$ \citep{pennec2006riemannian} or a (global) Cholesky unconstrained reparametrization \citep{hosseini2015matrix}.
Thus, we use and preserve the affine-invariant metric. Unlike existing works in manifold optimization, we preserve the metric in local coordinates instead of global coordinates.
Thus, we have to obey the metric transform rule needed for a coordinate change whenever we generate a local coordinate.

In the SPD manifold case, the Riemannian maps have a closed-form expression (see Table~\ref{tab:spd_table}). However, the updates in Eq.~\eqref{eq:rgd}-\eqref{eq:rgd_mom_eg} require computing matrix inversion/decomposition in the Riemannian maps, so such methods have $O(k^3)$ complexity and are impractical when $k$ is large.
To our best knowledge, very few Riemannian methods are developed and tested in high-dimensional (i.e., $k>10^6$), low floating-point precision, and stochastic cases.
We will propose an alternative approach to construct practical Riemannian methods for SPD submanifolds when these three cases are considered jointly.

For many SPD submanifolds\footnote{The ambient space of a SPD submanifold is a SPD manifold.
This submanifold does not have all degrees of freedom of the SPD manifold.
One trivial submanifold is the set of diagonal SPD matrices. } with the same (induced) metric, it is nontrivial to implement updates in  Eq.~\eqref{eq:rgd}-~\eqref{eq:rgd_mom_eg} since these needed Riemannian maps often do not admit a simple closed-form expression. 
For example, consider the following SPD submanifold, which can be used \citep{calvo1990distance} to represent a $(\!k\!-\!1\!)$-dimensional Gaussian with mean $\vmu$ and full covariance $\vSigma:=\vV-\vmu\vmu^T$, 

\vspace{-0.32cm}
 \resizebox{0.87\linewidth}{!}{
  \begin{minipage}[t]{\linewidth}
\begin{align*}
 \mathcal{M}=\Big\{ \gparam = \begin{bmatrix} \vV & \vmu \\ \vmu^T & 1 \end{bmatrix}  \in \real^{k \times k} \mid\gparam \succ 0 \Big\} .
\end{align*} \end{minipage}
}
\vspace{-0.13cm}

The Riemannian exponential map for this submanifold does not have a simple and closed-form expression \citep{calvo1991explicit}, not to mention other submanifolds induced by structured covariance $\vSigma$.
The exponential map also is unknown on  the following rank-one SPD submanifold, 
 
\vspace{-0.5cm}
\begin{center}
\resizebox{0.94\linewidth}{!}{
  \begin{minipage}[t]{\linewidth}
 \begin{align*}
 \mathcal{M} =\Big\{ \gparam =  \begin{bmatrix}a^2 & a \vb^T   \\ a \vb &  \vb\vb^T  +\mathrm{Diag}(\vc^2)   \end{bmatrix}
\in \real^{k \times k} \mid\gparam \succ 0 \Big\}.
\end{align*} \end{minipage} 
} \end{center}
\vspace{-0.25cm}

The existing Riemannian maps defined on the full SPD manifold such as the exponential map and the transport maps cannot be used on SPD submanifolds since these maps are neither computationally efficient in high dimensional cases nor preserve the submanifold structures. In particular, these maps do not guarantee that their output stays on a given SPD submanifold. 

To stay on a submanifold, a retraction map using global coordinate $\gparam$ is proposed as an approximation of the exponential map for the submanifold. Likewise, a vector transport map is proposed to approximate the Riemannian parallel transport map.
However, both retraction and vector transport maps vary from one submanifold to another. It can be difficult to design such maps for a new submanifold. 
A generic approach to designing such maps is to approximate the ODEs.
However, it is computationally challenging to even evaluate the ODEs at a point when the global coordinate $\gparam$ is used, since this requires computing the Christoffel symbols \scalebox{0.85}{ $\Gamma_{\, cb}^{a}( \gparam  )$} (see Appx.~\ref{apd:chris_sym}) arising in the ODEs. These symbols are defined by partial derivatives of the metric \scalebox{0.85}{ $\vF(\gparam)$} in Eq.~\eqref{eq:affinemetric}. Thus, their computation involves complicated 3rd order derivatives of a matrix and makes it hard to
preserve a given submanifold structure.

Other global-coordinate approaches such as the Frank-Wolfe algorithm recast a submanifold as a \emph{closed-set} constraint on an (higher-dimensional) ambient manifold and solve a constrained subproblem induced by the constraint.
The closed-set condition is often needed since the solution of the subproblem should be attainable. 
However, such a constraint varies from one submanifold to another. It can be non-trivial to construct such a closed-set constraint for a given submanifold.  Furthermore, it can be both mathematically and computationally challenging to obtain a closed-form expression to solve the constrained subproblem when the ambient manifold is high-dimensional. This is exactly the case for a SPD submanifold where its (SPD) ambient manifold is often high-dimensional.
Thus, it is essential to have a closed-form solution to the subproblem so that the Frank-Wolfe algorithm on a Riemannian submanifold is computationally efficient without introducing a computationally expensive inner-loop procedure needed for solving the subproblem. 
Unfortunately, many existing Frank-Wolfe methods on submanifolds have to introduce an inner-loop. This is undesirable even when these submanifolds contained in a 
(high-dimensional) manifold are indeed low-dimensional.

The Riemannian gradient computation on a SPD submanifold remains computationally challenging.
The existing Riemannian gradient computation on a full SPD manifold is neither computationally efficient nor straightforwardly 
applicable for a SPD submanifold since a Riemannian gradient on the manifold is not necessarily a 
Riemannian gradient on the submanifold. 
Thus, it is unclear how to efficiently compute a Riemannian gradient \scalebox{0.82}{$\vF^{-1}(\gparam) \vg(\gparam)$} on a SPD submanifold without explicitly inverting the metric, which can be another computationally intensive operation.

\begin{table*}[!t]
  \begin{minipage}[t]{.56\linewidth}
\scriptsize
\resizebox{0.97\textwidth}{!}{  
  \begin{tabular}{l|l|l}
 Operations on SPD in global $\gparam$
    &  Manifold  
    &  Submanifolds
    \\
    \hline
      Riemann. gradient $\vv$ at  $\gparam_0$ 
      &\scalebox{1.0}{ $\gparam_0 \vg  \gparam_0$ }
      & \scalebox{1.0}{$\vF^{-1}(\gparam_0)  \vg $ }%
      \\       
    Riemann. exponential \scalebox{0.99}{$ \RExp(\gparam_0,\vv)$}
      & \scalebox{0.99}{ $\gparam_0^{1 \! / \! 2} \MExp(\gparam_0^{-1 \! / \! 2} \vv \gparam_0^{-1 \! / \! 2}) \gparam_0^{1 \! / \! 2}$}  %
      & Unknown  %
      \\
             
  Riemann. transport \scalebox{0.99}{$ \hat{\transport}_{\sgparam_0\rightarrow\sgparam_1}(\vv)$}
      &\scalebox{0.99}{ $\vE \vv\vE^T; \,\, \vE:=(\gparam_1\gparam_0^{-1})^{1 \! / \! 2}$}%
      &  Unknown   %
    
     \\
   Euclidean transport \scalebox{0.99}{$ {\transport}_{\sgparam_0\rightarrow\sgparam_1}(\vg)$}
      &\scalebox{0.99}{ $ \vH  \vg \vH^T;\,\,  \vH:=\gparam_1^{-1} \vE \gparam_0 $} %
      &  Unknown   %
    
     \\
         
     \hline
\end{tabular}
}
\vspace{-0.1cm}
\caption{Manifold operations compatible with affine-invariant metric $\vF$,\\ where $\MExp(\vN)= \sum_{k=0}^{\infty} \frac{1}{k!} \vN^k $ is the matrix exponential function,\\ and $\vg$ is a (symmetric) Euclidean gradient at $\gparam_0$.  On submanifolds,\\ $\gparam$ denotes learnable vectorized parameters and $\vg$ is also vectorized.   }
\label{tab:spd_table}
  \end{minipage}\vspace{-0.45cm} %
    \begin{minipage}[t]{.4\linewidth}
    \centering
\scriptsize
\begin{tabular}{l|l|l}

    & SNC
    & GNC
    \\
    \hline
          
    $\gparam $
      &  $\vpsi_{\sgparam^{\text{(cur)}}}(  \lparam ) $  %
      & $\vphi_{\sgparam^{\text{(cur)}}}(  \lparam  )$     %
      \\

   $\gparam^{\text{(cur)}}$
      & $\vpsi_{\sgparam^{\text{(cur)}}}(  \lparam_0  )$ %
      & $\vphi_{\sgparam^{\text{(cur)}}}(  \lparam_0  )$ %
      \\

   $ \vF(  \lparam_0  )$
      & $\vI $%
      & $\vI$ %
      \\
  
   $\Gamma_{\, bc}^{a}( \lparam_0  )$
      & $0$
      &  can be nonzero
     \\
     
     \hline
\end{tabular}
\vspace{-0.1cm}
   \caption{Properties of Riemannian normal coordinates $\lparam$ defined at $\gparam^{\text{(cur)}}$, where the original $\lparam_0=\mathbf{0}$ represents $\gparam^{\text{(cur)}}$, $\vpsi$ is defined by the Riemannian exponential map, and $\vphi$ is a diffeomorphic and isometric map.}
\label{tab:normal_coord} 
  \end{minipage} %
\end{table*}

\vspace{-0.25cm}
\subsection{Natural-gradient Descent and its Challenges}
\vspace{-0.2cm}

A practical approach is natural-gradient descent (NGD), which approximates the Riemannian exponential map by ignoring the Christoffel symbols.
A NGD update is a linear approximation of the update of Eq.~\eqref{eq:rgd} given by

\vspace{-0.25cm}
\resizebox{0.95\linewidth}{!}{
  \begin{minipage}[t]{\linewidth}
\begin{align}
 \text{NGD}:\,\, \gparam \leftarrow \gparam -\stepsize \vF^{-1}(\gparam) \vg(\gparam) .
\end{align}\end{minipage}
}

\vspace{-0.15cm}
This approximation is also known as the Euclidean retraction map \citep{jeuris2012survey}.
Unfortunately, NGD in the global coordinate $\gparam$ does not guarantee that the update stays on a manifold even in the full SPD case. Moreover, computing Riemannian gradients remains challenging due to the metric inverse.
Structured NGD \citep{lin2021tractable}  addresses the SPD constraint of Gaussians for Bayesian posterior approximations by performing NGD on local coordinates.
Local coordinates could enable efficient Riemannian gradient computation by simplifying the metric inverse computation.
However, it is nontrivial to incorporate momentum in structured NGD due to the metric and the use of moving coordinates.
We address this issue and develop practical Riemannian momentum methods by using generalized normal coordinates.
Using our normal coordinates, we will explain and generalize structured NGD from a manifold optimization perceptive.
We further expand the scope of structured NGD to SPD submanifolds by going beyond the Bayesian settings. Our local-coordinate approach gives a computationally efficient paradigm for a class of SPD submanifolds where it can be nontrivial to design an efficient retraction map or solve a closed-form constrained subproblem
using a global coordinate while keeping the Riemannian gradient computation efficient and inverse-free.

\vspace{-0.3cm}
\subsection{Standard Normal Coordinates (SNCs) }
\vspace{-0.2cm}
{\bf Defn 1:}
A metric $\vF$ is orthonormal at $\lparam_0=\mathbf{0}$ if $\vF(\lparam_0) = \vI$, where $\vI$ is the identity matrix. 

\vspace{-0.08cm}
Normal coordinates can simplify calculations in differential geometry and general relativity.
However, these coordinates are seldom studied in the optimization literature. Given a reference point $\gparam^{\text{(cur)}}$, the standard (Riemannian) normal  coordinate (SNC) $\lparam $ at $\gparam^{\text{(cur)}}$ is defined as below, where $\gparam^{\text{(cur)}}$ and $\vF^{-1/2}(\gparam^{\text{(cur)}})$ are treated as constants in  coordinate $\lparam$:

\vspace{-0.2cm}
\resizebox{0.99\linewidth}{!}{
  \begin{minipage}[t]{\linewidth}
\begin{align}
    \gparam = \vpsi_{\sgparam^{\text{(cur)}}}(  \lparam  ):=\RExp(\gparam^{\text{(cur)}}, {\color{red} \vF^{-1/2}(\gparam^{\text{(cur)}})} \lparam ). \label{eq:standard_normal_coord}
\end{align} \end{minipage}
}

\scalebox{0.85}{$\vF^{-1/2}(\gparam^{\text{(cur)}})$} in Eq.~\eqref{eq:standard_normal_coord} is essential since it 
orthonormalizes the metric at $\lparam_0$ (i.e., \scalebox{0.85}{$\!\vF(\lparam_0)\!=\!\vI$})
and will simplify the Riemannian gradient computation in Eq.~\eqref{eq:rgd_standard_norm} and \eqref{eq:local_rgd_mom}.

\citet{lezcano2019trivializations,lezcano2020adaptive} consider (non-normal) local coordinates defined by the Riemannian exponential such as 
\scalebox{0.85}{$\gparam\!=\!\RExp(\gparam^{\text{(cur)}}, \vlambda )$}. However, the metric is not orthonormal in
their coordinates (i.e., \scalebox{0.85}{$\vF(\vlambda_0)\!\equiv \!\vF(\gparam^{\text{(cur)}})\!\neq\! \vI$} at \scalebox{0.85}{$\vlambda_0\!=\!\mathbf{0}$}). 
They use an ad-hoc metric $\vI$ instead of \scalebox{0.85}{$\vF(\vlambda_0)$} at $\vlambda_0$. Thus, their approach does not preserve the predefined metric $\vF$. Importantly, the Riemannian exponential is incompatible with the ad-hoc metric as the exponential is defined by metric $\vF$.
Another issue is that
the ad-hoc metric in their approach does not even obey the metric (tensor) transform rule needed for the change of their local coordinates at each iteration. 
In Sec.~\ref{sec:introducing_normal}, we will address these issues
via \emph{metric-aware} orthonormalizations and propose \emph{metric-preserving} trivializations  
even when the Riemannian exponential is unknown on submanifolds.

\begin{figure}%
 \vspace{-0.3cm}
    \centering
    \subfloat[\centering \scriptsize Normal coordinate $\lparam$ at  $\gparam^{\text{(cur)}}$]{{\includegraphics[width=0.4\linewidth]{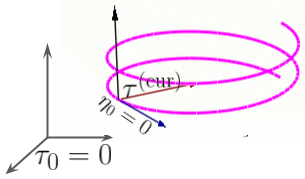} }}%
    \qquad
    \subfloat[\centering \scriptsize Normal coordinate $\vxi$ at  $\gparam^{\text{(new)}}$]{{\includegraphics[width=0.4\linewidth]{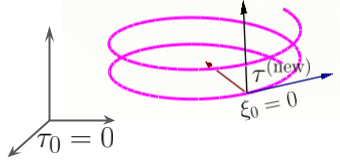} }}%
    \vspace{-0.2cm}
    \caption{A (orthonormal) SNC/GNC is generated at each iteration.}%
    \vspace{-0.6cm}
\label{fig:normal_coord_changes}
\end{figure}

A SNC has nice computational properties, summarized in Table \ref{tab:normal_coord}.
In coordinate $\lparam$, the origin  $\lparam_0:=\mathbf{0}$ represents the point  $\gparam^{\text{(cur)}}=\vpsi_{\sgparam^{\text{(cur)}}}( \lparam_0 )$ as illustrated in Fig.~\ref{fig:normal_coord_changes}.

RGD in Eq.~\eqref{eq:rgd} can be reexpressed as (Euclidean) gradient descent (GD) in local coordinate $\lparam$ since the metric $ \vF(\lparam_0)$ at $\lparam_0$ becomes the standard Euclidean metric $\vI$: 

\vspace{-0.05cm}
\hspace{3.5mm} \resizebox{0.87\linewidth}{!}{
  \begin{minipage}{\linewidth}
\begin{align}
\text{NGD/GD}:\,\,& \lparam_1   \leftarrow \lparam_0 -  \stepsize \vF^{-1}(\lparam_0)  \vg(\lparam_0) =\mathbf{0}  - \stepsize \vI^{-1} \vg(\lparam_0)  ,  \nonumber  \\
&\gparam^{\text{(new)}}    \leftarrow  \vpsi_{\sgparam^{\text{(cur)}}}(  \lparam_1  ), \label{eq:rgd_standard_norm}
\end{align} \end{minipage}
}

\vspace{-0.2cm}
where $\vg(\lparam_0) =\vF^{-1/2}(\gparam^{\text{(cur)}}) \vg(\gparam^{\text{(cur)}})$ is a Euclidean gradient evaluated at $\lparam_0$.
Since the metric $ \vF(\lparam_0)$ is orthonormal,  the GD update is also a NGD update in coordinate $\lparam$.  The orthonormalization of the metric makes it easy to add momentum into RGD while preserving 
the metric.

In the SPD case, Eq.~\eqref{eq:standard_normal_coord} can be simplified as $\vF^{-1/2}(\gparam) \lparam =    \gparam^{1/2}   \lparam   \gparam^{1/2}$, where $\lparam$ is a  symmetric matrix.  
Recall that under global parameterization $\gparam$, $\vF(\gparam^{\text{(cur)}} )\neq \vI$ for any $\gparam^{\text{(cur)}} \neq  \vI$.
However, by Eq.~\eqref{eq:affinemetric}, the metric $\vF(\lparam)$ is orthonormal, at local parameterization $\lparam_0$ associated to $\gparam^{\text{(cur)}}=\vpsi_{\sgparam^{\text{(cur)}}}(\lparam_0)$  as shown below.

\vspace{-0.25cm}
\hspace{3mm} \resizebox{0.89\linewidth}{!}{
  \begin{minipage}[t]{\linewidth}
\begin{align}
   \vF(\lparam_0) 
   = -2 \Unmyexpect{\gauss(\mathbf{0},  \gparam ))}\big[ \nabla_\slparam^2 \log \gauss(\mathbf{0}, \gparam )\big]\big|_{\slparam=\slparam_0} =\vI ,
\end{align}
\end{minipage}
}

\vspace{-0.15cm}
where  $\gparam =\vpsi_{\sgparam^{\text{(cur)}}}(\lparam)$. 
By Table \ref{tab:spd_table}, $\vpsi_{\sgparam^{\text{(cur)}}}(\lparam)$ in SNC $\lparam$ has a closed-form, where  $\MExp(\cdot)$ is the matrix exponential,

\vspace{-0.32cm}
\resizebox{0.93\linewidth}{!}{
  \begin{minipage}[t]{\linewidth}
\begin{align}
 \vpsi_{\sgparam^{\text{(cur)}}}(\lparam)   &= \big(\gparam^{\text{(cur)}}\big)^{1/2} \MExp( \lparam   ) \big(\gparam^{\text{(cur)}}\big)^{1/2}  .
\label{eq:orth_exp}
\end{align}
\end{minipage}
}
\vspace{-0.07cm}

Eq.~\eqref{eq:orth_exp} is only obtainable for SPD manifolds  if we use SNC $\lparam$, and in particular, the metric must be orthonormal at $\lparam_0$.

In the case of SPD submanifolds, it is hard to use a SNC as it relies on the intractable Riemannian exponential map, as seen in Eq.~\eqref{eq:standard_normal_coord}.
Recall that the Riemannian exponential map is defined by solving a non-linear ODE and it varies from one submanifold to another submanifold.
However, we will make use of the matrix exponential\footnote{It is a Lie-group exponential map. Importantly, this map remains the same on matrix subgroups.
In contrast, the Riemannian exponential map varies from one submanifold to another.
} in Eq.~\eqref{eq:orth_exp} to generalize normal coordinates and to reduce the computation cost of Riemannian gradients on SPD submanifolds.

\vspace{-0.2cm}
\section{Generalized  Normal  Coordinates (GNCs)}
\label{sec:introducing_normal}
\vspace{-0.2cm}
We propose new (metric-aware orthonormal) coordinates to simplify the momentum computation on a (sub)manifold with a predefined metric.
Inspired by SNCs, we identify their properties that enable efficient Riemannian optimization, and generalize normal coordinates by defining new coordinates satisfying these properties on the (sub)manifold.

\vspace{-0.1cm}
{\bf Defn 2:}
A local coordinate is a generalized (Riemannian) normal coordinate (GNC) at point $\gparam^{\text{(cur)}}$ denoted by $\gparam=\vphi_{\sgparam^{\text{(cur)}}}(\lparam)$  if
$\vphi_{\sgparam^{\text{(cur)}}}(\lparam)$ satisfies all the following assumptions. 

\vspace{-0.15cm}
{\bf Assumption 1:} \emph{The origin $\lparam_0=\mathbf{0}$ represents point $\gparam^{\text{(cur)}}= \vphi_{\sgparam^{\text{(cur)}}}(\lparam_0)$ in  coordinate $\lparam$.}

\vspace{-0.15cm}
{\bf Assumption 2:} \emph{The metric $\vF(\lparam_0)$ is orthonormal at $\lparam_0$.}

\vspace{-0.15cm}
{\bf Assumption 3:} \emph{ The map  $ \vphi_{\sgparam^{\text{(cur)}}}(\lparam)$ is bijective.   $\vphi_{\sgparam^{\text{(cur)}}}$ and $\vphi^{-1}_{\sgparam^{\text{(cur)}}}$ are smooth (i.e. $ \vphi_{\sgparam^{\text{(cur)}}}(\lparam)$ is diffeomorphic).  }

\vspace{-0.15cm}
{\bf Assumption 4:} \emph{The parameter space of $\lparam$ is a vector space. }

Assumption 1 enables simplification using the chain rule of high-order derivative calculations (i.e., the metric and Christoffel symbols in Appx.~\ref{apd:simple_fim},~\ref{apd:1st_approx_trans_eg}) by evaluating at zero, which is useful when computing Christoffel symbols.
Assumption 2 enables metric preservation in GD updates by dynamically orthonormalizing the metric.
By Assumption 3, (sub)manifold constraints disappear in these coordinates, and Assumption 4 ensures that scalar products and vector additions are well-defined, so that GD updates are valid.
For example, SNCs satisfy Assumptions 1-2 (see Table~\ref{tab:normal_coord}) and Assumption  3 due to the Riemannian exponential.
On a complete manifold, SNCs satisfy Assumption 4 \citep{absil2009optimization}. 
These assumptions make it easy to design new coordinates without the Riemannian exponential.  
Assumptions 1-4 together imply that \scalebox{0.78}{ $ \vphi_{\sgparam^{\text{(cur)}}}(\lparam)$} is a metric-preserving/isometric map from the tangent space at \scalebox{0.8}{$\gparam^{\text{(cur)}}$} with Riemannian metric \scalebox{0.8}{$\vF(\gparam^{\text{(cur)}})$} to a (local) coordinate space of \scalebox{0.83}{$\lparam$} identified as a tangent space at \scalebox{0.8}{$\lparam_0$} with the Euclidean metric \scalebox{0.8}{$\vF(\lparam_0)\!=\!\vI$} such as a matrix subspace in Sec.~\ref{sec:expanding}. 
This map is a \emph{metric-preserving} trivialization as (sub)manifold constraints are trivialized and the metric is preserved and orthonormal
in the coordinate space of $\lparam$. Thus, GD becomes NGD in the space.
Our approach differs from existing trivialization suggested by \citet{lezcano2019trivializations,lezcano2020adaptive}, as GD does not become NGD in their coordinates.
Related local-coordinate approaches  \citep{lezcano2019trivializations,lezcano2020adaptive,lin2021tractable,lin2021snd} do not orthonormalize the metric so adding Riemannian momentum can be inefficient.

We will propose GNCs to work with NGD/GD by satisfying Assumptions 1-4 while reducing the computational cost by ignoring  Christoffel symbols (see Table~\ref{tab:normal_coord}).
A GD update in a GNC is an approximation of the RGD update in Eq.~\eqref{eq:rgd_standard_norm}. This GD update can be understood as a NGD update with special local reparametrizations. 
As will be shown in Sec.~\ref{sec:sngd_special}, the GD resembles structured NGD in Gaussian cases, both of which ignore the symbols,

\vspace{-0.2cm}
\hspace{2.5mm}\resizebox{0.91\linewidth}{!}{
  \begin{minipage}[t]{\linewidth}
\begin{align}
\text{NGD/GD}:&\,
 \lparam_1  \leftarrow \lparam_0 -  \stepsize \vF^{-1}(\lparam_0)  \vg(\lparam_0) =   \mathbf{0}-\stepsize \vI^{-1} \vg(\lparam_0), \nonumber  \\
&\gparam^{\text{(new)}}    \leftarrow   \vphi_{\sgparam^{\text{(cur)}}}(  \lparam_1  ).
\end{align} 
\end{minipage} }

\vspace{-0.03cm} 
In the full SPD case,  \citet{lin2021tractable} establish a connection between the update in $\lparam$ and a retraction map in $\gparam$. 

\vspace{-0.4cm} 
\subsection{Adding Momentum using GNCs}
\label{sec:momentum}
\vspace{-0.2cm}
Since the metric is orthonormal at $\lparam_0$, we propose simply adding (Euclidean) momentum in the GD update in our normal coordinates. 
At each iteration, given the current point \scalebox{0.83}{$\gparam^{\text{(cur)}}$} in the global coordinate, we generate a GNC $\lparam$ at \scalebox{0.83}{$\gparam^{\text{(cur)}}$} and perform the update in coordinate $\lparam$, where we first assume momentum \scalebox{0.83}{$\vw^{ (\slparam_0)}$} is given at the current iteration.

\begin{tcolorbox}[enhanced,colback=white,%
    colframe=red!75!black, attach boxed title to top right={yshift=-\tcboxedtitleheight/2, xshift=-1.25cm}, title=our update with momentum in   GNC $\lparam$, coltitle=red!75!black, boxed title style={size=small,colback=white,opacityback=1, opacityframe=0}, size=title, enlarge top initially by=-\tcboxedtitleheight/2]
    \vspace{0.28cm}
    \resizebox{\linewidth}{!}{
  \begin{minipage}[t]{\linewidth}
\begin{align} 
\text{Momentum}:&\,
\vm^{(\slparam_0) } \leftarrow \alpha \vw^{ (\slparam_0)} + \stepsize   \vg(\lparam_0) , \nonumber \\
\text{NGD/GD}:&\, \lparam_1   \leftarrow \lparam_0 -  \vF^{-1}(\lparam_0)\vm^{(\slparam_0)}=\mathbf{0}-\vI^{-1}\vm^{(\slparam_0)} , \nonumber \\
 &\gparam^{\text{(new)}}  \leftarrow  \vphi_{\sgparam^{\text{(cur)}}}(\lparam_1), %
\label{eq:local_rgd_mom}
\end{align} 
\end{minipage} }
\end{tcolorbox}

\vspace{-0.06cm}
where $\alpha$ is the momentum weight and $\stepsize$ is the stepsize.

We now discuss how to include momentum in moving (local) coordinates $\lparam$ and $\vxi$, where  $\lparam$ and $\vxi$ are GNCs defined at $\gparam^{\text{(cur)}}$ and $\gparam^{\text{(new)}}$, respectively. Note that $\gparam^{\text{(new)}}$ is represented by $\lparam_1$ in current coordinate $\lparam$ and by $\vxi_0$ in new coordinate $\vxi$.
To compute momentum $\vw^{ (\xi_0) }$ in coordinate $\vxi$ at the next iteration, we perform the following two steps.

{\bf Step 1: (In Coordinate $\lparam$)} 
We transport momentum $ \vm^{ (\slparam_0) }$ at point $\gparam^{\text{(cur)}}$ via the Euclidean transport map to point $\gparam^{\text{(new)}}$, which is similar to the transport step in Eq.~\eqref{eq:rgd_mom_eg}.

Since performing NGD/GD alone ignores Christoffel symbols, we suggest ignoring the symbols in the approximation of the Euclidean transport map 
${\transport}_{\slparam_0 \rightarrow \slparam_1 }( \vm^{ (\slparam_0) } )$ using coordinate $\lparam$. This gives the following update:

\begin{tcolorbox}[enhanced,colback=white,%
    colframe=red!75!black, attach boxed title to top right={yshift=-\tcboxedtitleheight/2, xshift=-1.25cm}, title=momentum  transport in GNC $\lparam$, coltitle=red!75!black, boxed title style={size=small,colback=white,opacityback=1, opacityframe=0}, size=title, enlarge top initially by=-\tcboxedtitleheight/2]
    \vspace{0.18cm}
    \resizebox{\linewidth}{!}{
  \begin{minipage}{\linewidth}
\begin{align}
\text{Approximated Transport}:\,\,\vw^{ (\slparam_1) } \leftarrow  \vm^{ (\slparam_0) } . %
\label{eq:transport_approx}
\end{align} \end{minipage} }
\end{tcolorbox}

\vspace{-0.01cm}
The approximation keeps the dominant term of the map and ignores negligible terms.
In a global coordinate,
this approximation is known as the Euclidean vector transport map \citep{jeuris2012survey}  and the vector-transport-free map \citep{godaz2021vector}.
A similar approximation is also suggested in Riemannian sampling \citep{girolami2011riemann} to avoid solving an implicit leapfrog update. 
Using GNCs, we can make a better approximation by adding the second dominant term (see  Appx.~\ref{apd:1st_approx_trans_eg}) defined by the Christoffel symbols. 
For example, in SPD cases, the second dominant term (see Eq.~\eqref{eq:chris_error} in Appx.~\ref{apd:1st_approx_trans_eg}) can be explicitly computed and is negligible compared to the first term. The computation can be similarly carried out on submanifolds. Moreover, if GNC $\lparam$ is a symmetric matrix as will be shown in Eq.~\eqref{eq:gen_normal_param}, the second dominant term vanishes even if the symbols are non-vanishing. On the other hand, it is nontrivial to compute the second term in a global coordinate $\gparam$ on submanifolds.

{\bf Step 2: (At Point $\gparam^{\text{(new)}}$)} 
We coordinate-transform momentum $\vw^{ (\slparam_1) }$ from coordinate $\lparam$  to 
coordinate $\vxi$ and return the transformation as $\vw^{ (\xi_0) }$, where $\lparam_1$ and $\vxi_0$ represent  $\gparam^{\text{(new)}}$.

By construction of GNCs (shown in Fig.~\ref{fig:normal_coord_changes}), coordinates $\lparam$ and $\vxi$ represent the global coordinate $\gparam$ as 
 $\gparam = \vphi_{\sgparam^{\text{(cur)}}}(\lparam)
 = \vphi_{\sgparam^{\text{(new)}}}(\vxi)$, where $\vxi$ is the GNC  associated to $\gparam^{\text{(new)}}$ at the next iteration.
We transform Euclidean momentum $\vw^{ (\slparam_1) }$ as a Euclidean (gradient) vector from coordinate $\lparam$ to coordinate $\vxi$ via the (Euclidean) chain rule,

\vspace{-0.05cm}
\begin{tcolorbox}[enhanced,colback=white,%
    colframe=red!75!black, attach boxed title to top right={yshift=-\tcboxedtitleheight/2, xshift=-0.5cm}, title=momentum coordinate-transform for new GNC $\vxi$, coltitle=red!75!black, boxed title style={size=small,colback=white,opacityback=1, opacityframe=0}, size=title, enlarge top initially by=-\tcboxedtitleheight/2]
    \vspace{0.22cm}
\resizebox{0.95\linewidth}{!}{
  \begin{minipage}{\linewidth}
\begin{align}
     \left( \vw^{ (\xi_0)}\right)^{T}
      \!=\! \left(\vw^{ (\slparam_1) } \right)^{T} \vJ(\vxi_0)  ;\,\,\, \vJ(\vxi)\!:=\! \frac{ \partial \lparam } {\partial \vxi} ,
    \label{eq:jacobian_transform}
\end{align} \end{minipage}
}\end{tcolorbox}

\vspace{-0.2cm}
where $\vxi_0=\mathbf{0}$, $\lparam=\vphi^{-1}_{\sgparam^{\text{(cur)}}} \circ \vphi_{\sgparam^{\text{(new)}}}(\vxi) $, and $\vJ(\vxi) $ is the Jacobian. Thanks to GNCs, the vector-Jacobian product in Eq.~\eqref{eq:jacobian_transform} can be simplified by evaluating at $\vxi_0=\mathbf{0}$.

We can see that our update using local coordinates in Eq.~\eqref{eq:local_rgd_mom}-\eqref{eq:jacobian_transform} is a practical approximation of update~\eqref{eq:rgd_mom_eg} using a global coordinate. The Euclidean transport map is required for the use of the (Euclidean) chain rule and simplification of the vector-Jacobian product in Eq.~\eqref{eq:jacobian_transform}. Our update shares the same spirit of Cartan's method of moving frames \citep{ivey2003cartan} by using only Euclidean/exterior derivatives. 
The Christoffel symbols could be computed via a Lie bracket \footnote{There is a Lie group structure for the coordinate-transformation in the frame bundle of a general (sub)manifold.} due to Cartan's structure equations and the Maurer-Cartan form \citep{piuze2015maurer}. 

\vspace{-0.2cm}
\subsection{Designing GNCs for SPD Manifolds}
\label{sec:demystifying}
\vspace{-0.1cm}
We describe how to design GNCs on SPD manifolds.
This procedure explains the construction of existing coordinates in  \citet{lin2021tractable,godaz2021vector}.

To mimic the SNC in Eq.~\eqref{eq:orth_exp},  consider the matrix factorization $\gparam^{\text{(cur)}}= \vA^{\text{(cur)}}\big(\vA^{\text{(cur)}}\big)^T$, where  $\vA^{\text{(cur)}}$  is  invertible (not a Cholesky). 
This asymmetric factorization 
contains a \emph{matrix Lie group structure} in $\vA^{\text{(cur)}}$ for  submanifolds  while
 the symmetric one (i.e., $\big(\gparam^{\text{(cur)}}\big)^{1/2}$)  in Eq.~\eqref{eq:orth_exp} does not. 
 For example, the product of two distinct symmetric matrices is often not symmetric. Thus, the set of symmetric matrices does not have the matrix Lie group structure. 
 The Christoffel symbols are non-vanishing due to the asymmetry. Nevertheless, this factorization allows us
to obtain a
coordinate by approximating the map $\vpsi_{\sgparam^{\text{(cur)}}}$ in  the SNC as

\vspace{-0.1cm}
\hspace{0.8mm} \resizebox{0.95\linewidth}{!}{
  \begin{minipage}{1.1\linewidth}
\begin{align}
 \scriptstyle \vphi_{\sgparam^{\text{(cur)}}}(\lparam) & \scriptstyle \ : =  \vA^{\text{(cur)}} \MExp(\lparam) \left(\vA^{\text{(cur)}}\right)^T  \nonumber \\
 &\scriptstyle \ = \vA^{\text{(cur)}} \MExp(\half \lparam) \MExp^T(\half \lparam)\left(\vA^{\text{(cur)}}\right)^T   
 = \vA \vA^T, 
\label{eq:gen_normal_param}
\end{align}\end{minipage} }

\vspace{-0.2cm}
where  $\vA:=\vA^{\text{(cur)}} \MExp(\half \lparam)$ and  $\lparam$ is a symmetric matrix. Factorization $\gparam= \vA\vA^T$ can be non-unique in $\vA$. We only require   coordinate $\lparam$ to be unique instead of $\vA$. 
To satisfy uniqueness  in $\lparam$ required in Assumption 3, we can restrict $\lparam$ to be in a subspace of $\real^{k \times k}$ such as the symmetric matrix space.
Coordinate $\lparam$ is a GNC at $\gparam^{\text{(cur)}}$ (shown in Appx.~\ref{apd:gnormal_spd}).
Using other factorizations, we obtain more GNCs:

\vspace{-0.5cm}
\begin{itemize} 
\item
   \scalebox{0.9}{ $ \vphi_{\sgparam^{\text{(cur)}}}(\lparam)= \vB^{-T}\vB^{-1}$, with
$\vB:=\vB^{\text{(cur)}}\MExp(-\half \lparam) $}  \vspace{-0.2cm}
\item  
  \scalebox{0.9}{  $ \vphi_{\sgparam^{\text{(cur)}}}(\lparam)=\vC^T \vC$, with $\vC:=\MExp(\half \lparam)\vC^{\text{(cur)}} $ }
\end{itemize} 
\vspace{-0.36cm}
where $\lparam$ is a symmetric matrix in both cases.

The GNC considered in Eq.~\eqref{eq:gen_normal_param} is similar to local coordinates in structured NGD, where $\vA$ is referred to as an auxiliary coordinate. 
The authors of structured NGD \citep{lin2021tractable} introduce a similar local coordinate as  \scalebox{0.9}{ $\vA\!:=\!\vA^{\text{(cur)}}\MExp( \lparam)$} in Gaussian cases without providing the construction and mentioning other local coordinates. 
The metric is not orthonormal in their coordinates.
Our procedure sheds light on the construction and the role of local coordinates in structured NGD.
For example, our construction explains why  
 \scalebox{0.9}{$\vA\!=\!\vA^{\text{(cur)}}\MExp(\lparam)$} instead of  \scalebox{0.9}{$\vA\!=\!\MExp(\lparam) \vA^{\text{(cur)}}$} is used in structured NGD for the factorization  \scalebox{0.9}{$\gparam\!=\!\vA\vA^T$}.
Using  \scalebox{0.9}{$\vA\!=\!\MExp(\lparam) \vA^{\text{(cur)}}$} in structured NGD makes it difficult to compute natural-gradients since the metric is not orthonormal.  
In contrast, our coordinates explicitly orthonormalize the metric, which makes it easy to compute (natural) gradients and include momentum.  

Using the GNC  in Eq.~\eqref{eq:gen_normal_param}, our update in Eq.~\eqref{eq:local_rgd_mom}-\eqref{eq:jacobian_transform} can be simplified as shown in Appx.~\ref{apd:simple_update_spd}. By using our GNCs, we can obtain inverse-free updates.
When the matrix exponential $\MExp$ is approximated, our updates even become matrix-decomposition-free updates, which is useful in low numerical settings.

\citet{godaz2021vector} consider a special case for full SPD manifolds with symmetric $\lparam$ and a unique factor $\vA$.  However,
their method is non-constructive and limited to SPD manifolds. 
Our construction generalizes their method by allowing $\lparam$ to be an asymmetric matrix,  using a non-unique factor $\vA$, and extending their method to SPD submanifolds.

\vspace{-0.2cm}
\subsection{Designing  GNCs for SPD Submanifolds}
\label{sec:expanding}
\vspace{-0.1cm}
Although SNCs are unknown on SPD submanifolds, GNCs allow us to work on a class of SPD submanifolds by noting that $\vA$ is in the general linear  group $\mathrm{GL}^{k \times k}$ known as a \emph{matrix Lie group}.
Matrix structures are preserved under matrix multiplication and the matrix exponential. 
The coordinate space of $\lparam$ is a subspace of the tangent space of $\MExp(\lparam)$ at $\MExp(\lparam_0)=\vI$ known as the \emph{Lie algebra}. Recall that the Lie algebra of $\mathrm{GL}^{k \times k}$ is a square matrix space $\mathcal{R}^{k \times k }$.
Thus, Assumption 4 is satisfied. 
One example of a structured group is to consider a (unique) Cholesky factor $\vA$. In this case, $\vA$ is in a lower-triangular subgroup of $\mathrm{GL}^{k \times k}$. This group structure induces a SPD manifold.  Thus, we can obtain a new GNC for the SPD manifold by restricting $\lparam$ to be a lower-triangular matrix with proper scaling factors, where the lower-triangular matrix space of $\lparam$ is a subspace of the Lie algebra $\real^{k \times k}$.

The above observations and the example let us consider the following class of SPD submanifolds\footnote{Another Lie group structure is induced by a SPD submanifold.}:

\scalebox{0.82}{
\hspace{-0.2cm}\begin{minipage}{1.25\linewidth}
\begin{align}
 \mathcal{M}=\Big\{\!\gparam\!=\!\vA \vA^T
\!  \succ \! 0  \mid\vA \! \in \! \text{ Connected Subgroup of } \mathrm{GL}^{k \times k} \Big\}.
\label{eq:submanifolds}
\end{align} 
\end{minipage}}

Each of the submanifolds is induced by a (fixed-form) Lie subgroup. The Lie subgroup is
constructed by a GNC space of the submanifold as a subspace of the Lie algebra of the general linear group. We will construct diffeomorphic maps using the matrix (Lie-group) exponential to approximate the Riemannian exponential map on these submanifolds.
Each of the diffeomorphic maps induces a GNC.

We propose GNCs for these submanifolds so that our update preserves the Lie subgroup structure in $\vA$ by performing  updates in the GNCs as a subspace of the Lie algebra. Thanks to the GNCs, we can use $\vA$ to represent each of the submanifolds $\gparam$. 
Note that we can directly update and maintain $\vA$ instead of $\gparam$. Thus, 
a GNC for each of the submanifolds is readily available as $\gparam=\vphi_{\sgparam^{\text{(cur)}}}(\lparam)=\vA\vA^T$ in our local-coordinate approach. 
On the other hand, existing Riemannian methods directly update a global coordinate $\gparam$ and thus, a costly matrix decomposition such as $\gparam = \vA \vA^T$ may be required to satisfy the submanifold constraint. 

We give two examples of SPD submanifolds, where we construct GNCs. GNCs can be constructed for other submanifolds   considered in \citet{lin2021tractable}. For example, we will use a Heisenberg SPD submanifold suggested by \citet{lin2021tractable} in our experiments.
In Sec.~\ref{sec:sngd_special}, we present our update without the Bayesian and Gaussian assumptions. Our update recovers structured NGD as a special case by making a Gaussian assumption.
In Sec.~\ref{sec:low_rank_spd}, we demonstrate the scalability of our approach.

\vspace{-0.2cm}
\subsubsection{Gaussian Family as a SPD submanifold}
\label{sec:sngd_special}
\vspace{-0.1cm}

We will first consider a SPD submanifold in a non-Bayesian and non-Gaussian setting.
We then show how our update relates to structured NGD in Gaussian cases where Gaussian gradient identities are available.
Thus, the structured NGD update on a Gaussian family is a special case of our update on a higher-dimensional SPD submanifold. 

Consider the SPD submanifold, where $k=d+1$,

\resizebox{0.89\linewidth}{!}{
  \begin{minipage}{\linewidth}
\begin{align*}
 \mathcal{M}=\left\{ \gparam = \begin{bmatrix} \vV & \vmu \\ \vmu^T & 1 \end{bmatrix}  \in \real^{k \times k} \mid\gparam \succ 0 \right\} .
\end{align*}  \end{minipage}
}

Note that \scalebox{0.75}{ $\gparam^{-1}=  \begin{bmatrix} \vSigma^{-1}  & -\vSigma^{-1}\vmu \\ -\vmu^T\vSigma^{-1} & 1+ \vmu^T\vSigma^{-1}\vmu \end{bmatrix}  $, for $\gparam \in \mathcal{M}$ } and   $\vSigma:=\vV - \vmu \vmu^T$. Thus, $\vSigma \succ 0 $ since $\gparam^{-1} \succ 0$. Then, letting $\vSigma=\vL\vL^T$, $\mathcal{M}$ can be reexpressed as 

\resizebox{0.94\linewidth}{!}{
  \begin{minipage}{\linewidth}
\begin{align*}
 \mathcal{M}=\Big\{ \gparam = \vA \vA^T   \mid \vA:= \begin{bmatrix} \vL &\vmu \\ \mathbf{0} & 1 \end{bmatrix}, \vL \in \mathrm{GL}^{d \times d} \Big\} .
\end{align*}\end{minipage}
}

\vspace{-0.1cm}
Observe that $\vA$ is  in a  subgroup of \scalebox{0.86}{$\mathrm{GL}^{k \times k}$}.
We construct a GNC \scalebox{0.86}{$\vphi_{\sgparam^{\text{(cur)}}}(\lparam) = \vA \vA^T$} similar to the one in Eq.~\eqref{eq:gen_normal_param}, where 

\resizebox{0.94\linewidth}{!}{
  \begin{minipage}{\linewidth}
\begin{align}
\vA=\vA^{\text{(cur)}}\MExp\left(\begin{bmatrix} { \color{red} \half}  \lparam_L &{ \color{red} \frac{1}{\sqrt{2}} } \lparam_{\mu}\\ \mathbf{0} & 0 \end{bmatrix}\right), %
  \label{eq:gnormal_coord1}
\end{align} \end{minipage}
}

and \scalebox{0.8}{$\lparam= \{ \lparam_L , \lparam_{\mu} \}  $},   \scalebox{0.8}{$\lparam_L \in \real^{d \times d}$} is a symmetric matrix so that Assumption 3 is satisfied. 
The scalars highlighted in red are to satisfy Assumption 2 so that the metric is orthonormal.  

There is another GNC 
$\vphi_{\sgparam^{\text{(cur)}}} (\lparam) = \vA \vA^T$ where

\hspace{3.5mm} \resizebox{0.9\linewidth}{!}{
  \begin{minipage}{\linewidth}
\begin{align}
\vA= &\vA^{\text{(cur)}} \begin{bmatrix}  \MExp({\color{red} \half } \lparam_L) &{\color{red} \frac{1}{\sqrt{2} } }  \lparam_\mu  \\
    \mathbf{0} & 1 \end{bmatrix} \nonumber \\
   =&   \begin{bmatrix} \vL^{\text{(cur)}} \MExp(\half \lparam_L) &\vmu^{\text{(cur)}}+ \frac{1}{\sqrt{2} }  \vL^{\text{(cur)}}\lparam_\mu  \\
    \mathbf{0} & 1 \end{bmatrix} .
    \label{eq:gnormal_coord2}
\end{align} \end{minipage}
}

We can obtain  Eq.~\eqref{eq:gnormal_coord2} from   Eq.~\eqref{eq:gnormal_coord1} (shown in Appx.~\ref{apd:sngd_coordinate}).   

Using the GNC in either Eq.~\eqref{eq:gnormal_coord1} or Eq.~\eqref{eq:gnormal_coord2}, the Euclidean gradient needed in Eq.~\eqref{eq:local_rgd_mom} is given by

\resizebox{0.88\linewidth}{!}{
  \begin{minipage}{\linewidth}
\begin{align*}
    \vg(\lparam_0) =  \{   \vL^T \vg_1  \vL, \, \sqrt{2} \vL^T (\vg_1 \vmu + \vg_2 ) \}, 
\end{align*} 
\end{minipage}}

\vspace{-0.04cm}
where $\vL=\vL^{\text{(cur)}}$, $\vmu=\vmu^{\text{(cur)}}$, \scalebox{0.8}{$ \vg(\gparam  ^{\text{(cur)}})=\begin{bmatrix}    \vg_1  & \vg_{2}  \\ \vg^T_{2} & 0 \end{bmatrix}$} is a (symmetric) Euclidean gradient w.r.t. $\gparam \in \mathcal{M}$.
The vector-Jacobian product in Eq.~\eqref{eq:jacobian_transform} is easy to compute by using coordinate Eq.~\eqref{eq:gnormal_coord2}. Note that the computation of this product does depend on the choice of GNCs.

Our update can be used for SPD estimation such as metric learning, trace
regression, and dictionary learning from a deterministic matrix manifold optimization perspective. Neither Bayesian nor Gaussian assumptions are required. 

In  Gaussian cases, \citet{calvo1990distance} show that a $d$-dimensional Gaussian $\gauss(\vlat|\vmu,\vSigma)$ with mean $\vmu$ and covariance $\vSigma$ can be reexpressed as an augmented $\!(\!d\!+\!1\!)\!$-dimensional Gaussian $\gauss(\vx|\mathbf{0},\gparam)$ with zero mean and covariance $\gparam$, where $\vx^T=[\vlat^T, 1]$ and $\gparam$ is on this SPD submanifold.  
\citet{hosseini2015matrix} consider this reparametrization for maximum likelihood estimation (MLE) of Gaussian mixture models, but their method is only guaranteed to converge to this particular submanifold at the optimum as they use the Riemannian maps for the corresponding full SPD manifold. On the contrary, our update not only stays on this SPD submanifold at each iteration but is also applicable to other SPD submanifolds. 
Our approach expands the scope of structured NGD originally proposed as a Bayesian estimator to a maximum likelihood estimator for Gaussian mixtures with \emph{structured covariances} $\vSigma$ where existing methods such as \citet{hosseini2015matrix} and the expectation-maximization algorithm cannot be applied. 

To  show that structured NGD is a special case of our update, consider a dense covariance $\vSigma$  for simplicity. We can  extend the following results for
structured covariance cases.
In a dense case, the  NGD update \citep{lin2021tractable} for Gaussian $\gauss(\vmu,\vSigma)$ with the Fisher-Rao metric is
\vspace{-0.1cm}

\hspace{1mm} \resizebox{0.96\linewidth}{!}{
  \begin{minipage}{\linewidth}
\begin{align}
  \vmu  \leftarrow   \vmu -\gamma \vSigma \vg_\mu, \,\,\,\,
  \vL \leftarrow \vL \MExp(-\gamma \vL^T \vg_\Sigma \vL), %
  \label{eq:sngd}
\end{align} \end{minipage}
}
\vspace{-0.1cm}

where $\vSigma=\vL\vL^T $ and $\gamma$ is the stepsize for structured NGD. %

As shown in Appx.~\ref{apd:ngd_gaussian_identites}, 
we can reexpress $\vg_1$ and $\vg_2$ using Gaussian gradients $\vg_\mu$ and $\vg_\Sigma$ as
$\vg_1=\vg_\Sigma$ and $\vg_2=\half (\vg_\mu - 2  \vg_\Sigma\,  \vmu)$. Thus, we reexpress $ \vg(\lparam_0)$ as 

\vspace{-0.06cm}
\hspace{1.6mm} \resizebox{0.94\linewidth}{!}{
  \begin{minipage}{\linewidth}
\begin{align}
 \vg(\lparam_0)= \{   \vL^T \vg_\Sigma  \vL, \, \frac{1}{\sqrt{2}} \vL^T \vg_\mu \}.
 \label{eq:gauss_case_update}
\end{align}\end{minipage}
}

\vspace{-0.1cm}

Since the affine-invariant metric is twice the Fisher-Rao metric (see Eq.~\eqref{eq:affinemetric}) , we set stepsize $\stepsize= 2 \gamma $ and momentum weight $\alpha=0$.   
Our update (without momentum) in Eq.~\eqref{eq:local_rgd_mom}  recovers structured NGD in Eq.~\eqref{eq:sngd} by using the Gaussian gradients in Eq.~\eqref{eq:gauss_case_update} and coordinate \eqref{eq:gnormal_coord2}.

Using the GNC in  \eqref{eq:gnormal_coord2}, our update essentially performs NGD in the expectation parameter space \scalebox{0.8}{$\{\vmu, \vSigma+\vmu\vmu^T \}$} (induced by $\gparam$) of a Gaussian by considering the Gaussian as an exponential family. Likewise, using another GNC such as \scalebox{0.8}{$\gparam =\vB^{-T}\vB^{-1}$}, our update performs NGD in the natural parameter space \scalebox{0.8}{$\{-\vSigma^{-1}\vmu , \vSigma^{-1} \}$} (induced by $\gparam^{-1}$), which is the (Bregman) dual space of the expectation parameter space \citep{khan2017conjugate}. 
When \scalebox{0.8}{$\vA$} and \scalebox{0.8}{$\vB^{-T}$} have the same (sparse) group structure even for a  Gaussian with structured covariance $\vSigma$, 
our updates using each of the GNCs agree to first-order \scalebox{0.8} {$O(\stepsize)$} w.r.t. stepsize $\stepsize$ in the global coordinate $\gparam$.

\vspace{-0.1cm}
\subsubsection{Rank-one SPD Submanifold}
\label{sec:low_rank_spd}

Now, we give an example of sparse SPD matrices to illustrate the usage of our update in high-dimensional problems.

Consider the following SPD submanifold, where $k=d+1$.

\vspace{-0.05cm}
\hspace{2.8mm} \resizebox{0.9\linewidth}{!}{
  \begin{minipage}{\linewidth}
 \begin{align*} \quad 
\mathcal{M}\!=\!\Big\{ \gparam\!=\! \begin{bmatrix}a^2 & 
a \vb^T   \\  a \vb &  \vb\vb^T +\mathrm{Diag}(\vc^2)   \end{bmatrix}
\!\in\! \real^{k \times k} \mid\gparam \succ 0 \Big\}.
\end{align*} \end{minipage}
}

To construct GNCs, we reexpress $\mathcal{M}$ as

\vspace{-0.05cm}
\hspace{3mm} \resizebox{0.9\linewidth}{!}{
  \begin{minipage}{\linewidth}
\begin{align*} \quad \,
\mathcal{M}=\Big\{ \gparam = \vA \vA^T   \mid \vA:= \begin{bmatrix} a & \mathbf{0} \\   \vb & \mathrm{Diag}(\vc)  \end{bmatrix},  a>0, \vc>0  \Big\} ,
\end{align*} \end{minipage} } 

\vspace{-0.04cm}
where $\mathrm{Diag}(\vc)$ is in the diagonal subgroup of $\mathrm{GL}^{d \times d}$. 
 Observe that $\vA$ is in a subgroup of $\mathrm{GL}^{k \times k}$.
Thus, we can construct the following GNC,

\resizebox{0.9\linewidth}{!}{
  \begin{minipage}{\linewidth}
\begin{align*}
 \gparam=\vA \vA^T ;\,\,\,   \vA=\vA^{\text{(cur)}}\MExp({ \color{red} \vD } \odot \lparam) ,
\end{align*}\end{minipage}
}

where  \scalebox{0.8}{$\lparam\!=\!\begin{bmatrix} \slparam_a & \mathbf{0} \\ \lparam_b & \mathrm{Diag}(\lparam_c) \end{bmatrix}\!\in\!\real^{k \times k} $}, $\odot$ is elementwise product, and the constant matrix  \scalebox{0.8}{$\vD\!=\!\half \! \! \begin{bmatrix}  1 &  \mathbf{0}  \\ \sqrt{2} \mathbf{1}  & \vI_d \end{bmatrix}$} (with $\mathbf{1}$ denoting a vector of ones) enforces Assumption~2.

The Euclidean gradient $\vg(\lparam_0)$  in Eq.~\eqref{eq:local_rgd_mom} can be efficiently computed via sparse Euclidean gradients w.r.t. $\vA$.
It can also be computed via dense Euclidean gradients w.r.t. $\gparam$ as

\hspace{0.5mm} \resizebox{0.99\linewidth}{!}{
  \begin{minipage}{\linewidth}
\begin{align*}
\vg(\lparam_0) =\begin{bmatrix}  a(a g_1+2 \vg_2^T\vb) + \vb^T \vf & \mathbf{0} \\ \sqrt{2} \vc \odot (a \vg_2+ \vf)  & \mathrm{Diag}( \vc^2 \odot \mathrm{diag}( \vG_3 ) )   \end{bmatrix}, 
\end{align*} \end{minipage}
}

where $\vf=\vG_3 \vb$ and  \scalebox{0.8}{ $ \vg(\gparam  ^{\text{(cur)}})=\begin{bmatrix}    g_1  & \vg^T_{2}  \\ \vg_{2} & \vG_3 \end{bmatrix}$} is a (symmetric) Euclidean gradient w.r.t. $\gparam$. We assume that we can directly compute $\vf$ and $\mathrm{diag}( \vG_3 )$ without computing $\vG_3$. The computation of Eq.~\eqref{eq:jacobian_transform} can be found in
Appx.~\ref{apd:jac_vec_prod}.

On this submanifold,
it is easy to scale up our update (\ref{eq:local_rgd_mom}) to high-dimensional cases, by truncating the matrix exponential function as $\MExp(\vN)\approx \vI+\vN+\half\vN^2$. This truncation preserves the structure and non-singularity of $\vA$.  

\vspace{-0.2cm}
\section{Generalization for Deep Learning}
\label{sec:opt_dl}
\vspace{-0.2cm}
\begin{figure*}[!t]
\center

	\fbox{
			\begin{minipage}{.48\textwidth}
             \textbf{Our Inverse-free, Multiplication-only Update}
           
		\begin{algorithmic}[1]
               \STATE 
            \footnotesize   Each $T$ iter., update   \scalebox{0.9}{ $\vm_K$, $\vm_C$, $\vK$, $\vC$}   \\
                Obtain  $\vmu_{AA} \otimes \vmu_{GG}$ to approximate $\nabla_\mu^2 \ell(\vmu)$ \\  
             \scalebox{0.9}{   $\vm_K  \leftarrow\alpha_1 \vm_K + \frac{\stepsize_1}{2d}(\mathrm{Tr}(\vH_C) \vH_K +  c^2 \vK^T\vK- d\vI_p )$} \\
             \scalebox{0.9}{  $\vm_C  \leftarrow\alpha_1 \vm_C + \frac{\stepsize_1}{2p}(\mathrm{Tr}(\vH_K) \vH_C +  \kappa^2 \vC^T\vC- p\vI_d )$} \\
             \scalebox{0.9}{  $\vK \leftarrow \vK \MExp(-\vm_K) \approx \vK (\vI_p-\vm_K)$} \\
            \scalebox{0.9}{  $\vC \leftarrow \vC \MExp(-\vm_C) \approx \vC (\vI_d-\vm_C)$} \\   
        
            \STATE
          \scalebox{0.8}{    $\vM_\mu  \leftarrow \alpha_2 \vM_\mu  + \vC\vC^{T} \mathrm{vec}^{-1}( \nabla_\mu \ell(\vmu)) \vK\vK^{T} + \gamma  \mathrm{vec}^{-1}(  \vmu )   $}
             \STATE
            \scalebox{0.9}{   $ \vmu \leftarrow \vmu - \stepsize_2    \mathrm{vec}(\vM_\mu )   $}
				\end{algorithmic}
	\end{minipage}
	}
   \fbox{
			\begin{minipage}{.46\textwidth}
		\textbf{KFAC Optimizer}
   
				\begin{algorithmic}[1]
               \STATE 
            \footnotesize   Each $T$ iter., update  \scalebox{0.78}{ $\left(\mathbf{KK^T}\right)^{-1}$, $\left(\mathbf{CC^T}\right)^{-1}$, $\mathbf{KK^T}$, $\mathbf{CC^T}$ }  \\
                Obtain $\vmu_{AA} \otimes \vmu_{GG}$ to approximate $\nabla_\mu^2 \ell(\vmu)$ \\
          \scalebox{0.8}{       $\left(\mathbf{KK^T}\right)^{-1} \leftarrow  \theta \left(\mathbf{KK^T}\right)^{-1} + (1-\theta)  \vmu_{AA}$} \\
              \scalebox{0.8}{    $\left(\mathbf{CC^T}\right)^{-1}  \leftarrow \theta \left(\mathbf{CC^T}\right)^{-1}  + (1-\theta)   \vmu_{GG}$} \\
            \scalebox{0.8}{     $\mathbf{KK^T} \leftarrow (\left(\mathbf{KK^T}\right)^{-1}+ \lambda \vI_p )^{-1}$ } \\
              \scalebox{0.8}{    $\mathbf{CC^T}  \leftarrow (\left(\mathbf{CC^T}\right)^{-1}+ \lambda \vI_d )^{-1}$ }
            \STATE
          \scalebox{0.8}{    $\vM_\mu  \leftarrow \alpha_2 \vM_\mu  +  \mathbf{CC^T} \mathrm{vec}^{-1}( \nabla_\mu \ell(\vmu)) \mathbf{KK^T}   + \gamma   \mathrm{vec}^{-1}( \vmu ) $}
             \STATE
           \scalebox{0.9}{    $ \vmu \leftarrow \vmu - \stepsize_2    \mathrm{vec}(\vM_\mu )  $}

				\end{algorithmic}
	\end{minipage}
	}
 \vspace{-0.1cm}   \caption{ 
 In our update, we denote  \scalebox{0.9}{ $\vH_K := \vK^T \vmu_{AA} \vK\,$, \  \ $\vH_C  := \vC^T \vmu_{GG} \vC\,$,  \ \  $\kappa^2  := \lambda\mathrm{Tr}(\vK^T\vK)\,$,} and \scalebox{0.9}{$c^2  := \lambda\mathrm{Tr}(\vC^T\vC)$}, where \scalebox{0.9}{$\mathrm{vec}^{-1}(\vmu) \in \real^{d \times p}$,
 $\vC \in \real^{d \times d}$, $\vK \in \real^{p \times p}$}.
Note that we merge factors $\frac{1}{2\sqrt{d}}$ and $\frac{1}{2\sqrt{p}}$ in Eq.~\eqref{eq:mat_gauss_norm_coord} into the updates in $\vm_K$ and $\vm_C$, respectively (see Eq.~\eqref{eq:matgauss_just} in Appx.~\ref{apd:mat_gauss_dl} for a justification).
  We use the linear truncation of the matrix exponential function.  Our update does not require
explicit matrix inverses.  We can also pre-compute $\vC\vC^T$ and $\vK\vK^T$ when $\vT>1$.
In KFAC, a damping term $\lambda \vI$ is introduced to handle the singularity of \scalebox{0.9}{$ \left(\mathbf{KK^T}\right)^{-1}$} and \scalebox{0.9}{$\left(\mathbf{CC^T}\right)^{-1}$}. We introduce a similar damping term in $\kappa^2$ and $c^2$ (see  Appx.~\ref{apd:mat_gauss_dl} for a derivation) to improve numerical stability. 
Our update and KFAC include momentum weight $\alpha_2$ for layer-wise NN weights $\vmu$ and (L2) weight decay $\gamma$. In our update, we also introduce momentum weight $\alpha_1$ in the SPD preconditioner. Our update is more numerically robust than KFAC. Thus, our update can often use a larger stepsize $ \stepsize_2$ and a smaller damping weight $\lambda$ than KFAC.
\vspace{-0.22cm}
}
\label{fig:matDL_opt}
\end{figure*}

It is useful to design preconditioned GD such as Newton's method by exploiting the submanifold structure of the preconditioner.
We will use that structure to design inverse-free structured matrix optimizers for low-precision floating-point training schemes in DL. 

To solve a minimization problem $\min_{\mu \in \scaleto{\real^k}{6pt}}   \ell(\vmu) $, a Newton's step with $\stepsize=1$ is

\vspace{-0.1cm}
 
\resizebox{\linewidth}{!}{
  \begin{minipage}{1.1\linewidth}
\begin{align*}
\vS \leftarrow (1-\stepsize) \vS + \stepsize \nabla_\mu^2 \ell(\vmu), \,\,\,
\vmu\! \leftarrow\! \vmu - \stepsize \vS^{-1} \vg_\mu
\end{align*}\end{minipage}
}

\vspace{-0.1cm}
In stochastic settings, it is useful to consider an averaged update by setting $0<\stepsize<1$. However, the updated $\vS$ can be non-SPD.

\citet{lin2021snd} propose a Newton-like update derived from structured NGD.
Instead of directly updating $\vS$, the authors consider a SPD preconditioner $\vS:=\vB\vB^T$ and update $\vB$ as follows.
 
\resizebox{\linewidth}{!}{
  \begin{minipage}{1.1\linewidth}
\begin{align}
\vmu\! \leftarrow\! \vmu - \stepsize \vS^{-1} \vg_\mu,\,\,\,
\vB\!  \leftarrow\! \vB \MExp\big(  \frac{\stepsize}{2} \vB^{-1} \vg_{S^{-1}}   \vB^{-T}  \big),\label{eq:sngd_dl}
\end{align} \end{minipage}
}

where  $\vg_{S^{-1}} :=\half (\nabla_\mu^2 \ell(\vmu)-\vS) $, and $\vg_\mu:= \nabla_\mu \ell(\vmu)$. \citet{lin2021snd} show that the update in $\vB$ can be re-expressed in terms of $\vS$ as $\vS \leftarrow (1-\stepsize) \vS + \stepsize \nabla_\mu^2 \ell(\vmu) + O(\stepsize^2)$. More importantly, the updated $\vS$ is guaranteed to be SPD.

Similar to Newton's update, Eq.~\eqref{eq:sngd_dl} is linearly (Lie group) invariant in $\vmu$. When $\vS$ is a SPD submanifold, this update has a structural (Lie subgroup) invariance\footnote{This is a Lie-group structural invariance in $\vmu$ induced by a structured SPD preconditioner \citep{lin2021snd}.}. 
However, this update requires a matrix inverse which can be slow and numerically unstable in large-scale training due to the use of low-precision floating-point training schemes.

Eq.~\eqref{eq:sngd_dl} can be obtained by considering the inverse of the preconditioner \scalebox{0.9}{$\gparam=\vS^{-1}=\vB^{-T}\vB^{-1}$} as a SPD manifold. The update of $\vB$ can be obtained by using the GNC in Sec.~\ref{sec:demystifying} as \scalebox{0.9}{$\gparam= \vB^{-T} \vB^{-1}$} where \scalebox{0.9}{$\vB = \vB^{\text{(cur)}} \MExp(-\half \lparam)$}.  The minus sign (see Eq.~\eqref{eq:update_coordinate_b} in Appx.~\ref{apd:simple_update_spd}) is canceled out by another minus sign in the GD update of Eq.~\eqref{eq:local_rgd_mom}.

We can obtain a matrix-inverse-free update (see  Appx.~\ref{apd:simple_update_spd}):

\vspace{-0.1cm}
\resizebox{0.99\linewidth}{!}{
  \begin{minipage}{\linewidth}
\begin{align}
  \vmu \leftarrow \vmu -  \stepsize  \vA \vA^{T} \vg_\mu,
  \label{eq:no_mat_inv_opt}
\end{align}\end{minipage}
}

\vspace{-0.1cm}
 by changing the GNC to $\vS^{-1}=\gparam= \vA \vA^{T}$ where $\vA =\vA^{\text{(cur)}} \MExp(\half \lparam)$. Moreover, the update of $\vA$ is also matrix-inverse-free (see Sec.~\ref{sec:demystifying}
 and 
 Eq.~\eqref{eq:update_coordinate_a} in Appx.~\ref{apd:simple_update_spd}
 ). %
 
As shown in  Sec.~\ref{sec:sngd_special},
we can obtain the  update in \eqref{eq:no_mat_inv_opt} by considering $\{\vmu, \vSigma\}$ as a submanifold, where $\vSigma=\vS^{-1}$. Importantly, this implies that our Newton-like (2nd-order) update in the space of $\vmu$ can be viewed as  a gradient descent (1st-order) update in our (local) GNC spaces for the  SPD submanifold.

To approximate the Hessian $\nabla_\mu^2 \ell(\vmu)$ in DL as required in $\vg_{S^{-1}}$,
we consider the KFAC approximation \citep{martens2015optimizing}.
For simplicity, consider  the loss function $\ell(\vmu)$ defined by one hidden layer of a neural network (NN),  where $k=pd$, $\mathrm{vec}^{-1}(\vmu) \in \real^{d \times p}$ is a learnable weight matrix, $\mathrm{vec}(\cdot)$ is the  vectorization function.
In KFAC (summarized in Fig.~\ref{fig:matDL_opt}), the Hessian at each layer of a  NN with many layers is approximated by a Kronecker-product structure between two dense symmetric positive semi-definite matrices as $\nabla_\mu^2  \ell(\vmu)\approx \vmu_{AA} \otimes \vmu_{GG}   $, where matrices $\vmu_{AA} \in \real^{p \times p}$ and $\vmu_{GG}  \in \real^{d \times d}$ are computed as suggested by the authors and 
$\otimes$ denotes the Kronecker product. 

We consider a Kronecker-product structured submanifold with $k=pd$ to exploit the structure of the approximation: %

\vspace{-0.1cm}
\resizebox{0.88\linewidth}{!}{
  \begin{minipage}{\linewidth}
\begin{align*}
 \mathcal{M}=\Big\{ \gparam =  \vU \otimes \vW 
\in \real^{pd \times pd} \mid\gparam \succ 0 \Big\},
\end{align*} \end{minipage}
}

\vspace{-0.15cm}
where  
matrices $\vU \in \real^{p \times p }$ and $\vW \in \real^{d \times d }$ are both SPD. We can reexpress this submanifold as

\vspace{-0.1cm}
\hspace{1.4mm} \resizebox{0.98\linewidth}{!}{
  \begin{minipage}{\linewidth}
\begin{align}
  \mathcal{M}\!=\!\Big\{ \gparam\!=\!\vA \vA^T   \mid \vA\!:=\!  \vK \otimes \vC    \Big\} \label{eq:mat_sub_manifold},
 \end{align}  \end{minipage}
}

\vspace{-0.15cm}
where $\vU=\vK\vK^T$, $\vW=\vC\vC^T$, $\vK  \in   \real^{p \times p}$,   $\vC \in \real^{d \times d}$. $\vK$ and $\vC$ are (sub)groups of $\mathrm{GL}^{p \times p}$ and $\mathrm{GL}^{d \times d}$, respectively. 

By exploiting the Kronecker structure in $\vA=\vK \otimes \vC$,  we can reexpress the update of  $\vmu$ in Eq.~\eqref{eq:no_mat_inv_opt} as: 

\resizebox{\linewidth}{!}{
  \begin{minipage}{\linewidth}
\begin{align}
 \vmu  \leftarrow  \vmu  -  \stepsize \mathrm{vec}\left( \vC\vC^T \mathrm{vec}^{-1}( \vg_\mu ) \vK\vK^T \right).
\end{align} \end{minipage}
 }
\vspace{-0.4cm}

Now, we describe how to update $\vA=\vK \otimes \vC$ by using the structure of the SPD submanifold $\gparam=\vA\vA^T$. Unfortunately, the affine-invariant metric defined in Eq.~\eqref{eq:affinemetric} is singular. 
Note that the metric used in a standard product manifold is different from the metric for a Kronecker-product submanifold.
To enable the usage of this submanifold, we consider a block-diagonal approximation to update blocks $\vK$ and $\vC$. Given each block, we construct a normal coordinate to orthonormalize the metric with respect to the block while keeping the other block frozen. Such an approximation of the affine-invariant metric leads to a block-diagonal approximated metric.    
For blocks $\vK$ and  $\vC$, we consider the following blockwise GNCs $\lparam_K$ and $\lparam_C$, respectively:

\vspace{-0.15cm}
\resizebox{0.8\linewidth}{!}{
  \begin{minipage}{\linewidth}
\begin{align}
\vA \! =\!   \left( \!\vK^{\text{(cur)}} \MExp\left( \!   \frac{\lparam_K}{{ \color{red} 2 \sqrt{d}} }  \!  \right) \!\right) \!\otimes \! \vC^{\text{(cur)}}   , \nonumber \\
\vA  \!=\!  \vK^{\text{(cur)}} \!\otimes \! \left( \! \vC^{\text{(cur)}} \MExp\left( \!   \frac{\lparam_C}{{\color{red}2 \sqrt{p}} }  \!  \right) \!\right),  \label{eq:mat_gauss_norm_coord}
 \end{align} \end{minipage}
 }

\vspace{-0.07cm}
 where both  $\lparam_K \in \real^{p \times p }$ and $\lparam_C \in \real^{d \times d}$ are symmetric matrices and   $\vA^{\text{(cur)}}=\vK^{\text{(cur)}} \otimes\vC^{\text{(cur)}} $. The scalars highlighted in red are needed to orthonormalize the block-diagonal metric. 

Using these blockwise GNCs, 
our update is  summarized in Fig.~\ref{fig:matDL_opt} (see   Appx.~\ref{apd:mat_gauss_dl} for a derivation), where similar to KFAC, we use individual stepsizes to update $\vmu$ and $\vA$, and further 
introduce momentum weight $\alpha_2$ for $\vmu$, weight decay $\gamma$, and damping weight $\lambda$. In practice, we truncate the matrix exponential function: the quadratic truncation $\MExp(\vN) \approx \vI+\vN + \half \vN^2$ ensures the non-singularity of $\vK$ and $\vC$. In our DL experiments, we observed that the linear truncation $\MExp(\vN) \approx \vI+\vN$ also works well. 

We can also develop sparse Kronecker updates while original KFAC does not admit a sparse update. For example, our approach allows us to include sparse structures by considering $\vK$ and $\vC$ with sparse group structures in Eq.~\eqref{eq:mat_sub_manifold}.

Our approach allows us to go beyond the KFAC approximation by 
exploiting other structures in the Hessian approximation of $\nabla_\mu^2 \ell(\vmu)$, and  to develop inverse-free update schemes by using SPD submanifolds to approximate the structures.

\vspace{-0.25cm}
\section{Numerical Results}
\label{sec:results}
\begin{figure*}[t!]
\vspace{-0.1cm}
\captionsetup[subfigure]{aboveskip=-1pt,belowskip=-1pt}
        \centering
\hspace*{-1.0cm}
        \begin{subfigure}[b]{0.25\textwidth}
	\includegraphics[width=\textwidth]{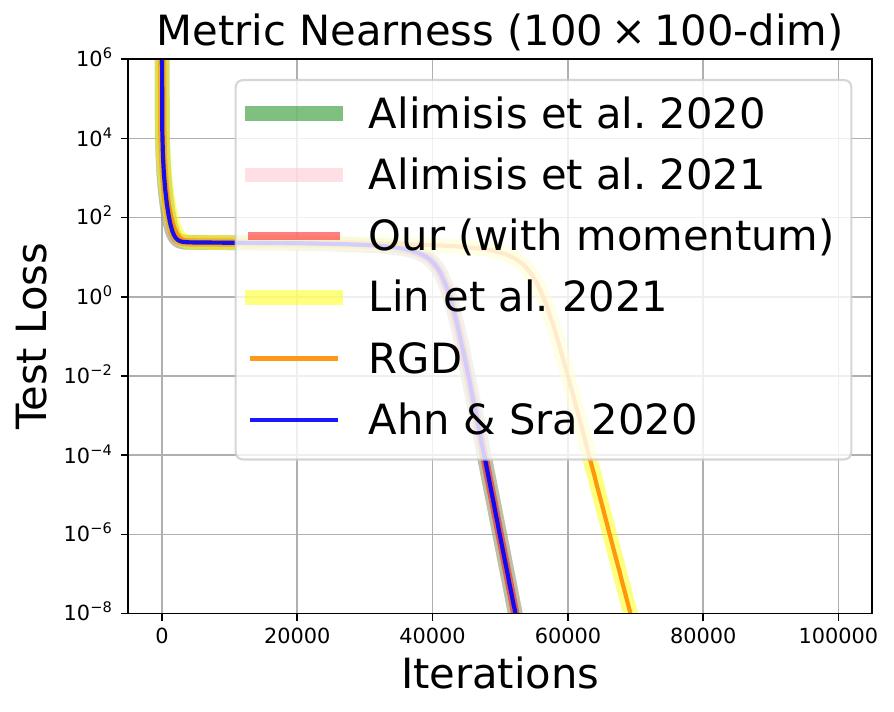}
                \caption{}
	      \label{fig:a}
        \end{subfigure}       
	\hspace*{-0.2cm}
        \begin{subfigure}[b]{0.25\textwidth}
	\includegraphics[width=\textwidth]{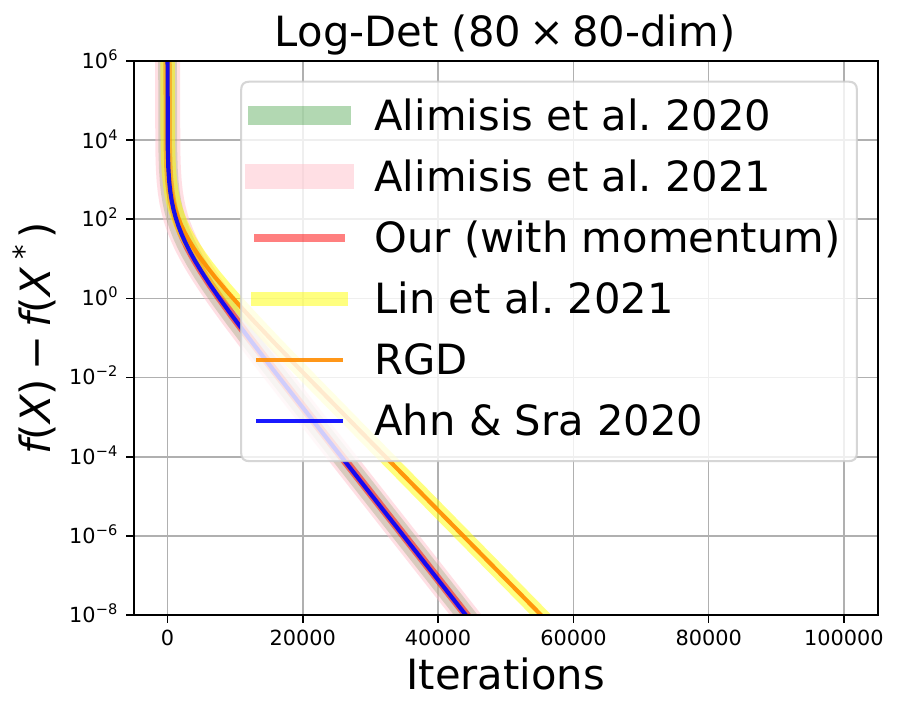}
                \caption{}
	      \label{fig:b}
        \end{subfigure}
	\hspace*{-0.18cm}
         \begin{subfigure}[b]{0.25\textwidth}
	\includegraphics[width=\textwidth]{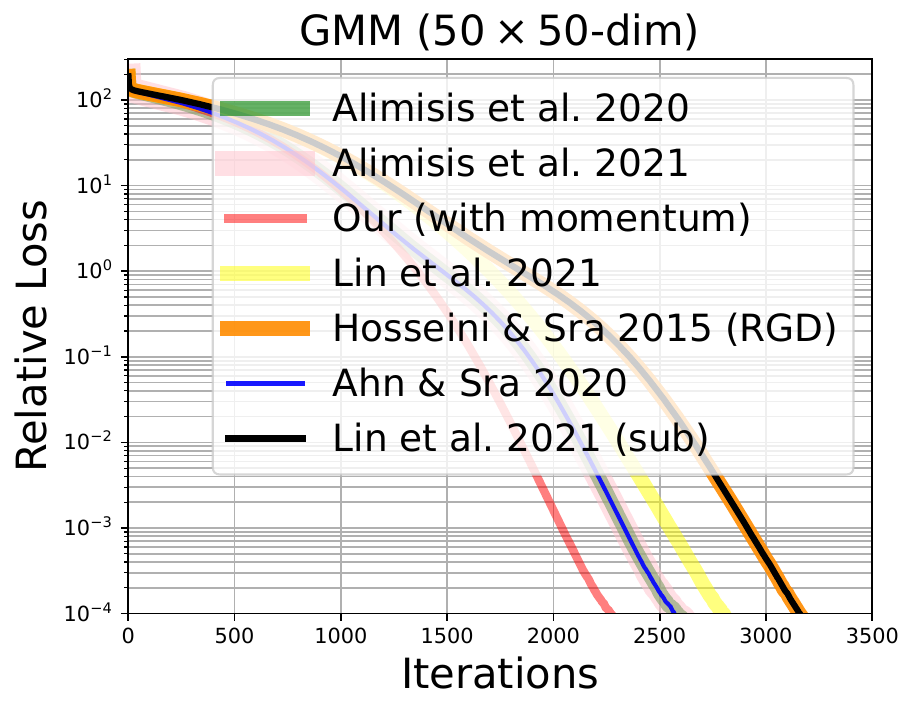}
                \caption{}
	      \label{fig:c}
        \end{subfigure}
	\hspace*{-0.18cm}
        \begin{subfigure}[b]{0.25\textwidth}
	\includegraphics[width=\textwidth]{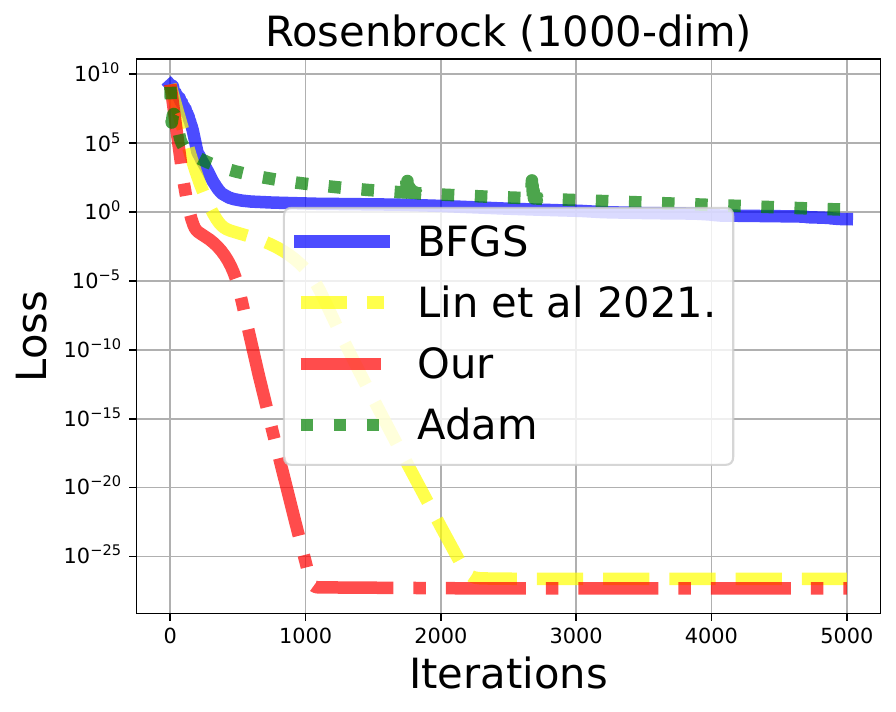}
                \caption{}
	      \label{fig:d}
        \end{subfigure}       
     \vspace{-0.45cm}   
\hspace*{-1.2cm}
          \caption{ 
The performance of our updates for  optimization problems.
Fig.~\ref{fig:a}-\ref{fig:b} show the performance on  SPD manifold optimization problems. Our update using approximations of the Riemannian maps achieves a similar performance as existing Riemannian methods using the exact Riemannian maps.
Fig.~\ref{fig:c} shows the performance on
a MLE problem on a Gaussian mixture.
The method denoted by ``sub" performs updates on a SPD submanifold (see sec.~\ref{sec:sngd_special}) while the other methods perform updates on a SPD manifold.
Note that the loss in Fig.~\ref{fig:c} is computed by augmented $(\!d\!+\!1\!)$-dim  Gaussian components suggested by \citet{hosseini2015matrix}. 
If we perform updates on the SPD manifold $\mathcal{S}_{++}^{k \times k}$ with $k\!=\!d+1$ instead of the submanifold, we cannot obtain the original (non-augmented) $d$-dim Gaussian components during the iterations since the updates are not guaranteed to stay on the submanifold. Thus, we cannot use the standard MLE loss defined by the $d$-dim Gaussians. %
In Fig.~\ref{fig:a}-\ref{fig:c}, we use the same stepsize and momentum weight for all methods. 
Note that our method and \citet{lin2021tractable} can use a larger stepsize than the other methods using 
the exact Riemannian maps. Our method and \citet{lin2021tractable} use 
the quadratic truncation while the other methods use the exact maps.
We observe that our method with truncation is more numerically robust than the other methods using the exact maps.
Fig.~\ref{fig:d} shows the performance using
a structured preconditioner to optimize a $1000$-dim function, where our update and structured NGD use Hessian information without computing the full Hessian. 
 }
\end{figure*}

\begin{figure*}[!ht]
\vspace{-0.1cm}
\captionsetup[subfigure]{aboveskip=-1pt,belowskip=-1pt}
        \centering
\hspace*{-1.0cm}
        \begin{subfigure}[b]{0.25\textwidth}
	\includegraphics[width=\textwidth]{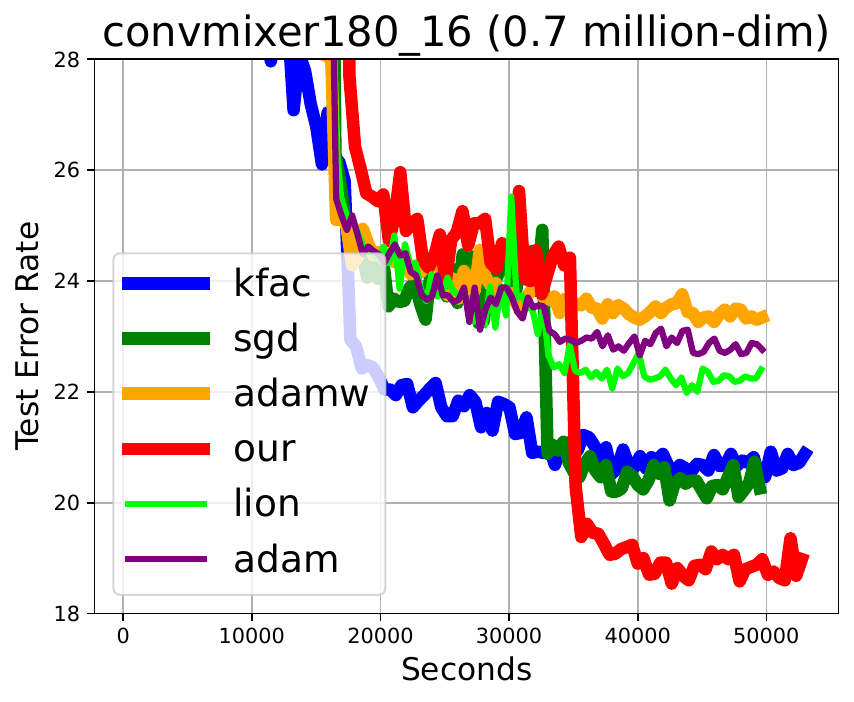}
        \end{subfigure}       
	\hspace*{-0.2cm}
        \begin{subfigure}[b]{0.25\textwidth}
	\includegraphics[width=\textwidth]{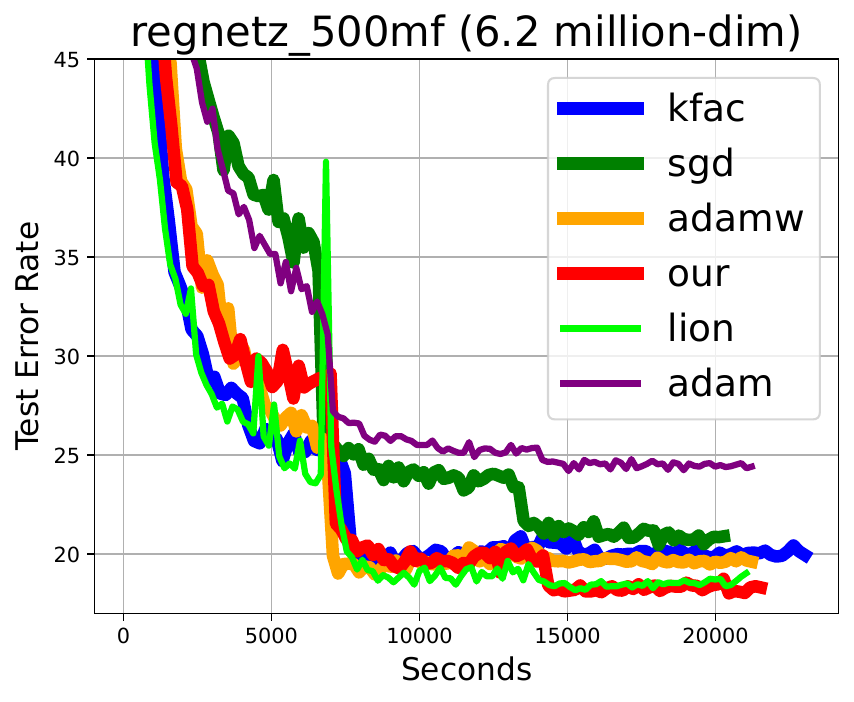}
        \end{subfigure}
	\hspace*{-0.15cm}
         \begin{subfigure}[b]{0.25\textwidth}
	\includegraphics[width=\textwidth]{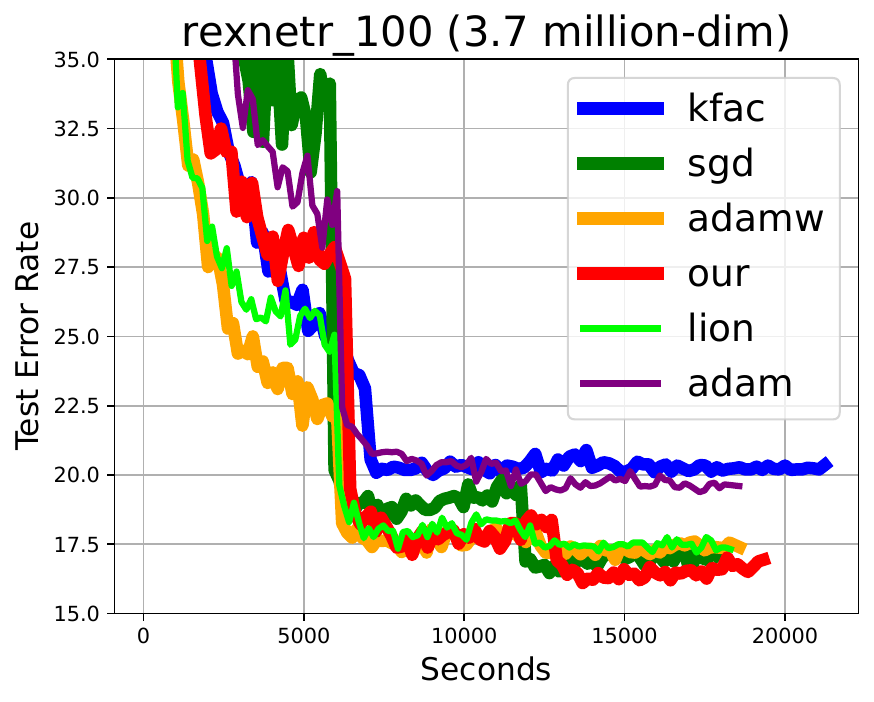}
        \end{subfigure}
	\hspace*{-0.15cm}
        \begin{subfigure}[b]{0.25\textwidth}
	\includegraphics[width=\textwidth]{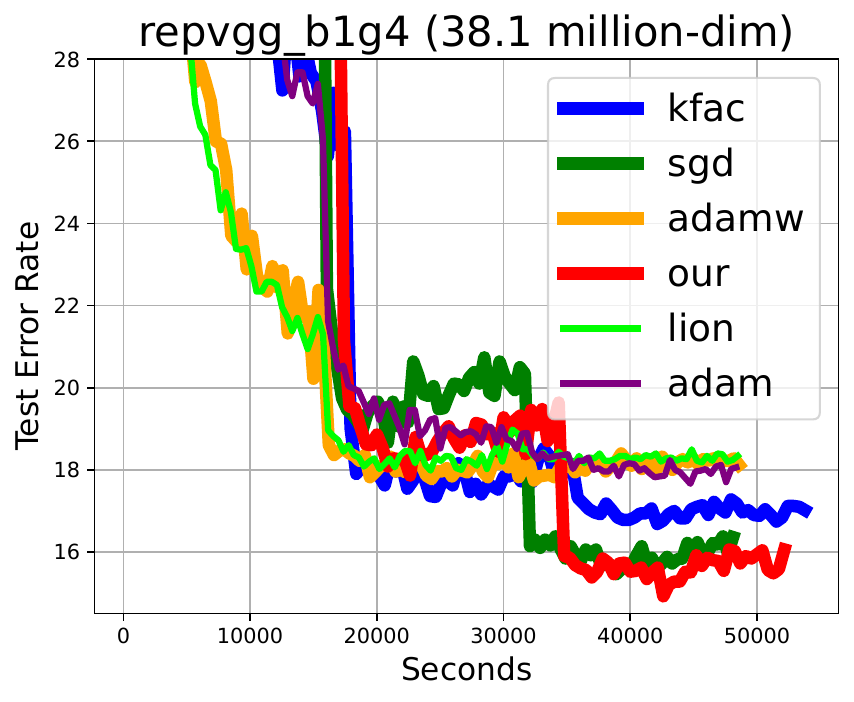}
        \end{subfigure}        
     \vspace{-0.45cm}   
\hspace*{-1.2cm}
\caption{ 
The error curves for optimization in deep NN models on the ``ImageNet-100" dataset. 
Our updates achieve lower test error rates than the other baseline methods for NN optimization.  
We report the number of learnable NN weights, $k$, in a round bracket shown in the title of each plot. For each NN optimization problem, our approach uses a structured and sparse $k$-by-$k$ SPD preconditioning matrix induced by a SPD submanifold.
As shown in Table~\ref{tab:spd_table}, it is computationally infeasible to use 
many Riemannian momentum methods since they are designed for (dense) SPD preconditioning matrices  and have $O(k^3)$ time complexity.
\vspace{-0.3cm}
 }
\label{fig:dnn_imagenet100}
\end{figure*}

\vspace{-0.2cm}
\subsection{Results on Synthetic Examples}
\vspace{-0.2cm}

To validate our proposed updates, we consider several optimization problems.
In the first three examples, we consider manifold optimization on SPD matrices $\mathcal{S}_{++}^{k \times k}$, where the Riemannian maps admit a closed-form expression (shown in Table~\ref{tab:spd_table}). 
We evaluate our method on the metric nearness problem considered in \citet{lin2021tractable}, a log-det optimization problem considered in \citet{han2021riemannian}, and a MLE problem of a Gaussian mixture model (GMM) considered in \citet{hosseini2015matrix}. 
We consider structured NGD \citep{lin2021tractable}, RGD, and existing Riemannian momentum methods such as the ones presented by
\citet{ahn2020nesterov}, \citet{alimisis2020continuous,alimisis2021momentum}  as baselines (see Appx.~\ref{apd:imple_baselines} for our  implementation of these methods). 
All methods are trained using the same stepsize and momentum weight.  Our updates and structured NGD can use a larger stepsize than the other methods. The exact Riemannian maps 
are not numerically stable in high-dimensional settings.    
From Fig.~\ref{fig:a}-\ref{fig:c}, we can see that our method performs as well as the Riemannian methods with the exact Riemannian maps in the global coordinate.
In the last example, we minimize a $1000$-dim Rosenbrock function. We consider the inverse of the preconditioner $\gparam=\vS^{-1}$ in Eq.~\eqref{eq:sngd_dl} as a SPD submanifold. We include momentum in $\gparam$ and $\vmu$. 
We compare our update with structured NGD, where both methods use the Heisenberg structure suggested by \citet{lin2021tractable} to construct a submanifold. Both methods make use of Hessian information without  computing the full Hessian. We consider other baselines: BFGS and Adam. We tune 
the stepsize for all methods. From Fig.~\ref{fig:d}, we can see that adding momentum in the preconditioner could be useful for optimization.

\begin{table*}[ht!]
\begin{minipage}[t]{\linewidth}
\centering
\resizebox{0.8 \textwidth}{!}{   %
 \begin{tabular}{c c c c c c}
		\toprule
		\begin{tabular}{c}
                  Dataset  \\
		\end{tabular} &
		Method & 
		\begin{tabular}{c}
                  VGG-16  \\
		\end{tabular} &
		\begin{tabular}{c}
                  PyramidNet-65  \\
		\end{tabular} &
		\begin{tabular}{c}
			ConvMixer256-12   \\
		\end{tabular}& 
  		\begin{tabular}{c}
			RegNetX-1.6GF   \\
		\end{tabular}   
		\\
                \midrule
		\multirow{3}{*}{
		  \begin{tabular}{c}
                    \\ 
                    CIFAR-100
		\end{tabular}}   
		& SGD    & $26.45$  & $25.65$     & $27.35$    & $\mathbf{21.62}$    \\
                & Adam         & $30.14$    & $28.95$    & $31.20$   &  $28.46$    \\
		& AdamW      & $31.50$   &  $28.61$ &  $30.5$ &  $26.94$ \\
		& Lion           & $31.83$   &  $28.14$ &  $29.10$ &  $25.53$ \\
                & KFAC  & $27.68$ &  $25.93$ &  $26.14$  & $23.11$ \\
  & Ours      & $\mathbf{26.06}$  & $\mathbf{25.39}$ & $\mathbf{25.79}$ &  $21.89$ \\
                \midrule
		\multirow{3}{*}{
		  \begin{tabular}{c}
                    \\ 
                    TinyImageNet-200
		\end{tabular}}   
		& SGD    & $40.18$     &  $41.94$   & $40.03$   &  $\mathbf{34.38}$    \\
                & Adam         & $44.17$   & $45.00$   &  $43.05$    & $43.72$     \\
		& AdamW            & $45.10$  &  $42.72$ & $46.46$ &  $40.26$ \\
		& Lion            & $45.53$  &  $41.04$ & $43.83$ &  $38.31$ \\
 & KFAC  &  $41.59$ & $41.23$ & $42.43$ &  $36.64$ \\
  & Ours   & $\mathbf{40.01}$  & $\mathbf{40.82}$ &  $\mathbf{39.09}$ &   $34.85$ \\
                \bottomrule \\
	\end{tabular}
 } 
\vspace{-0.5cm}
\caption{
More results about
the performance (test error rate) of the methods  considered in error curves on the other datasets (shown in Fig.~\ref{figure:dnn} in Appx.~\ref{app:more_results}). 
The results are
obtained by averaging over the last 10 iterations.   } 
\label{table:results1}  
\end{minipage}
\end{table*}

\begin{table*}[ht]
\begin{minipage}[t]{\linewidth}
\centering
\resizebox{0.8 \textwidth}{!}{   %
 \begin{tabular}{c c c c c c}
		\toprule
		\begin{tabular}{c}
                  Dataset  \\
		\end{tabular} &
		Method & 
		\begin{tabular}{c}
                  ConvMixer180-16  \\
		\end{tabular} &
		\begin{tabular}{c}
                RegNetZ-500MF  \\
		\end{tabular} &
		\begin{tabular}{c}
			  ReXNetr-100  \\
		\end{tabular}& 
  		\begin{tabular}{c}
			 RepVGG-B1G4    \\
		\end{tabular}   
		\\
                \midrule
		\multirow{3}{*}{
		  \begin{tabular}{c}
                    \\ 
                 ImagetNet-100 
		\end{tabular}}   
		& SGD   & $20.37$  & $20.80$     & $17.07$    & $16.15$    \\
                & Adam        & $22.79$    & $24.48$    & $19.58$   &  $17.95$    \\
		& AdamW      & $23.38$   &  $19.65$ &  $17.43$ &  $18.23$ \\
		& Lion        & $22.25$   &  $18.66$ &  $17.43$ &  $18.30$ \\
                & KFAC  & $20.70$ &  $20.07$ &  $20.24$  & $16.97$ \\
  & Ours   & $\mathbf{18.84}$  & $\mathbf{18.30}$ & $\mathbf{16.73}$ &  $\mathbf{15.81}$ \\
                \bottomrule \\
	\end{tabular}
 } 
\vspace{-0.5cm}
\caption{ 
Results about the performance (test error rate) of the methods in Fig.~\ref{fig:dnn_imagenet100}.
The results are obtained by averaging over the last 10 iterations.  \vspace{-0.5cm}  }
\label{table:results2} 
\end{minipage}
\end{table*}

\vspace{-0.2cm}
\subsection{Results in Deep Learning}
\label{sec:results_dl}
 \vspace{-0.2cm}
 
To demonstrate our method
as a practical Riemannian method in high-dimensional cases, we consider image classification tasks with NN architectures ranging from classical to modern models: VGG \citep{simonyan2014very} with the batch normalization, PyramidNet \citep{han2017deep}, RegNetX \citep{radosavovic2020designing}, RegNetZ \citep{dollar2021fast}, RepVGG \citep{ding2021repvgg}, ReXNetr \citep{han2021rethinking}, and ConvMixer \citep{trockman2022patches}. We use the KFAC approximation for convolution layers \citep{grosse2016kronecker}.  Table~\ref{table:nn_models} in Appx.~\ref{app:more_results} summarizes the number of learnable parameters. We consider three complex datasets ``CIFAR-100'',  ``TinyImageNet-200''\footnote{\scriptsize \url{github.com/tjmoon0104/pytorch-tiny-imagenet}}, and ``ImageNet-100''\footnote{\scriptsize \url{kaggle.com/datasets/ambityga/imagenet100}}.
The hyper-parameter configuration of our update and KFAC can be found in Table~\ref{table:hyperparameters_opt} in Appx.~\ref{app:more_results}. We also consider other baselines such as SGD with momentum, Adam,  AdamW, and Lion \citep{chen2023symbolic}. 
A L2 weight decay is included in all methods.
We set the weight decay to be $0.1$ for Lion, $0.001$ for SGD and Adam,  and 
$0.01$ for AdamW, KFAC, and our method.  For AdamW and Lion, we use the weight decay suggested by  \citet{chen2023symbolic}.  We choose the best weight decay over the set $\{0.1,0.01,0.001,0.0001\}$ for SGD and Adam. For KFAC and our method, we use the same weight decay as the one used in AdamW.
We train all models from scratch for 120 epochs with mini-batch size 128.
For all methods, we tune the initial stepsize and then divide the stepsize by 10 every 40 epochs, as suggested by \citet{wilson2017marginal}.
Note that our method can take a larger stepsize and a smaller damping weight than KFAC for all the NN models. Our method has similar running times as KFAC, as shown in Fig.~\ref{fig:dnn_imagenet100} in the main text, and Fig.~\ref{figure:dnn_time} in Appx.~\ref{app:more_results}.
We report the test error rate (i.e., $\text{error rate} = 100 - \text{accuracy percentage}$) for all methods in Tables~\ref{table:results1} and \ref{table:results2}.
From Fig.~\ref{fig:dnn_imagenet100}, we can see that our method performs better than KFAC and achieves competitive performances among other baselines. More results on other datasets can be found in Fig.~\ref{figure:dnn} in Appx.~\ref{app:more_results}.

\vspace{-0.2cm}
\section{Conclusion}
\vspace{-0.2cm}

We propose GNCs to simplify existing Riemannian momentum methods via \emph{metric-preserving} trivializations, which results in practical momentum-based NGD updates with metric-inverse-free Riemannian/natural gradient computation.
We exploit Lie-algebra structures in GNCs of SPD manifolds and use the structures to construct SPD submanifolds so that 
our updates on each of the submanifolds preserve a 
Lie-subgroup structure.
We show that a NGD update on a Gaussian family is a special case of  our update on a higher-dimensional SPD submanifold. 
Our approach further expands the scope of structured NGD to SPD submanifolds arising in applications of ML and enables the usage of structured NGD beyond Bayesian and Gaussian settings from a manifold optimization perspective.
We further develop matrix-inverse-free structured optimizers for deep learning by exploiting the submanifold structure of SPD preconditioners. An interesting application is to design customized optimizers for a given neural network architecture by
investigating a range of submanifolds.
Overall, our work 
provides a new way to design practical manifold optimization methods while taking care of numerical stability in high-dimensional, low-numerical precision, and noisy settings.

%
\section*{Acknowledgements}
\vspace{-0.2cm}
This research was partially supported by the Canada CIFAR AI Chair Program, the NSERC grant RGPIN-2022-03669, the NSF under grants CCF-2112665, DMS-1345013,
DMS-1813635, and the AFOSR under grant FA9550-18-1-0288.

\bibliography{refs}
\bibliographystyle{icml2023}

\clearpage
\newpage
\clearpage
\pagebreak
\onecolumn
\begin{appendices}
Outline of the Appendix:
\begin{itemize}
\item 
Appendix \ref{app:more_results} contains more numerical results.
 \item 
Appendix \ref{app:summary} summarizes the normal coordinates used  in this work.
\item
The rest of the appendix contains proofs of the claims and derivations  for examples considered in the main text.
 \end{itemize}

 \hfill

\section{Additional Results}
 \label{app:more_results}

 \hfill

\begin{figure*}[!htbp]
  \begin{minipage}[t]{\linewidth}
	\centering
	\hspace*{-1.0cm}
 \includegraphics[width=0.25\linewidth]{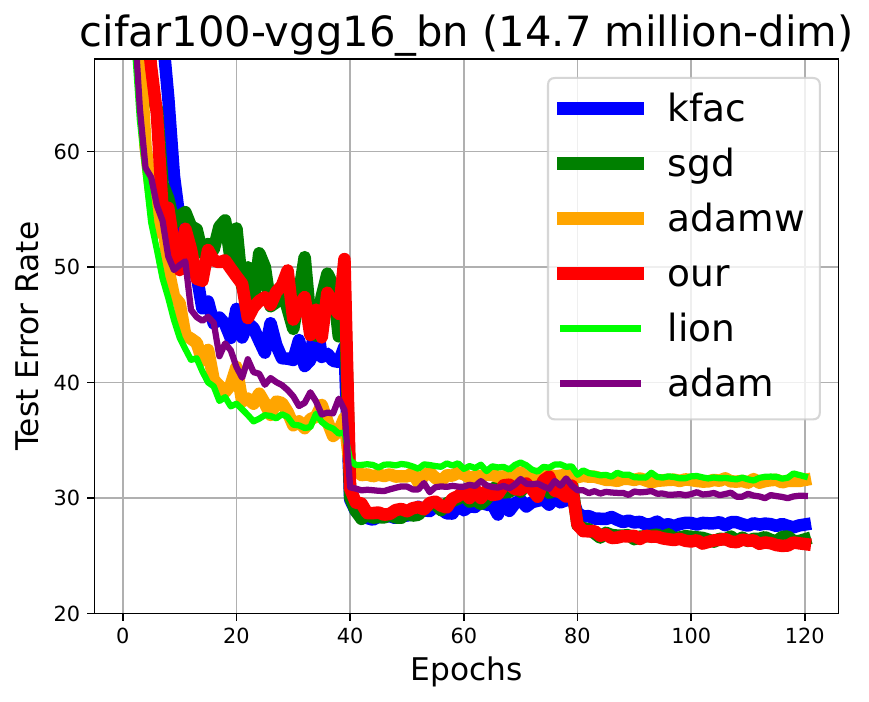}
\hspace*{-0.12cm}		
\includegraphics[width=0.25\linewidth]{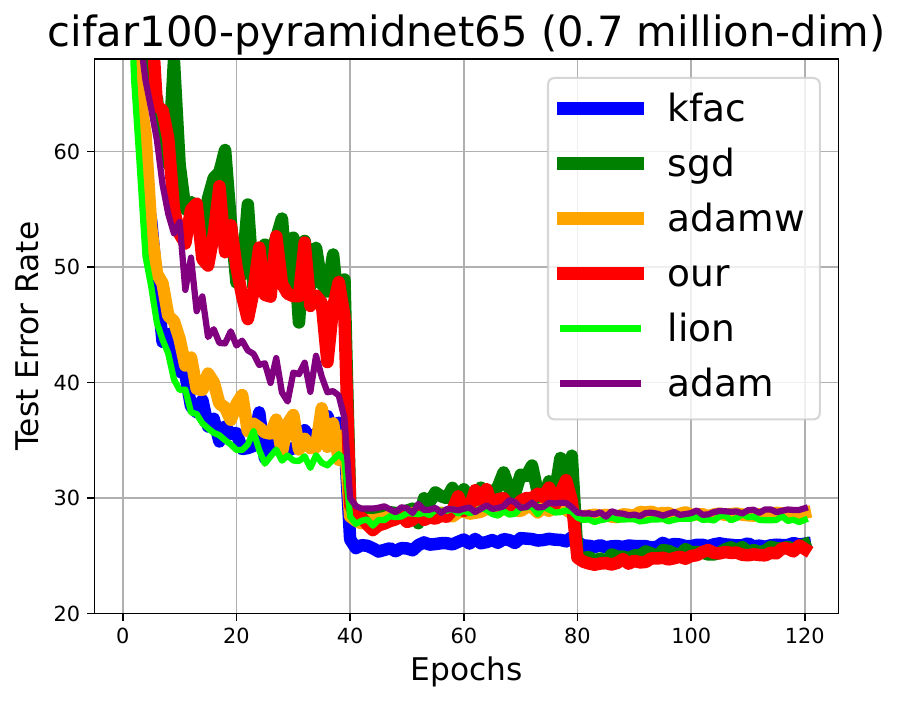}
\hspace*{-0.12cm}	
\includegraphics[width=0.25\linewidth]{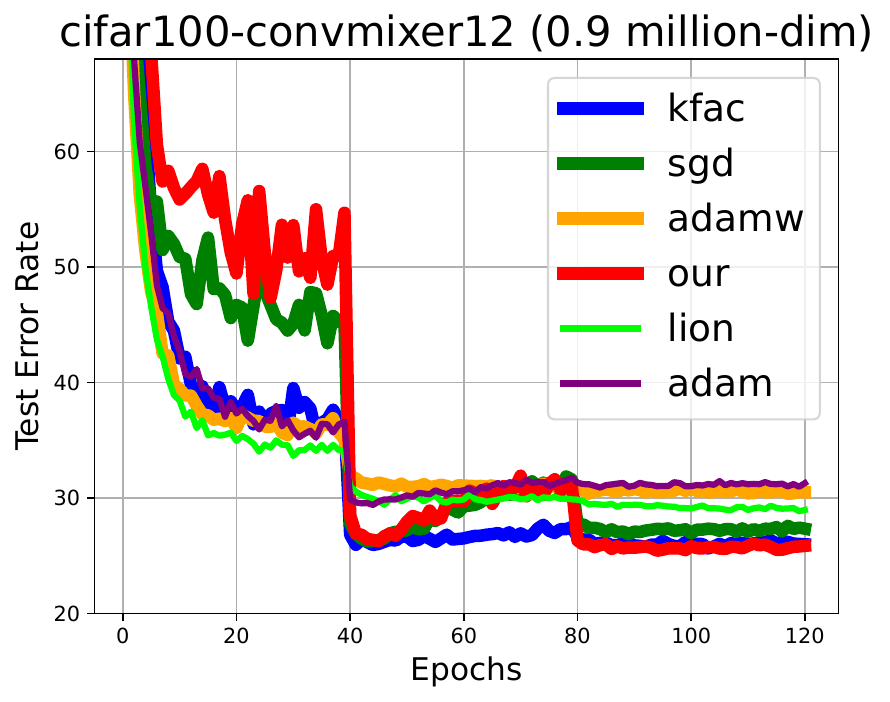}
\hspace*{-0.12cm}	
\includegraphics[width=0.25\linewidth]{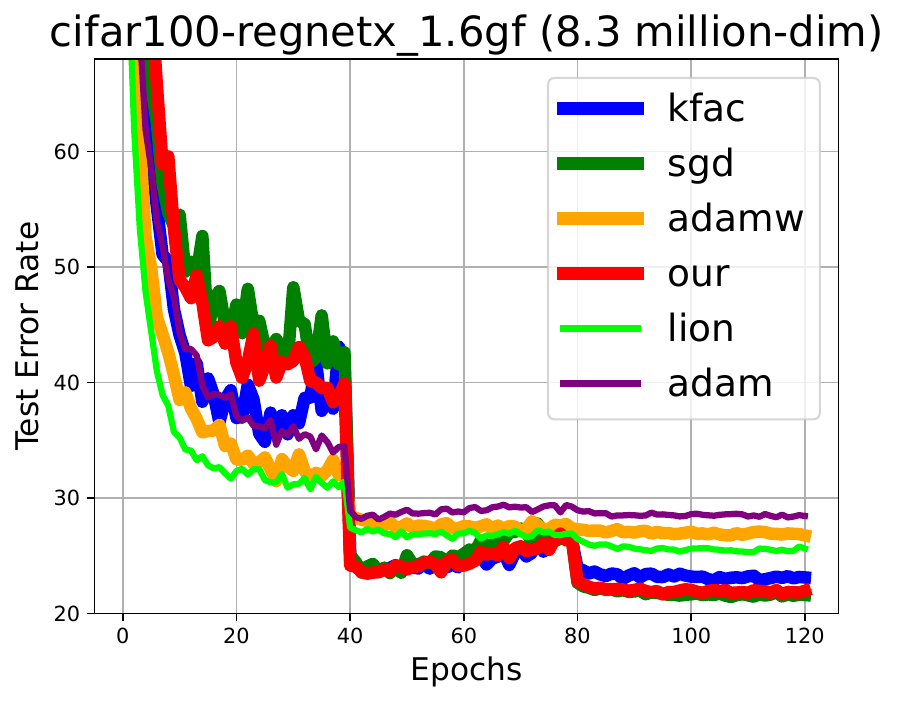}
	\hspace*{-1.2cm}	
 \end{minipage}
  \begin{minipage}[t]{\linewidth}
 	\centering
	\hspace*{-1.0cm}
 	\includegraphics[width=0.25\linewidth]{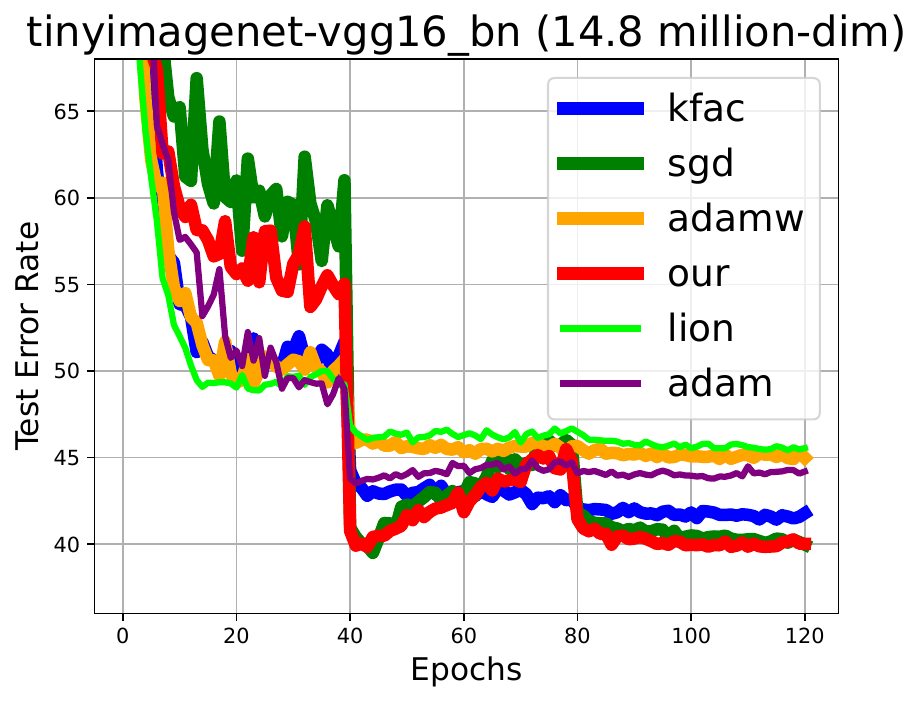}
\hspace*{-0.12cm}		\includegraphics[width=0.25\linewidth]{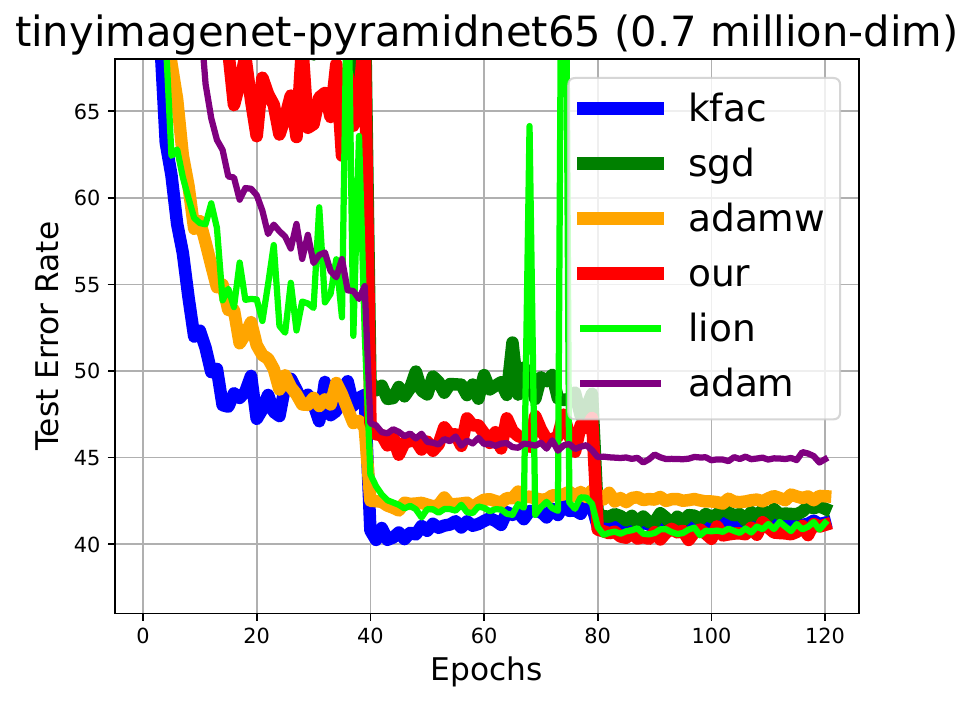}
\hspace*{-0.12cm}		\includegraphics[width=0.25\linewidth]{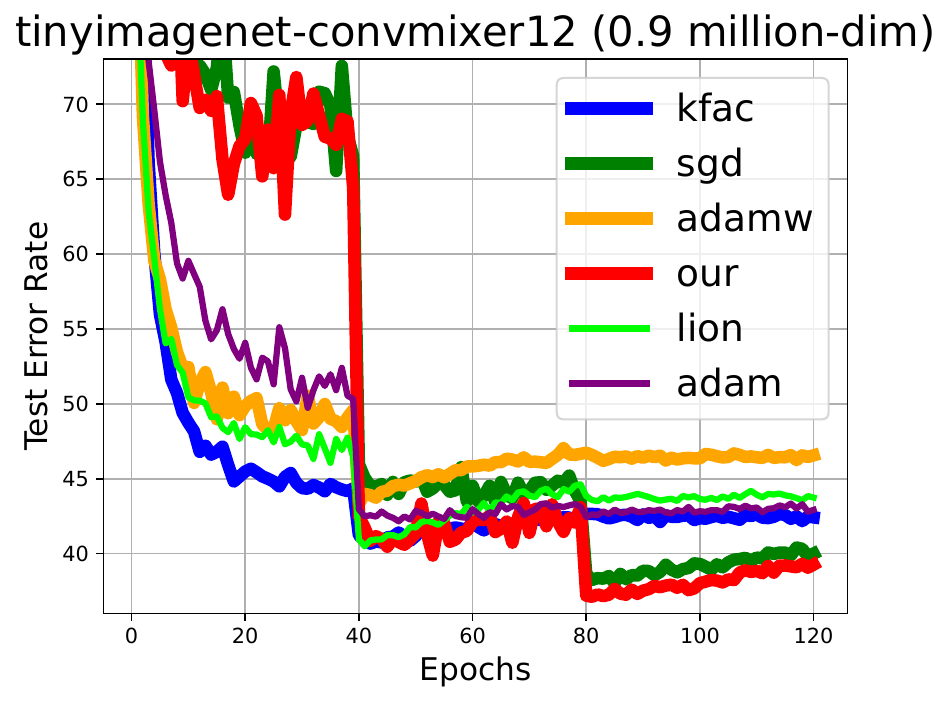}
\hspace*{-0.12cm}		\includegraphics[width=0.25\linewidth]{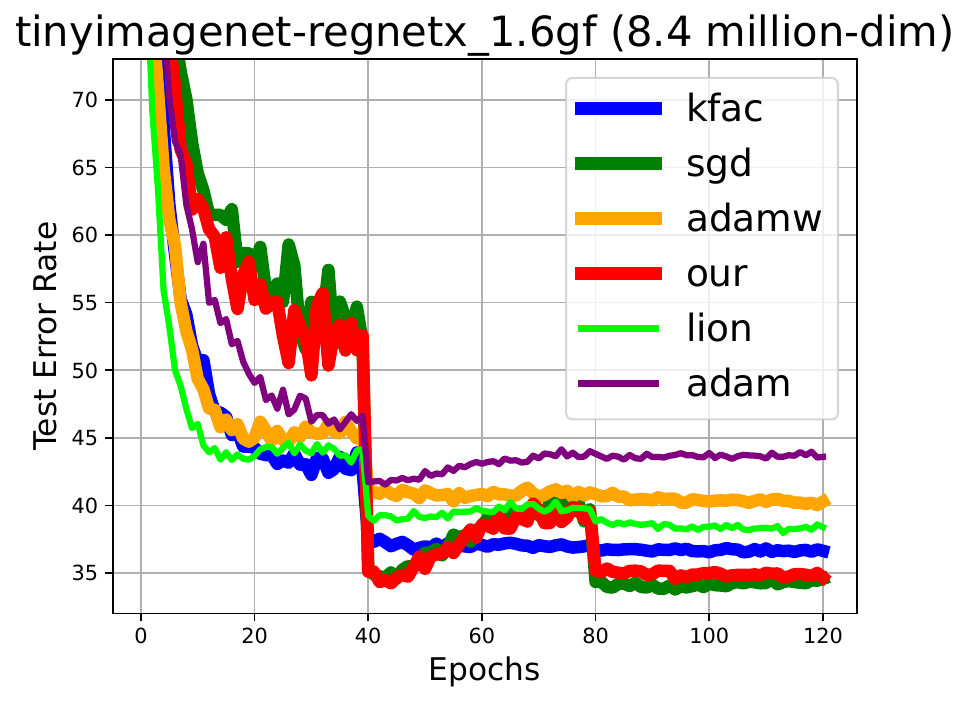}
	\hspace*{-1.2cm}
  \end{minipage}  \vspace*{-7.5mm}
 	\caption{Performance of NN optimizers on more datasets. SGD performs best in the classical model and fairly in the modern models. Our updates achieve competitive test error rates compared to baselines and perform better than KFAC in many cases. 
}
\label{figure:dnn}
\end{figure*}

\begin{figure*}[ht]
	\centering
 \vspace{4mm}
	\hspace*{-1.0cm}
\includegraphics[width=0.25\linewidth]{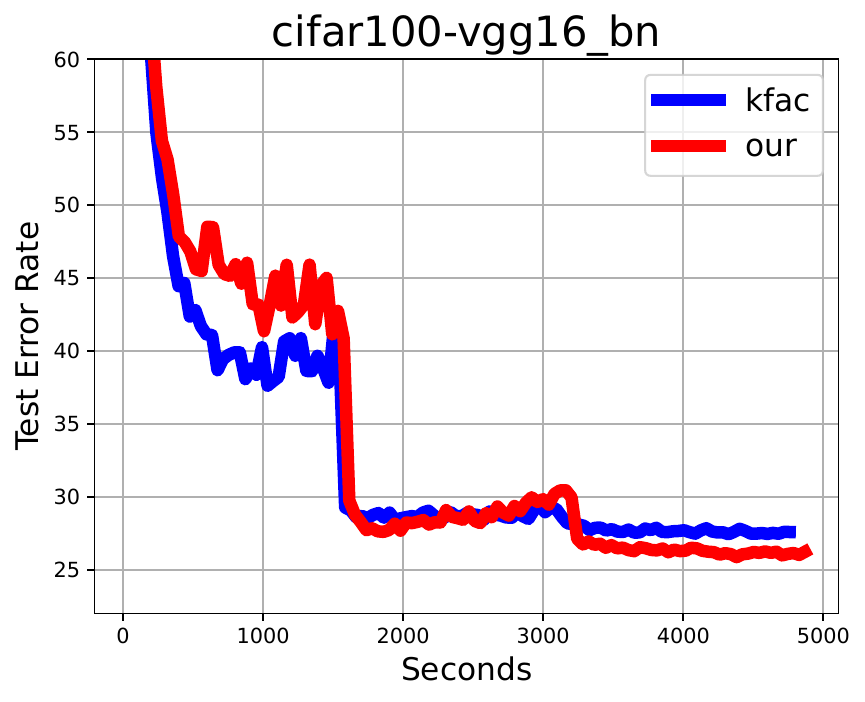}
\hspace*{-0.12cm}		\includegraphics[width=0.25\linewidth]{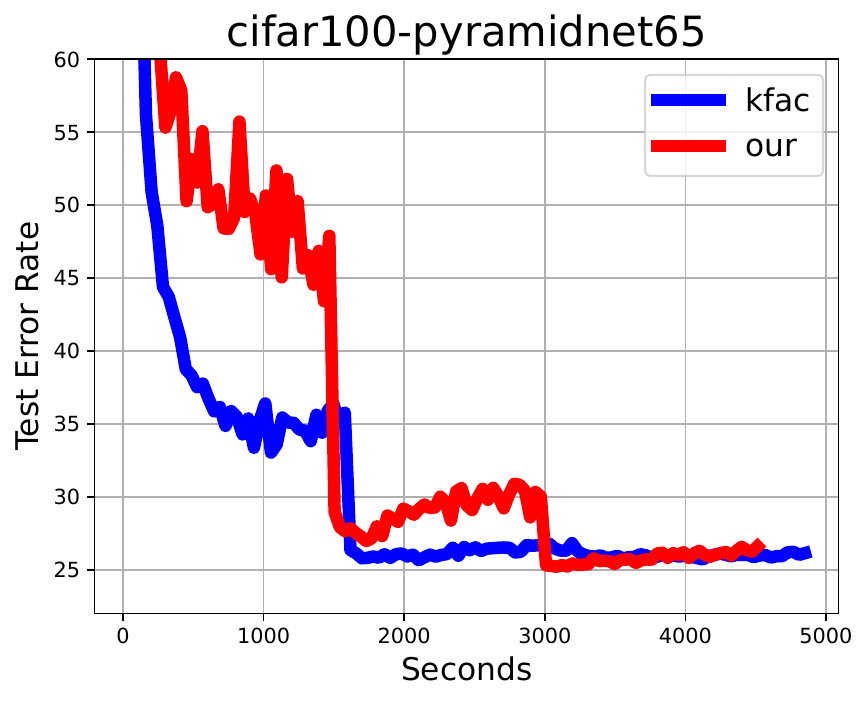}
\hspace*{-0.12cm}		\includegraphics[width=0.25\linewidth]{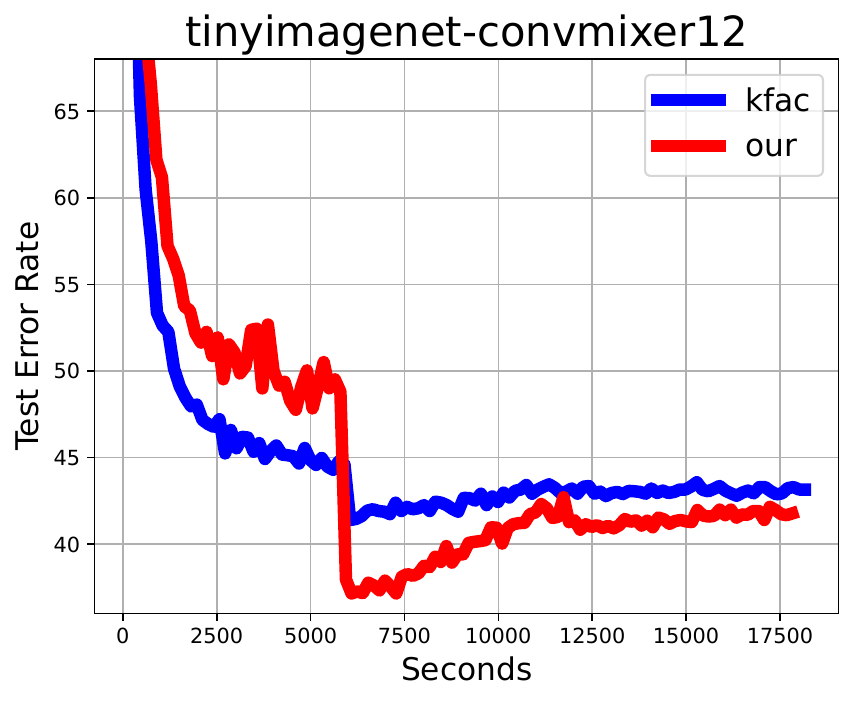}
 \hspace*{-0.12cm}		\includegraphics[width=0.25\linewidth]{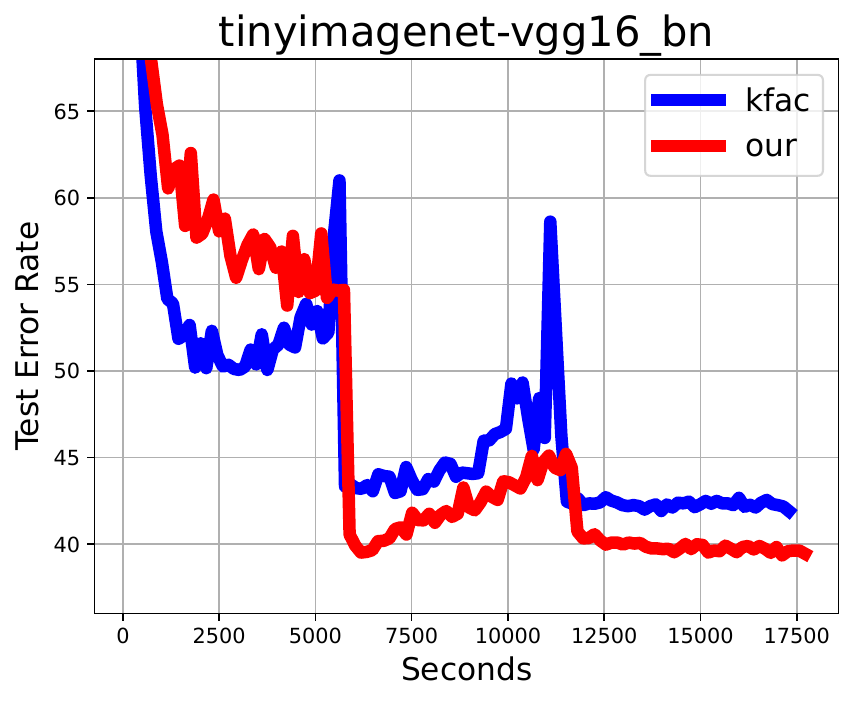}
	\hspace*{-1.2cm}
 \vspace{-2.5mm}
	\caption{Additional results of our method and KFAC using a new random seed. We use these two methods to train NN models in 120 epochs.  We report the performance of the methods in terms of running time.
}
	\label{figure:dnn_time}
\end{figure*}

 \begin{table*}[!htbp]
	\centering
	\footnotesize
	\begin{tabular}{c c c c}
		\toprule
		\begin{tabular}{c}
		Hyperparameter
		\end{tabular} &
		\begin{tabular}{c}
                Meaning
		\end{tabular} &
		\begin{tabular}{c}
		Our Method in Fig.~\ref{fig:matDL_opt}
		\end{tabular} &
		\begin{tabular}{c}
		KFAC in Fig.~\ref{fig:matDL_opt}
		\end{tabular} \\
                \midrule
		\multirow{1}{*}{
		  \begin{tabular}{c}
                    $\stepsize_2$
		\end{tabular}}   
		& Standard stepsize       &  Tuned &  Tuned \\
                \midrule
		\multirow{1}{*}{
		  \begin{tabular}{c}
                    $\alpha_2$
		\end{tabular}}   
		&   Standard momentum weight  & 0.9   & 0.9   \\
                \midrule
		\multirow{1}{*}{
		  \begin{tabular}{c}
                    $\gamma$
		\end{tabular}}   
		&  (L2) weight decay       &  0.01 &  0.01 \\
                  \midrule
		\multirow{1}{*}{
		  \begin{tabular}{c}
                    $\lambda$
		\end{tabular}}   
		& Damping weight       &   0.005; 0.0005 &    0.005; 0.0005 \\
                    \midrule
		\multirow{1}{*}{
		  \begin{tabular}{c}
                    $T$
		\end{tabular}}   
		& Update frequency      &  $10$; $15$;  $25$; $60$ &   $10$; $15$; $25$; $60$ \\
                    \midrule
		\multirow{1}{*}{
		  \begin{tabular}{c}
                    $\theta$
		\end{tabular}}   
		& Moving average in KFAC      & NA   & 0.95 \\
                      \midrule
		\multirow{1}{*}{
		  \begin{tabular}{c}
                    $\stepsize_1$
		\end{tabular}}   
		&  Stepsize to update our preconditioner       & 0.01 & NA \\
                      \midrule
		\multirow{1}{*}{
		  \begin{tabular}{c}
                    $\alpha_1$
		\end{tabular}}   
		&  Momentum weight to update our preconditioner       & 0.5 & NA \\
                \bottomrule \\
	\end{tabular}
   \caption{Hyperparameter configuration  in our update and KFAC. We first choose the damping weight $\lambda$ based on the performance of KFAC and use the same value in our update. For both methods, we set $\lambda= 0.0005, 0.005$ in RepVGG-B1G4, and other models, respectively. To reduce the iteration cost of both methods, we only update the preconditioner at every $T$ iterations.
   For RepVGG-B1G4, we update the preconditioner at every $T=60$ iterations.
   For RegNetZ-500MF and ConvMixer180-16, we update the preconditioner at every $T=25$ iterations.
   For ReXNetr-100,  we update the preconditioner at every $T=15$ iterations. 
   For the rest models,  we update the preconditioner at every $T=10$ iterations. 
   The value of the hyperparameter $\theta$ is chosen as suggested at \url{https://github.com/alecwangcq/KFAC-Pytorch}. 
   We consider the first 500 iterations as a warm-up period to update our preconditioner by using a smaller stepsize $\stepsize_1$: we set $\stepsize_1=0.0002$ for the first 100 iterations, increase it to $\stepsize_1=0.002$ for the next 400 iterations, and finally fix it to $\stepsize_1=0.01$ for the remaining iterations. 
   }
	\label{table:hyperparameters_opt}
      \end{table*}
      
 \begin{table*}[!htbp]
   \begin{minipage}[t]{\linewidth}
	\centering
	\footnotesize
	\begin{tabular}{c c c c c}
		\toprule
		\begin{tabular}{c}
                  Dataset  \\
		\end{tabular} &
		\begin{tabular}{c}
                  VGG-16-BN  \\  
		\end{tabular} &
		\begin{tabular}{c}
                  PyramidNet-65 
                   \\  
		\end{tabular} &
		\begin{tabular}{c}
			ConvMixer256-12    \\ 
		\end{tabular} &
  		\begin{tabular}{c}
                  RegNetX-1.6GF   \\
		\end{tabular} 
                \\
                \midrule
		\multirow{1}{*}{
		  \begin{tabular}{c}
                    CIFAR-100
		\end{tabular}}   
		&   14,774,436     & 707,428  & 911,204 & 8,368,436 \\
                \midrule
		\multirow{1}{*}{
		  \begin{tabular}{c}
                    TinyImageNet-200
		\end{tabular}}   
		&  14,825,736      &  733,128  &  936,904 &  8,459,736 \\
                \bottomrule \\
	\end{tabular}
 \end{minipage}
 
    \begin{minipage}[!htbp]{\linewidth}
	\centering
	\footnotesize
	\begin{tabular}{c c c c c}
		\toprule
		\begin{tabular}{c}
                  Dataset  \\
		\end{tabular} &
		\begin{tabular}{c}
                  RegNetZ-500MF   
		\end{tabular} 
            & \begin{tabular}{c}
              RepVGG-B1G4    
		\end{tabular}
  & \begin{tabular}{c}
              ConvMixer180-16    
		\end{tabular}
  & \begin{tabular}{c}
              ReXNetr-100    
		\end{tabular}
                \\
                \midrule
		\multirow{1}{*}{
		  \begin{tabular}{c}
                    ImageNet-100
		\end{tabular}}   
		&   6,200,242  & 38,121,956 & 721,900 & 3,730,404
  \\
                \bottomrule \\
	\end{tabular}
 \end{minipage} \vspace*{-5mm}
   \caption{Number of Learnable Parameters in the NN Models Considered}
	\label{table:nn_models}
      \end{table*}

 \begin{table*}[!htbp]
 \vspace{5mm}
	\centering
	\footnotesize
	\begin{tabular}{c c c c c}
		\toprule
		\begin{tabular}{c}
                  Dataset  \\
		\end{tabular} &
		\begin{tabular}{c}
                  Input Dimension   
		\end{tabular} &
		\begin{tabular}{c}
                  Number of Classes  
		\end{tabular} &
  		\begin{tabular}{c}
                  Number of Training Points  
		\end{tabular} &
  		\begin{tabular}{c}
                 Number of Test Points    
		\end{tabular} 
             \\
                \midrule
		\multirow{1}{*}{
		  \begin{tabular}{c}
                    CIFAR-100
		\end{tabular}}   
		&    $32 \times 32$    &    100  & 50,000  & 10,000 \\
                \midrule
		\multirow{1}{*}{
		  \begin{tabular}{c}
                    TinyImageNet-200
		\end{tabular}}   
		&     $64 \times 64$   &   200 & 100,000 & 10,000  
  \\
            \midrule
  		\multirow{1}{*}{
		  \begin{tabular}{c}
                   ImageNet-100
		\end{tabular}}   
		&    $224 \times 224$    &  100 & 130,000 &  5,000 \\  
                \bottomrule \\
	\end{tabular}
 \vspace*{-5mm}
   \caption{Statistics of the Datasets}
	\label{table:data_stats}
      \end{table*}

\newpage

\section{Summary of  Generalized Normal Coordinates}
\label{app:summary}

\begin{table}[H]
\begin{minipage}{\columnwidth}
\centering
\begin{tabular}{|l|l|l|l}
   \hline
SPD (sub)manifold  $\mathcal{M}$
& Name
   & Our normal coordinate 
    \\
    \hline
$\{ \gparam \in \real^{k \times k} \big| \gparam \in \mathcal{S}_{++}^{k \times k} \}$  
& Full   manifold  
& See Eq.~\eqref{eq:gen_normal_param}
    \\
with affine-invariant metric $\vF$ && \\
    
    \hline

$\Big\{ \gparam=\begin{bmatrix} \vV & \vmu \\ \vmu^T & 1 \end{bmatrix} \in \real^{k \times k} \Big| \gparam \in \mathcal{S}_{++}^{k \times k}  \Big\}$  
&  Submanifold induced
  & See Eq.~\eqref{eq:gnormal_coord1},\eqref{eq:gnormal_coord2} 
    \\ with affine-invariant metric $\vF$ 
&  by Siegel embedding 
&\\
    \hline

$\Big\{ \gparam= \vU \otimes \vW \in \real^{pd \times pd} \Big| \vU \in \mathcal{S}_{++}^{p \times p} , \vW \in \mathcal{S}_{++}^{d \times d}  \Big\}$  
& Kronecker-product   submanifold 
  & See Eq.~\eqref{eq:mat_gauss_norm_coord} 
    \\
with an approximated affine-invariant metric $\vF$ && \\
    \hline

\end{tabular}
\end{minipage}
\vspace{-0.15cm}
  \caption{ Summary of our normal coordinates, where $\mathcal{S}_{++}$ denotes   the set of SPD matrices }
 \label{tab:updates}
 \vspace{0.1cm}
\end{table}

\section{Background}

In this section, we will assume $\gparam$ is a (learnable) vector to simplify  notations. For a SPD matrix $\vM \in \mathcal{S}_{++}^{k \times k}$, we could consider $\gparam = \mathrm{vech}(\vM) $, where $\mathrm{vech}(\vM)$ returns a $\frac{k(k+1)}{2}$-dimensional array obtained by vectorizing only the lower triangular part of $\vM$, which is known as the half-vectorization function.
 
\subsection{Fisher-Rao Metric}
\label{apd:fim}

Under parametrization $\gparam$, the Fisher-Rao Metric is defined as
\begin{align}
\vF(\gparam^{\text{(cur)}}) :=
- E_{q(\lat|\sgparam)} [ \nabla_\sgparam^2 \log q(\vlat|\gparam) ] \big|_{\gparam=\gparam^{\text{(cur)}} },
\end{align} where $q(\vlat|\gparam)$ is a probabilistic distribution parameterized by $\gparam$, such as a Gaussian distribution with zero mean and covariance $\gparam$.

\subsection{Christoffel Symbols}
\label{apd:chris_sym}
Given a Riemannian metric $\vF$, the Christoffel symbols of the first kind  are defined as
\begin{align}
\Gamma_{d,a b}(\gparam_1) :=\half [\partial_a F_{bd}(\gparam) + \partial_b F_{ad}(\gparam) - \partial_d F_{ab}(\gparam) ]\Big|_{\sgparam=\sgparam_1},
\label{eq:chris_1st}
\end{align} where    $F_{bd}(\gparam)$ denotes the $(b,d)$ entry of the metric $\vF_\sgparam$ and $\partial_b$ denotes the partial derivative w.r.t. the $b$-th entry of $\gparam$. 

The Christoffel symbols of the second kind are defined as \begin{equation} \Gamma_{\, ab}^{c}( \gparam ) := \sum_d F^{cd}(\gparam)  \Gamma_{d,ab}(\gparam) , \end{equation}  where
   $F^{cd}(\gparam)$ denotes the (c,d) entry of the inverse $\vF^{-1}$ of the metric.
Observe that the Christoffel symbols of the second kind
involve  computing all partial derivatives of the metric $\vF$ and the inverse of the metric.

\subsection{Riemannian  Exponential Map}
\label{apd:expmap}
The Riemannian  exponential map is defined via a geodesic, which generalizes the notion of a straight line to a manifold.
The geodesic $\vr(t)$ satisfies the following second-order nonlinear system of ODEs with initial values $\vr(0)=\vx$ and $\dot{\vr}(0)=\vnu$, where $\vx$ denotes a point on the manifold and $\vnu$ is a Riemannian gradient,
\begin{align}
\text{geodesic ODE:} \,\,\,\,   \ddot{r}^{c}(t) + \sum_{a,b} \Gamma_{\, ab}^{c}( \vr(t)  )  \dot{r}^{a}(t) \dot{r}^{b}(t) =0,
\end{align} where ${r}^{c}(t)$ denotes the $c$-th entry of $\vr(t)$.

The Riemannian  exponential map is defined as
\begin{align}
    \RExp (\vx,\vnu) := \vr(1) ,
\end{align} where $\vx$ denotes an initial point and $\vnu$ is an initial Riemannian gradient so that 
$\vr(0)=\vx$ and $\dot{\vr}(0)=\vnu$.

\subsection{Riemannian (Parallel) Transport Map}
\label{apd:trans_rg}
In a curved space, the transport map along a given curve generalizes the notion of parallel transport. In Riemannian optimization, we consider the transport map along  a geodesic curve. 
Given a geodesic curve $\vr(t)$, a smooth Riemannian gradient field  denote by $\vv(t)$ that satisfies the following first-order linear system of ODEs with initial value $\vv(0)=\vnu$,
\begin{align}
 \text{transport ODE:} \,\,\,\,   \dot{v}^{c}(t) + \sum_{a,b} \Gamma_{\, ab}^{c}( \vr(t)  )  {v}^{a}(t) \dot{r}^{b}(t) =0. \label{eq:parallel_trans_riem}
\end{align} 

\vspace{-4.5mm} 

The transport map $\hat{\transport}_{\sgparam^{\text{(cur)}} \rightarrow \sgparam^{\text{(new)}}}(\vnu)$  transports  the Riemannian gradient $\vnu$ at  $\gparam^{\text{(cur)}}$ to  $\gparam^{\text{(new)}}$ as follows,
\begin{align}
    \hat{\transport}_{\sgparam^{\text{(cur)}} \rightarrow \sgparam^{\text{(new)}}}(\vnu):=\vv(1),
\end{align} 

\vspace{-4mm} 

where $\vr(0)=\gparam^{\text{(cur)}}$, $\vr(1)=\gparam^{\text{(new)}}$, and $\vv(0)=\vnu$. It can be computationally challenging to solve this linear ODE due to the presence of the Christoffel symbols.

\subsection{Euclidean (Parallel) Transport Map}
\label{apd:trans_eg}
Given a geodesic curve $\vr(t)$,   a smooth Euclidean gradient field denote by $\vomega(t)$  on manifold $\mathcal{M}$ that satisfies the following first-order linear system of ODEs with initial value  $\vomega(0)=\vm$,
\begin{align}
  \text{transport ODE:} \,\,\,\,    \dot{\omega}_{c}(t) - \sum_{a,b} \Gamma_{\, cb}^{a}( \vr(t) )  {\omega}_{a}(t) \dot{r}^{b}(t) =0 .\label{eq:parallel_trans_eucl}
\end{align} 

\vspace{-4.5mm} 

The transport map $\transport_{\sgparam^{\text{(cur)}} \rightarrow \sgparam^{\text{(new)}}}(\vg)$  transports the Euclidean gradient $\vg$  at  $\gparam^{\text{(cur)}}$ to $\gparam^{\text{(new)}}$ as shown below,
\begin{align}
\transport_{\sgparam^{\text{(cur)}} \rightarrow \sgparam^{\text{(new)}}}(\vg):=\vomega(1),
\end{align} 

\vspace{-4mm} 

where $\vr(0)=\gparam^{\text{(cur)}}$, $\vr(1)=\gparam^{\text{(new)}}$, and $\vomega(0)=\vg$.

\subsection{Update \ref{eq:rgd_mom_eg} is Equivalent to \citet{alimisis2020continuous}}
\label{apd:equ_alimi}

Note that $\vm$ and $\vz$ are initialized by zero.
Due to Eq.~\eqref{eq:equ_trans_maps}, updates
\eqref{eq:rgd_mom} and \eqref{eq:rgd_mom_eg} are equivalent since
$\vm^{\text{(cur)}}=\vF(\gparam^{\text{(cur)}})\vnu^{\text{(cur)}}$ and 
$\vw^{\text{(new)}}=\vF(\gparam^{\text{(new)}})\vz^{\text{(new)}}$, where $\vm^{\text{(cur)}}$ and $\vw^{\text{(new)}}$ are defined in Eq.~\eqref{eq:rgd_mom_eg} while 
$\vnu^{\text{(cur)}}$ and $\vz^{\text{(new)}}$ are defined in Eq.~\eqref{eq:rgd_mom}

\section{Simplification of the vector-Jacobian Product}
\label{apd:jac_vec_prod}
The vector-Jacobian product in \eqref{eq:jacobian_transform} could, in general, be computed by automatic differentiation. We give two cases for SPD (sub)manifolds where the product can be explicitly simplified. For notation simplicity, we denote
 $\vm =\vm^{(\lparam_0)} $ and $\vw=\vw^{(\lparam_1)}$. Thus $\vw=\vm$  when we use the approximation in Eq.~\eqref{eq:transport_approx}.

\subsection{Symmetric Cases}
Suppose that $\lparam $ is symmetric, in which case $\vm$ is also symmetric due to update \eqref{eq:local_rgd_mom}.
We further denote  $\vA_0=\vA^{\text{(cur)}}$ and  $\vA_1=\vA^{\text{(new)}}$, where $\vA_1 = \vA_0\MExp( -\half \vm^{(\lparam_0)} )= \vA_0\MExp( -\half \vm  ) $.

Recall that $\gparam =\vA_0 \MExp( \lparam )\vA_0^T = \vA_1 \MExp(  \vxi )\vA_1^T$. Thus, we have
\begin{align}
    \MExp( \lparam ) = \MExp( -\half \vm  ) \MExp(  \vxi ) \MExp^T( -\half \vm  ) =  \MExp( -\half \vm  ) \MExp(  \vxi ) \MExp( -\half \vm ).
\end{align} By the Baker--Campbell--Hausdorff formula, we have
\begin{align}
\MExp^{-1}(\MExp(\vN) \MExp(\vM)  ) &= \vN + \vM + \sum_i w_i ( \vM^{a_i} \vN  \vM^{b_i} - \vM^{c_i} \vN  \vM^{d_i}  ) + O(\vN^2) ,  \\[-1em]
\MExp^{-1}(\MExp(\vM) \MExp(\vN)   ) &=  \vN + \vM   + \sum_i w_i ( \vM^{a_i} \vN  \vM^{b_i} - \vM^{c_i} \vN  \vM^{d_i}  ) + O(\vN^2) ,
\end{align} 

\vspace{-5mm} 

where $a_i$, $b_i$, $c_i$, $d_i$ are non-negative integers satisfying $a_i+b_i=c_i+d_i>0$, and $w_i$ is a coefficient.  

Since we evaluate the Jacobian at $\vxi_0=\mathbf{0}$, we can get rid of the higher-order term $O(\vxi^2)$, which leads to the following simplification,
\begin{align}
\lparam = \MExp^{-1} ( \MExp(-\half \vm) \MExp(\vxi)\MExp(-\half \vm)  ) =  \vxi -\vm + \sum_i w_i ( \vm^{a_i} \vxi  \vm^{b_i} - \vm^{c_i} \vxi  \vm^{d_i}  ) + O(\vxi^2) .
\end{align}
Recall that $\vm=\vw$ is symmetric. The vector-Jacobian product can be simplified as 
\begin{align}
\vw^T \vJ(\vxi_0)  & = \vm^T \big[ \frac{\partial \lparam}{\partial \vxi} \big|_{\xi=\xi_0} \big]  \nonumber  \\
&= \nabla_\xi \mathrm{Tr} (\vm^T \lparam)\big|_{\xi=\xi_0} \nonumber \\
&= \nabla_\xi \mathrm{Tr} \big(\vm^T [  \vxi -\vm + \sum_i w_i ( \vm^{a_i} \vxi  \vm^{b_i} - \vm^{c_i} \vxi  \vm^{d_i}  ) + O(\vxi^2)    ] \big)\big|_{\xi=\xi_0} \nonumber  \\
&=  \nabla_\xi \mathrm{Tr} \big(\vm^T [  \vxi + \sum_i w_i ( \vm^{a_i} \vxi  \vm^{b_i} - \vm^{c_i} \vxi  \vm^{d_i}  )    ] \big)\big|_{\xi=\xi_0} \nonumber  \\
&=  \nabla_\xi \mathrm{Tr} \big(\vm  [  \vxi + \sum_i w_i ( \vm^{a_i} \vxi  \vm^{b_i} - \vm^{c_i} \vxi  \vm^{d_i}  )    ] \big)\big|_{\xi=\xi_0} \nonumber  \\
&=  \nabla_\xi \mathrm{Tr} \big( \vm    \vxi  + \sum_i w_i ( \vm^{a_i+b_i+1} \vxi   - \vm^{c_i+d_i+1} \vxi    )   \big)\big|_{\xi=\xi_0} \nonumber \\
&=  \nabla_\xi \mathrm{Tr} (\vm    \vxi    )\big|_{\xi=\xi_0} \nonumber \\
& = \vm=\vw. 
\label{eq:jacob_vec_sym}
\end{align}

\hfill 

\subsection{Triangular Cases}
Without loss of generality, we assume $\lparam $ is lower-triangular, in which case $\vm$ is also lower-triangular due to update \eqref{eq:local_rgd_mom}.
Similarly, we denote  $\vA_0=\vA^{\text{(cur)}}$ and  $\vA_1=\vA^{\text{(new)}}$, where $\vA_1 = \vA_0\MExp( - \vD \odot \vm^{(\lparam_0)} )= \vA_0\MExp( - \vD \odot \vm  ) $, and where $\vD=  \frac{1}{\sqrt{2}} \mathrm{Tril}(\mathbf{1}) + (\half - \frac{1}{\sqrt{2}}\mathbf{1}) \vI$ is chosen so that the metric is orthonormal at $\lparam_0$, and $\mathrm{Tril}(\mathbf{1})$ denotes a lower-triangular matrix of ones.

Recall that $\gparam =\vA_0 \MExp(\vD \odot \lparam )\MExp^T(\vD \odot \lparam )\vA_0^T = \vA_1 \MExp(\vD \odot  \vxi )\MExp^T(\vD \odot  \vxi )\vA_1^T$. Thus, we have
\begin{align}
    \MExp( \vD \odot \lparam )  \MExp^T( \vD \odot \lparam ) 
    &= \MExp( -\vD \odot \vm  )  \MExp( \vD \odot \vxi )  \MExp^T( \vD \odot  \vxi )  \MExp^T( -\vD \odot \vm  ) .
\end{align}

Since $\lparam$, $\vm$ and $\vxi$ are lower-triangular,
$\MExp( \vD \odot \lparam ) $, $\MExp( -\vD \odot \vm  )$, and $ \MExp( \vD \odot \vxi )$ are also lower-triangular. Moreover, all the eigenvalues of  $\MExp( \vD \odot \lparam ) $, $\MExp( -\vD \odot \vm  )$, and $ \MExp( \vD \odot \vxi )$  are positive.

Note that we make use of the uniqueness of the Cholesky decomposition since  $\MExp( \vD \odot \lparam )$ can be viewed as a Cholesky factor.
Thus, $\MExp( \vD \odot \lparam ) = \MExp( -\vD \odot \vm  )  \MExp( \vD \odot \vxi ) $. 

By the Baker--Campbell--Hausdorff formula, we have
\begin{align}
\MExp^{-1} \big( \MExp( \vM  )  \MExp( \vN  )  \big)  = \vM + \vN + \half  \LieBracket{\vM}{\vN}  + O( \LieBracket{\vM}{ \LieBracket{\vM}{\vN} }  ) + O(\vN^2),
\end{align} where $\LieBracket{ \vM }{  \vN  }=\vM\vN-\vN\vM$ is the Lie bracket. 
Thus, we have
\begin{align}
 \lparam = \left(\vD \odot \vxi -\vD \odot \vm  + \half   \LieBracket{ -\vD \odot \vm }{  \vD \odot \vxi  }  ) \right) \oslash \vD + O( \LieBracket{\vm}{ \LieBracket{\vm}{\vxi} }  + O(\vxi^2),
\end{align} where $ \oslash$ denotes elementwise division.

We get rid of the higher-order term $ O(\vxi^2)$ by evaluating $\vxi=\vxi_0=\mathbf{0}$.

Recall that $\vm=\vw$.   The vector-Jacobian  product can be simplified as 
\begin{align}
\vw^T \vJ(\vxi_0)  & = \vm^T \big[ \frac{\partial \lparam}{\partial \vxi} \big|_{\xi=\xi_0} \big]  \nonumber \\
&= \nabla_\xi \mathrm{Tr}( \vm^T  \lparam ) \big|_{\xi=\xi_0} \nonumber \\
&= \nabla_\xi \mathrm{Tr}( \vm^T   (\vD \odot \vxi -\vD \odot \vm  + \half   \LieBracket{ -\vD \odot \vm }{  \vD \odot \vxi  }  \oslash \vD ) + O( \LieBracket{\vm}{ \LieBracket{\vm}{\vxi} }  ) ) \big|_{\xi=\xi_0} \nonumber\\
&= \nabla_\xi \mathrm{Tr}( (\vm  \oslash \vD )^T   (\vD \odot \vxi -\vD \odot \vm  + \half   \LieBracket{ -\vD \odot \vm }{  \vD \odot \vxi  } +  O( \LieBracket{\vm}{ \LieBracket{\vm}{\vxi} }  )  ) ) \big|_{\xi=\xi_0} \nonumber\\
&= \underbrace{\vm}_{ O(\stepsize) } + \half  \vD \odot \underbrace{ \LieBracket{(-\vD \odot \vm )^T  }{ \vm  \oslash \vD   }}_{ O(\stepsize^2) } + \underbrace{ O\left( \LieBracket{ \vm  }{  \LieBracket{ \vm  }{ \vm^T  } } \right) }_{ O(\stepsize^3)  } .
\end{align}

\hfill 

\section{Simplification of the Metric Calculation at $\lparam_0$}
\label{apd:simple_fim}

Consider $\gparam=\vphi_{\sgparam^{\text{(cur)}}}(\lparam)=\vA\vA^T \in \mathcal{S}_{++}^{k \times k}$, where  $\vA=\vA^{\text{(cur)}}\MExp(\vD \odot \lparam)$ .

For notation simplicity, we let  $\vA_0=\vA^{\text{(cur)}}$ and $\gparam_0= \gparam^{\text{(cur)}}=\vA_0\vA_0^T$. 
Let $\tilde{\lparam}$ denote the vector representation of the learnable part of $\lparam$.
By definition of the affine-invariant metric, we have
\begin{align}
\vF(\lparam_0) &= -2 \Unmyexpect{\gauss(\mathbf{0}, \gparam)}\big[ \nabla_{ \tilde{\slparam} }^2 \log \gauss(\mathbf{0}, \gparam) \big] \big|_{\slparam=\slparam_0} \nonumber \\
&= \Unmyexpect{\gauss(x|\mathbf{0}, \gparam)}\big[ \nabla_{ \tilde{\slparam} }^2 \big\{ \mathrm{Tr} [ \vx\vx^T  \vA_0^{-T} \MExp^T( -\vD \odot \lparam) \MExp( - \vD \odot \lparam)  \vA_0^{-1}   ] +  2 \log \mathrm{det} \ \MExp(\vD \odot \lparam) \big\}   \big] \big|_{\slparam=\slparam_0} \nonumber  \\
&= \nabla_{ \tilde{\slparam} }^2 \big\{ \mathrm{Tr} [ \Unmyexpect{\gauss(x|\mathbf{0}, \gparam_0)}\big[\vx\vx^T\big]  \vA_0^{-T} \MExp^T( -\vD \odot \lparam) \MExp( - \vD \odot \lparam)  \vA_0^{-1}   ] +  2 \log \mathrm{det} \ \MExp(\vD \odot \lparam) \big\}    \big|_{\slparam=\slparam_0} \nonumber \\
&= \nabla_{ \tilde{\slparam} }^2 \big\{ \mathrm{Tr} [ [\vA_0\vA_0^T]      \vA_0^{-T}  \MExp^T( -\vD \odot \lparam) \MExp( - \vD \odot \lparam) \vA_0^{-1}   ] +  2 \log \mathrm{det} \ \MExp(\vD \odot \lparam) \big\}    \big|_{\slparam=\slparam_0} \nonumber \\
&= \nabla_{ \tilde{\slparam} }^2 \big\{ \mathrm{Tr} [   \MExp^T( -\vD \odot \lparam) \MExp( - \vD \odot \lparam)   ] + 2 \log \mathrm{det} \ \MExp(\vD \odot \lparam) \big\}    \big|_{\slparam=\slparam_0} \nonumber \\
&=  \nabla_{ \tilde{\slparam} }^2 \big\{ \mathrm{Tr} [   \MExp^T( -\vD \odot \lparam) \MExp( - \vD \odot \lparam)   ] +  2 \mathrm{Tr}[ \vD \odot \lparam ]  \big\}    \big|_{\slparam=\slparam_0} \,\, \text{\small(ignore linear terms for a 2nd order derivative)} \nonumber \\
&= \nabla_{ \tilde{\slparam} }^2 \big\{ \mathrm{Tr} [   \MExp^T( -\vD \odot \lparam) \MExp( - \vD \odot \lparam)   ]  \big\}    \big|_{\slparam=\slparam_0} .
\label{eq:metric_simple}
\end{align}
Note that we express $\MExp( - \vD \odot \lparam)$ as $\MExp( - \vD \odot \lparam)=\vI -\vD\odot \lparam +\half(\vD\odot \lparam)^2 + O(\lparam^3)  $. Since we evaluate the metric at $\lparam_0=\mathbf{0}$, we can get rid of the higher-order term $ O(\lparam^3)$, which leads to the following simplification,
 \begin{align}
&  \nabla_{\slparam_{ij}}   \mathrm{Tr} [   \MExp^T( -\vD \odot \lparam) \MExp( - \vD \odot \lparam)   ] \nonumber \\
& \qquad \quad = 2   \mathrm{Tr} [ \MExp( - \vD \odot \lparam)  [ \nabla_{\slparam_{ij}} \MExp^T( -\vD \odot \lparam)]    ] \nonumber \\
&  \qquad \quad = 2D_{ij}
 \nabla_{ \slparam_{ij} }  \big\{ \mathrm{Tr} [ \MExp( - \vD \odot \lparam)   [  - \vE_{ij} + \half \vE_{ij}(\vD \odot \lparam ) + \half (\vD \odot \lparam ) \vE_{ij} +
 O(\lparam^2) ]^T  ]  \big\}  .
 \end{align}
 Thus, we have
  \begin{align}
  & \nabla_{\slparam}   \mathrm{Tr} [   \MExp^T( -\vD \odot \lparam) \MExp( - \vD \odot \lparam)   ] \nonumber \\
  & \qquad  =   \vD \odot [-2 \MExp( - \vD \odot \lparam) + \MExp( - \vD \odot \lparam)(\vD \odot \lparam )^T + (\vD \odot \lparam )^T  \MExp( - \vD \odot \lparam)] +O(\lparam^2) .
  \end{align}

To show $\vF(\lparam_0)=\vI$, we show that $\vF(\lparam_0) \vv = \vv$ for any $\vv$. Let $\vV$ be the matrix  representation of $\vv$, which has the same structure as $\lparam$, such as being symmetric or being lower-triangular. Then,
\begin{align*}
& \vF(\lparam_0) \vv  \\ 
&=  \nabla_{ \tilde{\slparam} }^2 \big\{ \mathrm{Tr} [   \MExp^T( -\vD \odot \lparam) \MExp( - \vD \odot \lparam)   ]  \big\}    \big|_{\slparam=\slparam_0} \vv \\
&= \nabla_{ \tilde{\slparam} }  \big\{ \vv^T  \nabla_{ \tilde{\lparam}  }   \mathrm{Tr} [   \MExp^T( -\vD \odot \lparam) \MExp( - \vD \odot \lparam)   ]  \big\} \big|_{\slparam=\slparam_0}, \,\,\, (\text{note: } \tilde{\slparam}, \vv \text{ are vectors}) \\
&= \nabla_{ \tilde{\slparam} }  \mathrm{Tr}\big\{ \vV^T   \nabla_{\slparam}   \mathrm{Tr} [   \MExp^T( -\vD \odot \lparam) \MExp( - \vD \odot \lparam)   ]  \big\}\big|_{\slparam=\slparam_0} \,\,\, (\text{note: } \lparam, \vV \text{ are matrices})  \\
&= \nabla_{ \tilde{\slparam} }  \mathrm{Tr}\big\{ \vV^T ( \vD \odot [-2 \MExp( - \vD \odot \lparam) + \MExp( - \vD \odot \lparam)(\vD \odot \lparam )^T + (\vD \odot \lparam )^T  \MExp( - \vD \odot \lparam)] +O(\lparam^2))     \big\} \big|_{\slparam=\slparam_0} .
\end{align*}

We can get rid of the higher-order term $O(\lparam^2)$ by evaluating at $\lparam=\lparam_0=\mathbf{0}$ and noting that \begin{equation} \mathrm{Tr}(\vA^T (\vD\odot \vB) )= \sum( \vA \odot (\vD \odot  \vB) ) = \mathrm{Tr}( (\vD\odot\vA)^T \vB ) .\end{equation}
Also note that
\begin{align}
& \nabla_{  \slparam_{ij} }  \mathrm{Tr}\big\{ (\vD \odot \vV)^T  [-2 \MExp( - \vD \odot \lparam) + \MExp( - \vD \odot \lparam)(\vD \odot \lparam )^T + (\vD \odot \lparam )^T  \MExp( - \vD \odot \lparam)]      \big\} \big|_{\slparam=\slparam_0} \nonumber \\
& \qquad =  \mathrm{Tr}\big\{ (\vD \odot \vV)^T 2 D_{ij} [  \vE_{ij} +  \vE_{ij}^T ]  \big\} \big|_{\slparam=\slparam_0}.
\end{align}
Thus, we have 
\begin{align}
& \nabla_{  \slparam  }  \mathrm{Tr}\big\{ (\vD \odot \vV)^T  [-2 \MExp( - \vD \odot \lparam) + \MExp( - \vD \odot \lparam)(\vD \odot \lparam )^T + (\vD \odot \lparam )^T  \MExp( - \vD \odot \lparam)]      \big\} \big|_{\slparam=\slparam_0} \nonumber \\
& \qquad = 2 \vD \odot ( \vD \odot \vV + (\vD \odot \vV )^T ) .
\end{align}

\subsection{Symmetric Cases}
\label{apd:metric_trivi_sym}

When $\lparam$ is symmetric, $\vV$ is also symmetric so
\begin{align}
& \vF(\lparam_0) \vv \nonumber \\ 
&= \nabla_{  \tilde{\slparam} }  \mathrm{Tr}\big\{ \vV^T ( \vD \odot [-2 \MExp( - \vD \odot \lparam) + \MExp( - \vD \odot \lparam)(\vD \odot \lparam )^T + (\vD \odot \lparam )^T  \MExp( - \vD \odot \lparam)] +O(\lparam^2))     \big\} \big|_{\slparam=\slparam_0} \nonumber \\
&= \mathrm{vech} (  2 \vD \odot ( \vD \odot \vV +  \vD \odot \vV  ) ) = 4 \mathrm{vech} (  \vD^2 \odot \vV ).
\end{align}
When $\vD=\half \mathbf{1}$, we have $4 \mathrm{vech} (  \vD^2 \odot \vV )=\mathrm{vech} (\vV) =\vv $, where $\mathbf{1}$ is a matrix of ones.
Thus, $\vF(\lparam_0)=\vI$ .

\subsection{Triangular Cases}
Without loss of generality, we assume that $\lparam$ is lower-triangular, in which case $\vV$ is lower-triangular, and thus
\begin{align}
& \vF(\lparam_0) \vv \nonumber  \\ 
&= \nabla_{   \tilde{\slparam}  }  \mathrm{Tr}\big\{ \vV^T ( \vD \odot [-2 \MExp( - \vD \odot \lparam) + \MExp( - \vD \odot \lparam)(\vD \odot \lparam )^T + (\vD \odot \lparam )^T  \MExp( - \vD \odot \lparam)] +O(\lparam^2))     \big\} \big|_{\slparam=\slparam_0} \nonumber \\
&= \mathrm{tril}( 2 \vD \odot ( \vD \odot \vV +  \vD \odot \vV^T  )\,\,\,\,  (\vV^T \text{ is upper-triangular. Thus } \mathrm{tril}(\vV^T) = \mathrm{Diag}(\vV^T) =  \mathrm{Diag}(\vV) ) \nonumber \\
&=  \mathrm{tril}(  2 \vD^2 \odot (  \vV + \mathrm{Diag}(\vV)  ) ),
\end{align} where  $\mathrm{tril}(\cdot)$  represents a vector representation of the learnable part of a lower-triangular matrix and $\mathrm{Tril}(\cdot)$ denotes a lower-triangular matrix.

When $\vD= \frac{1}{\sqrt{2} } \mathrm{Tril}(\mathbf{1}) + (\half - \frac{1}{\sqrt{2} }) \vI   $, we have $\mathrm{tril}(  2 \vD^2 \odot (  \vV + \mathrm{Diag}(\vV)  ) )  = \mathrm{tril}(\vV) =\vv $, where $ \mathrm{Tril}(\mathbf{1}) $ is a  lower-triangular matrix of ones.
Thus, $\vF(\lparam_0)=\vI$.

\section{An Accurate Approximation of the Euclidean Transport Map}
\label{apd:1st_approx_trans_eg}
We consider the first-order approximation of the Euclidean transport
\begin{align}  
\vm^{(\slparam_1)} =  {\transport}_{\slparam_0\rightarrow \slparam_1}(\vm^{(\slparam_0)} )  = \vomega(1) \approx \underbrace{ \vomega(0) }_{\vm^{(\slparam_0)} }+ \dot{\vomega}(0),
\end{align} where we have to evaluate the Christoffel symbols as discussed below.

By the transport ODE in Eq.~\eqref{eq:parallel_trans_eucl}, we can compute $\dot{\vomega}(0)$ via
\begin{align}
    \dot{\omega}_{c}(0) - \sum_{a,b} \Gamma_{\, cb}^{a}( \vr(0)  )  {\omega}_{a}(0) \dot{r}^{b}(0) =0,
\end{align}  where 
$\vr(0)=\lparam_0$ is the current point and $\dot{\vr}(0)$ is the Riemannian gradient so that $\lparam_1 =\RExp(\lparam_0, \dot{\vr}(0))$.
In our case, as shown in Eq.~\eqref{eq:local_rgd_mom}, we have that $\dot{\vr}(0)= - \vF^{-1}(\slparam_0)\vm^{(\slparam_0)} = \vm^{(\slparam_0)}  $ and $\vomega(0)=\vm^{(\slparam_0)}$.

Note that the metric and Christoffel symbols are evaluated at  $\lparam_0=\mathbf{0}$. The computation can be simplified due to the orthonormal metric as
$\vF^{-1}(\lparam_0)=\vI$ and \begin{equation} \Gamma_{\, cb}^{a}( \lparam_0 ) = \sum_d F^{ad}(\lparam_0) \Gamma_{d,\, cb}( \lparam_0 ) = \sum_d \delta^{ad}  \half [\partial_c F_{bd}(\lparam_0) + \partial_b F_{cd}(\lparam_0) - \partial_d F_{cb}(\lparam_0) ] ,\end{equation} where $F^{ad}(\lparam_0)=\delta^{ad}$.
Thus, we have the following simplification,
\begin{align}
    \dot{\omega}_{c}(0) & =   - \half \sum_{b,d}     [\partial_c F_{bd}(\lparam_0) + \partial_b F_{cd}(\lparam_0) - \partial_d F_{cb}(\lparam_0) ]   ( \vm^{(\slparam_0)} )^d      (\vm^{(\slparam_0)})^b  \nonumber \\
    &= - \half \sum_{b,d}   \partial_c F_{bd}(\lparam_0) ( \vm^{(\slparam_0)} )^d      (\vm^{(\slparam_0)})^b.
\end{align} 
For notation simplicity, we let $\vm=\vm^{(\slparam_0)} $ and $\vA_0=\vA^{\text{(cur)}}$.

For normal coordinate $\vA= \vA_0 \MExp(\vD \odot \lparam)$, we can obtain the following result. The calculation is similar to the metric calculation in Appx.~\ref{apd:simple_fim}.
 \begin{align}
      \dot{\vomega} (0) =    
     \underbrace{    \vD \odot \left( \LieBracket{\vD \odot \vm   }{ \left(\vD \odot \vm \right)^T }  \right)   }_{O(\stepsize^2) } ,\label{eq:chris_error}
\end{align} where $\LieBracket{\vN}{\vM}:=\vN\vM-\vM\vN$ is the Lie bracket, the $\stepsize$ is the stepsize used in Eq.~\eqref{eq:local_rgd_mom}.

When $\lparam$ is symmetric, we know that $\vm$ is symmetric. Thus, we have $\dot{\vomega} (0)=\mathbf{0}$.

\hfill \\

\section{Structured NGD as a Special Case}

\vspace{2mm}

\subsection{Normal Coordinate for Structured NGD} 
\label{apd:sngd_coordinate}
We can obtain coordinate~\eqref{eq:gnormal_coord2} from coordinate~\eqref{eq:gnormal_coord1}.

In  Eq.~\eqref{eq:gnormal_coord1}, the normal coordinate is defined as  $\vA = \vA_0 \MExp\left(  \begin{bmatrix}  \half  \lparam_L & \frac{1}{\sqrt{2}}\lparam_\mu \\ \mathbf{0} & 0 \end{bmatrix} \right) $, where we use $\vA_0$ to denote $\vA^{\text{(cur)}}$. 
Note that
\begin{align}
 \MExp\left(  \begin{bmatrix}  \half  \lparam_L & \frac{1}{\sqrt{2}}\lparam_\mu \\ \mathbf{0} & 0 \end{bmatrix} \right)   
 = \begin{bmatrix} \MExp(  \half  \lparam_L) & \frac{1}{\sqrt{2}}\lparam_\mu +O(\lparam_L \lparam_\mu ) \\ \mathbf{0} & 1 \end{bmatrix} .
\end{align}

The main point is that $O(\lparam_L \lparam_\mu ) $ vanishes in the metric computation since we evaluate the metric at $\lparam_0=\{\lparam_L , \lparam_\mu\}=\mathbf{0}$. Thus, we can ignore  $O(\lparam_L \lparam_\mu ) $, and recover   Eq.~\eqref{eq:gnormal_coord2}:
\begin{align}
\vA & = \vA_0 \begin{bmatrix}  \MExp( \half  \lparam_L) & \frac{1}{\sqrt{2}}\lparam_\mu   \\ \mathbf{0} & 1 \end{bmatrix} = \begin{bmatrix}  \vL_0 &  \vmu_0 \\ \mathbf{0} & 1 \end{bmatrix}  \begin{bmatrix}   \MExp( \half \lparam_L) & \frac{1}{\sqrt{2}}\lparam_\mu  \\ \mathbf{0} & 1 \end{bmatrix} = \begin{bmatrix}  \vL_0 \MExp( \half \lparam_L)  &  \vmu_0 + \frac{1}{\sqrt{2}}  \vL_0 \lparam_\mu   \\ \mathbf{0} & 1 \end{bmatrix}  .
\end{align}

To show that $O(\lparam_L \lparam_\mu ) $ vanishes in the metric computation, we have to show that all the cross terms between $\lparam_L$ and $\lparam_\mu$ of the metric vanish. Using Eq.~\eqref{eq:metric_simple},
\begin{align} 
& \Unmyexpect{\gauss(\mathbf{0}, \gparam)}\big[\nabla_{ \lparam_{L_{jk}} } \nabla_{ \lparam_{\mu_i}  } \log \gauss(\mathbf{0}, \gparam) \big] \big|_{\slparam=\slparam_0}  \nonumber \\
&  \qquad \ \  = \nabla_{ \lparam_{L_{jk}} } \nabla_{ \lparam_{\mu_i}  } \left\{ \mathrm{Tr} \left[   \MExp^T\left( - \begin{bmatrix}  \half  \lparam_L & \frac{1}{\sqrt{2}}\lparam_\mu \\ \mathbf{0} & 0 \end{bmatrix} \right) \MExp\left( - \begin{bmatrix}  \half  \lparam_L & \frac{1}{\sqrt{2}}\lparam_\mu \\ \mathbf{0} & 0 \end{bmatrix} \right)   \right]  \right\}    \big|_{\slparam=\slparam_0}.
\end{align}

\newpage
We can drop higher order terms since we evaluate at $\lparam_0=\{\lparam_L , \lparam_\mu\}=\mathbf{0}$. We get
\hspace*{-0.2cm}
\begin{align}
\addtolength{\arraycolsep}{-2pt}
& \nabla_{ \lparam_{L_{jk}} }  \nabla_{ \lparam_{\mu_i}  } \left\{ \mathrm{Tr} \left[   \MExp^T
\left(\! -\!
\begin{varmatrix}[delim=b,size=\normalsize,sep=3pt]
\half  \lparam_L & \frac{1}{\sqrt{2}}\lparam_\mu \\ \mathbf{0} & 0 
\end{varmatrix}
\! \right) \MExp\left( \!-\! 
\begin{varmatrix}[delim=b,size=\normalsize,sep=3pt]
\half  \lparam_L & \frac{1}{\sqrt{2}}\lparam_\mu \\ \mathbf{0} & 0 \end{varmatrix} 
\! \right)   \! \right] \! \right\} \nonumber  \\
&\!=\!  2 \nabla_{ \lparam_{L_{jk}} }  \left\{ \mathrm{Tr} \left[ \left\{ \nabla_{ \lparam_{\mu_i}  }   \MExp^T\left(\! -\! 
\begin{varmatrix}[delim=b,size=\normalsize,sep=3pt]
\half  \lparam_L & \frac{1}{\sqrt{2}}\lparam_\mu \\ \mathbf{0} & 0 \end{varmatrix} \!
\right) \! \right\} \MExp\left( \!-\! 
\begin{varmatrix}[delim=b,size=\normalsize,sep=3pt]
\half  \lparam_L & \frac{1}{\sqrt{2}}\lparam_\mu \\ \mathbf{0} & 0 \end{varmatrix} 
\! \right)  \! \right] \! \right\}  \nonumber \\
&\!=\! 2 \nabla_{ \lparam_{L_{jk}} } 
 \mathrm{Tr} \left[ \! \left(\! -\!
 \begin{varmatrix}[delim=b,size=\normalsize,sep=3pt]
 \mathbf{0} & \frac{1}{\sqrt{2}} \ve_i \\ \mathbf{0} & 0 
 \end{varmatrix}
 +  \! \half \! 
 \begin{varmatrix}[delim=b,size=\normalsize,sep=3pt]
 \mathbf{0} & \frac{1}{\sqrt{2}} \ve_i \\ \mathbf{0} & 0 
 \end{varmatrix} 
\begin{varmatrix}[delim=b,size=\normalsize,sep=3pt]
 \half  \lparam_L & \frac{1}{\sqrt{2}}\lparam_\mu \\ \mathbf{0} & 0 
 \end{varmatrix}
\! +\! \half \! 
\begin{varmatrix}[delim=b,size=\normalsize,sep=3pt]
\half  \lparam_L & \frac{1}{\sqrt{2}}\lparam_\mu \\ \mathbf{0} & 0 
\end{varmatrix} 
\begin{varmatrix}[delim=b,size=\normalsize,sep=3pt] 
 \mathbf{0} & \frac{1}{\sqrt{2}} \ve_i \\ \mathbf{0} & 0 
 \end{varmatrix}
 \! 
 \right)^T \MExp\left(\! -\! 
\begin{varmatrix}[delim=b,size=\normalsize,sep=3pt]
 \half  \lparam_L & \frac{1}{\sqrt{2}}\lparam_\mu \\ \mathbf{0} & 0 \end{varmatrix} 
 \!\right) \!   \right] \nonumber \\
&\!=\!  2 \nabla_{ \lparam_{L_{jk}} } 
 \mathrm{Tr} \left[ \!  \left(\! -\!
 \begin{varmatrix}[delim=b,size=\normalsize,sep=3pt]
 \mathbf{0} & \frac{1}{\sqrt{2}} \ve_i \\ \mathbf{0} & 0 
 \end{varmatrix}
 \!+  \! \half  \!
 \begin{varmatrix}[delim=b,size=\normalsize,sep=3pt]
 \half  \lparam_L & \frac{1}{\sqrt{2}}\lparam_\mu \\ \mathbf{0} & 0 \end{varmatrix}
\begin{varmatrix}[delim=b,size=\normalsize,sep=3pt]
 \mathbf{0} & \frac{1}{\sqrt{2}} \ve_i \\ \mathbf{0} & 0 
 \end{varmatrix} 
 \!  \right)^T \MExp\left(\! -\! 
 \begin{varmatrix}[delim=b,size=\normalsize,sep=3pt]
 \half  \lparam_L & \frac{1}{\sqrt{2}}\lparam_\mu \\ \mathbf{0} & 0 \end{varmatrix} 
 \!\right) \!  \right] \nonumber \\
 &=  2  \mathrm{Tr} \left[\!
\begin{varmatrix}[delim=b,size=\normalsize,sep=3pt]
   \mathbf{0} & \frac{1}{\sqrt{2}} \ve_i \\ \mathbf{0} & 0 
   \end{varmatrix}^T   
\begin{varmatrix}[delim=b,size=\normalsize,sep=3pt]
   \frac{1}{2} \vE_{jk}  & \  \mathbf{0} \\ \mathbf{0} & 0 
\end{varmatrix}
   \! + \! \half \! \left(\! 
   \begin{varmatrix}[delim=b,size=\normalsize,sep=3pt]
   \half  \vE_{jk} & \mathbf{0} \\ \mathbf{0} & 0 
   \end{varmatrix} 
   \begin{varmatrix}[delim=b,size=\normalsize,sep=3pt]
   \mathbf{0} & \frac{1}{\sqrt{2}} \ve_i \\ \mathbf{0} & 0 
   \end{varmatrix} 
   \!\right)^T  
 \right] =0 ,
\end{align}
and therefore
\begin{equation} \Unmyexpect{\gauss(\mathbf{0}, \gparam)}\big[\nabla_{ \lparam_{L_{jk}} } \nabla_{ \lparam_{\mu_i}  } \log \gauss(\mathbf{0}, \gparam) \big] \big|_{\slparam=\slparam_0}  = 0 .\end{equation}

\hfill 

 \subsection{Gaussian Identities in Structured NGD} 
\label{apd:ngd_gaussian_identites}
 Recall that the manifold is defined as
\begin{align*}
 \mathcal{M}=\left\{ \gparam = \begin{bmatrix} \vV & \vmu \\ \vmu^T & 1 \end{bmatrix}  \in \real^{(d+1) \times (d+1)} \mid \gparam \succ 0 \right\} .
\end{align*}
To use Gaussian gradient identities, we first change the notation from $\vmu $ to $\vm$ to avoid confusion: 
\begin{align*}
 \mathcal{M}=\left\{ \gparam = \begin{bmatrix} \vV & \vm \\ \vm^T & 1 \end{bmatrix}  \in \real^{(d+1) \times (d+1)} \mid \gparam \succ 0 \right\} .
\end{align*}

In Sec.~\ref{sec:sngd_special}, we can compute the Euclidean gradient
\begin{align}
    \vg(\lparam_0) =  \{   \vL^T \vg_1  \vL, \, \sqrt{2} \vL^T (\vg_1 \vm + \vg_2 ) \}, 
\end{align} where $ \vg(\gparam  ^{\text{(cur)}})=\begin{bmatrix}    \vg_1  & \vg_{2}  \\ \vg^T_{2} & 0 \end{bmatrix}$ is a (symmetric) Euclidean gradient w.r.t. $\gparam \in \mathcal{M}$.

By the chain rule, we have
\begin{align}
\frac{\partial  \ell}{\partial m_i}= \mathrm{Tr} \Bigg( \underbrace{  \left(\frac{\partial  \ell}{\partial \gparam}\right)^T }_{ \vg^T(\gparam) } \frac{\partial \gparam }{\partial m_i} \Bigg) = \mathrm{Tr} \left( \begin{bmatrix}    \vg_1  & \vg_{2}  \\ \vg^T_{2} & 0 \end{bmatrix}  \begin{bmatrix} \mathbf{0} & \ve_i \\ \ve_i^T & 1 \end{bmatrix} \right) = 2 \vg_2^T \ve_i ,
\end{align}
so  $\vg_m = \frac{\partial  \ell}{\partial m} =2\vg_2$. Similarly, we have $\vg_V =  \frac{\partial  \ell}{\partial V}  = \vg_1$.

Note that in Gaussian cases, we have
$\vmu=\vm$ and $\vSigma =\vV-\vm\vm^T$, and thus we have
\begin{align}
\nabla_{\mu_i} \ell & =  \mathrm{Tr} \left( \left(\frac{\partial  \ell}{\partial m} \right)^T \frac{\partial  \vm}{\partial \mu_i} \right) +  \mathrm{Tr} \left( \left(\frac{\partial  \ell}{\partial V} \right)^T \frac{\partial  \vV}{\partial \mu_i} \right) =\vg_m^T \ve_i + \mathrm{Tr} \left(  \vg_V^T (\ve_i \vm^T+\vm \ve_i^T) \right) ,  \\
\nabla_{\Sigma_{jk}} \ell &=  \mathrm{Tr} \left( \left(\frac{\partial  \ell}{\partial m} \right)^T \frac{\partial  \vm}{\partial \Sigma_{jk}} \right) +  \mathrm{Tr} \left( \left(\frac{\partial  \ell}{\partial V} \right)^T \frac{\partial  \vV}{\partial \Sigma_{jk}} \right) =0 + \mathrm{Tr} \left(  \vg_V^T  \vE_{jk} \right),
\end{align} which implies that
\begin{align}
\vg_\mu &= \vg_m + (\vg_V+\vg_V^T) \vm = \vg_m + 2 \vg_V \vm = 2\vg_2 + 2\vg_1 \vmu ,\nonumber \\
\vg_\Sigma &= \vg_V = \vg_1 \label{eq:gauss_grad_eq_matrix} .
\end{align}
Thus, $\vg_1$ and $\vg_2$ can be reexpressed using Gaussian gradients $\vg_\mu$ and $\vg_\Sigma$ as
$\vg_1=\vg_\Sigma$ and $\vg_2=\half (\vg_\mu - 2  \vg_\Sigma\,  \vmu)$.

\section{SPD Manifolds}

\subsection{Generalized Normal Coordinates}
\label{apd:gnormal_spd}

We first show that the local coordinate
$\gparam=\vphi_{\sgparam^{\text{(cur)}}}(\lparam)=\vA\vA^T$ is a generalized normal coordinate defined at the reference point $\gparam^{\text{(cur)}}=  \vA^{\text{(cur)}} \big(\vA^{\text{(cur)}}\big)^T$, where $\lparam \in \real^{k \times k}$ is symmetric, and  $\vA=\vA^{\text{(cur)}}\MExp(\half \lparam)$.

It is easy to verify that Assumption 1 holds since $\vphi_{\sgparam^{\text{(cur)}}}(\lparam_0)=\gparam^{\text{(cur)}}$ at $\lparam_0=\mathbf{0}$.

As shown in Appx.~\ref{apd:metric_trivi_sym}, the metric is orthonormal at $\lparam_0=\mathbf{0}$, so Assumption 2 holds.

Recall that the standard normal coordinate is $\gparam= \vpsi_{\sgparam^{\text{(cur)}}}(\lparam)   = \big(\gparam^{\text{(cur)}}\big)^{1/2} \MExp( \lparam   ) \big(\gparam^{\text{(cur)}}\big)^{1/2}  $, where Assumption 3 holds.
Our generalized normal coordinate is defined as $\gparam= \vpsi_{\sgparam^{\text{(cur)}}}(\lparam)   =  \vA^{\text{(cur)}} \MExp( \lparam   ) \big( \vA^{\text{(cur)}} \big)^T $, where $\lparam$ is symmetric.
The only difference between these two coordinates is a multiplicative constant. 
Differentiability and smoothness remain the same.
The injectivity for symmetric $\lparam$ is due to the uniqueness of the symmetric square root of a matrix. 
Thus, Assumption 3 holds in our coordinate.
This statement can be extended to the case where $\lparam$ is a triangular matrix due to the uniqueness of the Cholesky decomposition.

The space of symmetric matrices $\lparam \in \real^{k \times k}$ is an abstract vector space since scalar products and  matrix additions of symmetric matrices are also symmetric. As a result, Assumption 4 holds. 

\subsection{Euclidean Gradients in Normal Coordinates}
As mentioned in Sec.~\ref{sec:demystifying}, there are many generalized  normal coordinates such as  
\begin{itemize} 
\item $\vphi_{\sgparam^{\text{(cur)}}}(\lparam) = \vA \vA^T$, where  $\lparam$ is symmetric and  $\vA:=\vA^{\text{(cur)}} \MExp(\half \lparam)$  ,
\item
    $ \vphi_{\sgparam^{\text{(cur)}}}(\lparam)= \vB^{-T}\vB^{-1}$,  where $\lparam$ is symmetric and 
$\vB:=\vB^{\text{(cur)}}\MExp(-\half \lparam) $,  
\item  
    $ \vphi_{\sgparam^{\text{(cur)}}}(\lparam)=\vC^T \vC$, where $\lparam$ is symmetric and  $\vC:=\MExp(\half \lparam)\vC^{\text{(cur)}} $. 
\end{itemize}  

We show how to compute the Euclidean gradient $\vg(\lparam_0) $ needed in Eq.~\eqref{eq:local_rgd_mom}, where we assume that the Euclidean gradient $\nabla_{\sgparam} \ell = \vg(\gparam)$ w.r.t. $\gparam$ is given. 
Let us consider $ \gparam= \vphi_{\sgparam^{\text{(cur)}}}(\lparam) = \vA \vA^T$. By the chain rule, we have
\begin{align}
\nabla_{\slparam_{ij}} \ell = \mathrm{Tr} \big( \vg^T(\gparam)    \nabla_{ \slparam_{ij} } \gparam \big) \big|_{\lparam=\lparam_0} =
 \mathrm{Tr} \big(  \vg^T(\gparam) \vA^{\text{(cur)}} \vE_{ij} (\vA^{\text{(cur)}})^T ) ,
\end{align}
so
\begin{align}
\vg(\lparam_0)= (\vA^{\text{(cur)}})^T  \vg(\gparam) \vA^{\text{(cur)}} 
\label{eq:egd_coordinate_a} .
\end{align}

Similarly, when $ \gparam= \vB^{-T}\vB^{-1}$, we have 
\begin{align}
\nabla_{\slparam_{ij}} \ell = \mathrm{Tr} \big( \vg^T(\gparam)    \nabla_{ \slparam_{ij} } \gparam \big) \big|_{\lparam=\lparam_0} =
 \mathrm{Tr} \big(  \vg^T(\gparam) (\vB^{\text{(cur)}})^{-T} \vE_{ij} (\vB^{\text{(cur)}})^{-1} ) ,
\end{align} which gives 
\begin{align}
\vg(\lparam_0)=   (\vB^{\text{(cur)}})^{-1}   \vg(\gparam)(\vB^{\text{(cur)}})^{-T}. \label{eq:egd_coordinate_b}
\end{align} 

\hfill 

\subsection{Simplification of Our Update}
\label{apd:simple_update_spd}

Consider the normal coordinate $\vphi_{\sgparam^{\text{(cur)}}}(\lparam) = \vA \vA^T$, where  $\lparam$ is symmetric and  $\vA:=\vA^{\text{(cur)}} \MExp(\half \lparam)$.

We can compute the Euclidean gradient as $\vg(\lparam_0)=(\vA^{\text{(cur)}})^T  \vg(\gparam) \vA^{\text{(cur)}} $.

Using the approximation in Eq.~\eqref{eq:transport_approx}, we have 
\begin{align}
\vw^{ (\slparam_1) } \leftarrow  \vm^{ (\slparam_0) } .
\end{align}
Since $\lparam$ is symmetric, we can further show that the accurate approximation also gives the same update since the second dominant term vanishes as shown in Eq.~\eqref{eq:chris_error}.

By Eq.~\eqref{eq:jacob_vec_sym},  the vector-Jacobian product needed in Eq.~\eqref{eq:jacobian_transform} can be expressed as 
\begin{align}
\vw^{ (\xi_0) } = \vJ(\vxi_0)  \vw^{ (\slparam_1) } =\vw^{ (\slparam_1) }.
\end{align}

Thus, we have $\vw^{ (\xi_0) } = \vm^{ (\slparam_0) } $.
As a consequence, our update (defined in Eq.~\eqref{eq:local_rgd_mom}) can be simplified as below, where we can drop all the superscripts and let $\vw=\vm$,
\begin{align} 
\text{Momentum}:&\,
\vm \leftarrow \alpha \vm  + \stepsize \overbrace{ (\vA^{\text{(cur)}})^T  \vg(\gparam) \vA^{\text{(cur)}}}^{ = \vg(\lparam_0) }, \nonumber \\
\text{GD}:&\, \lparam_1   \leftarrow  -\vm  , \nonumber \\
 &\gparam^{\text{(new)}}  \leftarrow  \vphi_{\sgparam^{\text{(cur)}}}(\lparam_1)= \vA^{\text{(cur)}} \MExp( \lparam_1 ) \big( \vA^{\text{(cur)}} \big)^T  \Longleftrightarrow \vA^{\text{(new)}}  \leftarrow \vA^{\text{(cur)}}  \MExp(   \half \lparam_1 ) .\label{eq:update_coordinate_a} 
\end{align} 

Using Eq.~\eqref{eq:egd_coordinate_b}, 
we can also obtain the following update if the normal coordinate $\vphi_{\sgparam^{\text{(cur)}}}(\lparam) =  \vB^{-T}\vB^{-1}$ is used, where  $\lparam$ is symmetric and   $\vB:=\vB^{\text{(cur)}}\MExp(-\half \lparam)$,
\begin{align} 
\text{Momentum}:&\,
\vm \leftarrow \alpha \vm  +\stepsize \overbrace{    (\vB^{\text{(cur)}})^{-1}   \vg(\gparam)(\vB^{\text{(cur)}})^{-T} }^{ =   \vg(\lparam_0) }, \nonumber \\
\text{GD}:&\, \lparam_1   \leftarrow  -\vm  , \nonumber \\
 &\gparam^{\text{(new)}}  \leftarrow  \vphi_{\sgparam^{\text{(cur)}}}(\lparam_1)=\big( \vB^{\text{(cur)}} \big)^{-T} \MExp(- \lparam_1 ) \big( \vB^{\text{(cur)}} \big)^{-1} \Longleftrightarrow \vB^{\text{(new)}}  \leftarrow \vB^{\text{(cur)}}  \MExp( - \half \lparam_1 ). \label{eq:update_coordinate_b}
\end{align} 

\vspace{3mm}

\section{SPD Kronecker-product Submanifolds}
\label{apd:mat_gauss_dl}
We consider the SPD submanifold
\begin{align*}
 \mathcal{M}\!=\!\Big\{ \gparam\!=\!\vA \vA^T \in \mathcal{S}_{++}^{(pd) \times (pd)}  \mid \vA\!:=\!  \vK \otimes \vC,   \vK \!\in \! \real^{p \times p}, \vC \!\in\! \real^{d \times d} \Big\} ,
 \end{align*} where $\vU=\vK\vK^T \succ 0$,  $\vW=\vC\vC^T \succ 0$, and both  $\vK$ and $\vC$ are dense and non-singular. 

\subsection{Blockwise Normal Coordinates}
As mentioned in Sec.~\ref{sec:opt_dl}, we consider a block-diagonal approximation of the affine-invariant metric. For block $\vK$, we consider the coordinate
\begin{align}
\vA & =  
\big(\vK^{\text{(cur)}} \MExp(  \frac{1}{2 \sqrt{d}} \lparam_K) \big) \otimes   \vC^{\text{(cur)}}  ,
 \end{align} where   $\lparam_K \in \real^{p \times p }$  is symmetric and   $\vA^{\text{(cur)}}=\vK^{\text{(cur)}} \otimes\vC^{\text{(cur)}} $.  
 
 We will show that  the block-diagonal approximated metric is orthonormal at $\lparam_K   =\mathbf{0}$  under  coordinate   $\lparam_K$.

For notation simplicity, we let $\vK_0 = \vK^{\text{(cur)}} $,   $\vC_0 =\vC^{\text{(cur)}} $, and $\gparam_0 =\gparam^{\text{(cur)}}  $. Let $\tilde{\lparam}_K$ denote the learnable part of $\lparam_K$.

By  the Kronecker-product, we have
$ \mathrm{vec}^T(\vX) (\vU  \otimes  \vW) \mathrm{vec}(\vX) = 
 \mathrm{vec}^T(\vX) \mathrm{vec}(\vW \vX \vU^T )  = \mathrm{Tr}( \vX^T \vW \vX \vU^T)
$, where $\vx:= \mathrm{vec}(\vX)$ and $\vX \in \real^{d \times p}$.

By definition, the metric  $\vF$ w.r.t. block $\vK$ in coordinate $\lparam_K$ is 
\begin{align}
&\vF_K(\mathbf{0}) \!=\! -2 \Unmyexpect{\gauss(\mathbf{0}, \gparam)}\big[ \nabla_{ \tilde{\slparam}_K }^2 \log \gauss(\mathbf{0}, \gparam) \big] \big|_{\slparam_K=0} \nonumber \\
&\!=\! \Unmyexpect{\gauss(x|\mathbf{0}, \gparam)}\big[ \nabla_{ \tilde{\slparam}_K }^2 \big\{ \mathrm{Tr} [ \vx^T\left(   
\big( \vK_0^{-T} \MExp(  -\frac{1}{  \sqrt{d}} \lparam_K) \vK_0^{-1}  \big) \otimes  \big(  \vC_0^{-T}  \vC_0^{-1}  \big)\right) \vx
]     \big] \big|_{\slparam_K=0} \,\,\, (\text{drop linear terms in the log-det term})\nonumber \\
&\!=\! \Unmyexpect{\gauss(x|\mathbf{0}, \gparam)}\big[ \nabla_{ \tilde{\slparam}_K }^2 \big\{ \mathrm{Tr} [ \vX^T  \big(  \vC_0^{-T}  \vC_0^{-1}  \big)   \vX
\big( \vK_0^{-T} \MExp(  -\frac{1}{  \sqrt{d}} \lparam_K) \vK_0^{-1}  \big)  
]   \big\}  \big] \big|_{\slparam_K=0}\, (\text{note: }  \vx^T (\vU  \otimes  \vW) \vx= \mathrm{Tr}( \vX^T \vW \vX \vU^T) ) \nonumber \\
&\!=\! d \Unmyexpect{\gauss(x|\mathbf{0}, \gparam)}\big[ \nabla_{ \tilde{\slparam}_K }^2 \big\{ \mathrm{Tr}[  \MExp(  -\frac{1}{  \sqrt{d}} \lparam_K) \big\}] \big] \big|_{\slparam_K=0} \,\, (\text{note: } \Unmyexpect{\gauss(x|\mathbf{0}, \gparam_0)}[ \vX^T  \big(  \vC_0^{-T}  \vC_0^{-1}  \big)   \vX] = d\vK_0\vK_0^T) .
\end{align}

It is easy to show that $\vF_K(\mathbf{0})=\vI$ w.r.t. block $\vK$ in coordinate $\lparam_K$, which means Assumption 2 holds.

Since block $\vC$ is frozen, we can prove as in Appx.~\ref{apd:gnormal_spd} that all assumptions are satisfied for the coordinate $\lparam_K$. 

Similarly, for block $\vC$, we can consider the coordinate
\begin{align}
\vA & =  
\vK^{\text{(cur)}}   \otimes \big(  \vC^{\text{(cur)}}  \MExp(  \frac{1}{2 \sqrt{p}} \lparam_C) \big) ,
 \end{align}   where   $\lparam_C \in \real^{d \times d }$  is symmetric and   $\vA^{\text{(cur)}}=\vK^{\text{(cur)}} \otimes\vC^{\text{(cur)}} $, and show that it defines a normal coordinate. 

\hfill 

\subsection{ (Euclidean) Gradient Computation for Deep Learning  }
We consider $\gparam=\vSigma = (\vK\vK^T ) \otimes (\vC\vC^T )$.

As suggested by \citet{lin2021snd}, the Euclidean gradient w.r.t. $\gparam$ is computed as  $\vg_\Sigma:=\half (\nabla_\mu^2 \ell(\vmu)-\vSigma^{-1}) $.

In KFAC \citep{martens2015optimizing}, the Hessian is approximated as 
$\nabla_\mu^2  \ell(\vmu)\approx \vmu_{AA} \otimes \vmu_{GG}   $, where matrices $\vmu_{AA} \in \real^{p \times p}$ and $\vmu_{GG}  \in \real^{d \times d}$ are two dense symmetric positive semi-definite matrices and are computed as suggested by the authors.

To handle the singularity of $\vmu_{AA}$ and $\vmu_{GG}$, \citet{martens2015optimizing} introduce a damping term $\lambda$  when it comes to inverting   $\vmu_{AA}$ and  $\vmu_{GG}$ such as $ \vmu^{-1}_{AA} \approx (\vmu_{AA}+\lambda \vI_p)^{-1} $  and 
$ \vmu^{-1}_{GG} \approx (\vmu_{GG}+\lambda \vI_d)^{-1} $.

In our update, we use the KFAC approach to approximate  the Hessian. We add a damping term by including it in $\vg_\Sigma$   as  \begin{align} \vg_\Sigma   \approx \half ( \underbrace{ \vmu_{AA} \otimes \vmu_{GG}}_{ \approx \nabla_\mu^2 \ell(\vmu) } + \lambda \vI_{pd}    -\vSigma^{-1})  ,\end{align} where $\vI_{pd} = \vI_p \otimes \vI_d$ and $\vSigma^{-1} = ( \vK^{-T} \vK^{-1} ) \otimes (\vC^{-T} \vC^{-1} ) $.

The Euclidean gradient $\vg(\lparam_{K_0})$ w.r.t. $\lparam_E$ can be computed as 
\begin{align}
\frac{\partial \ell}{\partial \slparam_{K_{ij} } } = \mathrm{Tr}\big ( \big(\frac{\partial \ell}{\partial \gparam }\big)^T  \frac{\partial \gparam 
 }{\partial \slparam_{K_{ij} }}   \big)= \mathrm{Tr} \big( [\vg_\Sigma]^T \frac{\partial \gparam 
 }{\partial \slparam_{K_{ij} }} \big) .
\end{align}

There are three terms in the Euclidean gradient w.r.t. $\gparam=\vSigma$: \begin{align} \vg_\Sigma \approx \half ( \vmu_{AA} \otimes \vmu_{GG} + \lambda \vI_p \otimes \vI_d - ( \vK^{-T} \vK^{-1} ) \otimes (\vC^{-T} \vC^{-1} ) . \end{align}

Thus, the Euclidean gradient w.r.t. $\lparam$ can be decomposed into three parts. For notation simplicity, we let $\vK_0 = \vK^{\text{(cur)}} $,   $\vC_0 =\vC^{\text{(cur)}} $, and $\gparam_0 =\gparam^{\text{(cur)}}  $. 

The first part of $\frac{\partial \ell}{\partial \slparam_{K_{ij} } }$ can be computed via
\begin{align}
  \half   \mathrm{Tr} \big( [ \vmu_{AA} \otimes \vmu_{GG} ]^T \frac{\partial \gparam 
 }{\partial \slparam_{K_{ij} }} \big) & =  \frac{1}{2 \sqrt{d} } \mathrm{Tr} \big[ ( \vmu_{AA}^T \vK_0 \vE_{ij} \vK_0^T ) \otimes (  \vmu_{GG}^T \vC_0 \vC_0^T ) \big] \nonumber \\ & =  \frac{1}{2 \sqrt{d} }\mathrm{Tr}(  \vmu_{GG}^T \vC_0 \vC_0^T ) \mathrm{Tr}   ( \vmu_{AA}^T \vK_0 \vE_{ij} \vK_0^T ).
\end{align} 
We obtain the expression for the first part of $\frac{\partial \ell}{\partial \slparam_{K} }$  as  $\frac{1}{2 \sqrt{d} }\mathrm{Tr}( \vC_0^T \vmu_{GG}  \vC_0  ) \vK_0^T \vmu_{AA} \vK_0 $.

Similarly, we can obtain the second and third parts, which gives altogether the Euclidean gradient $\vg(\lparam_K)$ via
\begin{align}
\vg(\lparam_{K_0}) = \frac{1}{2 \sqrt{d} } [  \mathrm{Tr}( \vC_0^T \vmu_{GG}  \vC_0  ) \vK_0^T \vmu_{AA} \vK_0 +\lambda  \mathrm{Tr}(    \vC_0^T \vC_0 ) \vK_0^T \vK_0 - d\vI_p] .
\end{align}

Likewise, the Euclidean gradient $\vg(\lparam_C)$ is
\begin{align}
\vg(\lparam_{C_0}) = \frac{1}{2 \sqrt{p} }[  \mathrm{Tr}( \vK_0^T \vmu_{AA}  \vK_0  ) \vC_0^T \vmu_{GG} \vC_0 +\lambda  \mathrm{Tr}(    \vK_0^T \vK_0 ) \vC_0^T \vC_0 - p\vI_d ] .
\end{align}

\subsection{ Derivation of the Update}
We consider the update   for block $\lparam_K$.
By  the approximation in Eq.~\eqref{eq:transport_approx}, we have 
\begin{align}
\vw^{ (\slparam_{K_1}) } \leftarrow  \vm^{ (\slparam_{K_0}) } ,
\end{align} for block $\lparam_K$.
Since $\lparam_K$ is symmetric, we can further show that the accurate approximation also gives the same update since the second dominant term vanishes as shown in Eq.~\eqref{eq:chris_error}.

Since $\lparam_K$ is symmetric (see Eq.~\eqref{eq:jacob_vec_sym}), the vector-Jacobian product needed in Eq.~\eqref{eq:jacobian_transform} can be expressed as 
\begin{align}
\vw^{ (\xi_{K_0}) } = \vJ(\vxi_{K_0})  \vw^{ (\slparam_{K_1}) } =\vw^{ (\slparam_{K_1}) }.
\end{align} 

Thus, we have $\vw^{ (\xi_{K_0}) } = \vm^{ (\slparam_{K_0}) } $ for $\lparam_K$.

As a consequence (similar to Appx.~\ref{apd:simple_update_spd}), our update (defined in Eq.~\eqref{eq:local_rgd_mom}) for block $\lparam_K$ can be expressed as below, where we drop all superscripts and let $\vw=\vm$,
\begin{align} 
\text{Momentum}:&\,
\vm_K \leftarrow \alpha \vm_K  + \stepsize \vg(\lparam_{K_0}), \nonumber \\
\text{GD}:&\, \lparam_{K_1}   \leftarrow  -\vm_K  , \nonumber \\
 & \vK \leftarrow \vK^{\text{(cur)}} \MExp( \frac{1}{2\sqrt{d}   }  \lparam_{K_1} ) . 
\end{align} 

Since we initialize $\vm_K $ to $\mathbf{0}$, we can merge factor $\frac{1}{2\sqrt{d}}$ into $\vm_K $ as shown below.
\begin{align} 
\text{Momentum}:&\,
\vm_K \leftarrow \alpha \vm_K  +  \frac{\stepsize}{2\sqrt{d}   } \vg(\lparam_{K_0}), \nonumber \\
\text{GD}:&\, \lparam_{K_1}   \leftarrow  -\vm_K  , \nonumber \\
 & \vK \leftarrow \vK^{\text{(cur)}} \MExp(   \lparam_{K_1} ) . \label{eq:matgauss_just}
\end{align} 

Note that the affine-invariant metric is defined as twice of the Fisher-Rao metric. 

To recover structured NGD, we have to set our stepsize $\stepsize$ to twice the stepsize of structured NGD.
Letting $\stepsize=2\beta_2$, we can reexpress the above update for block $\lparam_K$ as
\begin{align} 
&\vm_K \leftarrow \alpha \vm_K  +  \frac{\beta_2}{\sqrt{d}} \vg(\lparam_{K_0}), \nonumber \\
 & \vK \leftarrow \vK^{\text{(cur)}} \MExp( - \vm_K  ) . 
\end{align} 

A similar update for the block $\lparam_C$ can also be obtained. 

\hfill

\section{Implementation for the Baseline Methods }
\label{apd:imple_baselines}
We consider the following manifold optimization problem on a SPD full manifold:
\begin{align}
    \min_{\gparam \in \mathcal{S}_{++}^{k \times k} } \ell(\gparam)
\end{align}
Recall that a Riemannian gradient w.r.t. $\gparam$ is $\hat{\vg}(\gparam):=\gparam \left(\nabla_\sgparam \ell \right) \gparam= - \nabla_{\sgparam^{-1}} \ell$.

The Riemannian gradient descent (RGD) is
\begin{align}
    \gparam^{\text{(new)}} \leftarrow \RExp(\gparam^{\text{(cur)}}, -\stepsize \hat{\vg}(\gparam^{\text{(cur)}}) ) .
\end{align}

The update of \citet{alimisis2020continuous} is shown below, where we initialize $\vz$ by $0$:
\begin{align}
\vnu^{\text{(cur)}} & \leftarrow   \alpha \vz^{\text{(cur)}} + \stepsize   \hat{\vg}(\gparam^{\text{(cur)}}) \nonumber \\
 \gparam^{\text{(new)}} & \leftarrow \RExp (\gparam^{\text{(cur)}}, -\vnu^{\text{(cur)}} ) \nonumber \\
 \vz^{\text{(new)}} & \leftarrow \hat{\transport}_{\sgparam^{\text{(cur)}}\rightarrow \sgparam^{\text{(new)}}}(\vnu^{\text{(cur)}}) 
 \end{align}

The update of \citet{alimisis2021momentum} is shown below, where  we initialize $\vy$   by $
\gparam$ :
\begin{align}
\vz^{\text{(new)}} & \leftarrow \RExp(\gparam^{\text{(cur)}}, -\stepsize  \hat{\vg}(\gparam^{\text{(cur)}}) ) \nonumber \\
\vy^{\text{(new)}} & \leftarrow \RExp(\vy^{\text{(cur)}}, -\frac{\stepsize}{1-\alpha} \hat{\transport}_{\sgparam^{\text{(cur)}}\rightarrow y^{\text{(cur)}}  }( \hat{\vg}(\gparam^{\text{(cur)}}) )   ) \nonumber  \\
\gparam^{\text{(new)}} & \leftarrow \RExp (\vy^{\text{(new)}}, \alpha \RExp^{-1}(\vy^{\text{(new)}}, \vz^{\text{(new)}} ) )
\end{align}

The update of \citet{ahn2020nesterov} is shown below, where we initialize  $\vz$  by $
\gparam$:
\begin{align}
    \vy^{\text{(new)}} & \leftarrow \RExp(\gparam^{\text{(cur)}}, -\stepsize  \hat{\vg}(\gparam^{\text{(cur)}}) ) \nonumber \\
\vz^{\text{(new)}} & \leftarrow \RExp(\gparam^{\text{(cur)}}, \frac{\alpha}{1-\alpha}  \RExp^{-1}(\gparam^{\text{(cur)}}, \vz^{\text{(cur)}} ) - 2\stepsize \hat{\vg}(\gparam^{\text{(cur)}})    ) \nonumber \\
  \gparam^{\text{(new)}} & \leftarrow \RExp (\vy^{\text{(new)}}, \alpha \RExp^{-1}(\vy^{\text{(new)}}, \vz^{\text{(new)}} ) )
\end{align}

We properly select momentum weights and stepsizes in \citet{ahn2020nesterov} and  \citet{alimisis2020continuous,alimisis2021momentum} so that these updates are equivalent in Euclidean cases. 

Recall that our update with momentum in the GNC $\gparam=\vC^T \vC$ (see Sec.~\ref{sec:demystifying}) is
 \begin{align} 
\vm^{\text{(new)}} & \leftarrow \alpha \vm^{\text{(cur)}}  +\stepsize   \overbrace{ (\vC^{\text{(cur)}})^{-T}   \hat{\vg}(\gparam^{\text{(cur)}} )(\vC^{\text{(cur)}})^{-1}}^{= \vC^{\text{(cur)}} \left( \nabla_{\sgparam} \ell(\gparam^{\text{(cur)}}) \right) \left(\vC^{\text{(cur)}}\right)^T }    \nonumber \\
 \lparam_1  & \leftarrow  \mathbf{0} - \vm^{\text{(new)}}   \nonumber \\
 \vC^{\text{(new)}}  & \leftarrow \MExp(\half  \lparam_1 )  \vC^{\text{(cur)}}   \nonumber \\
 \gparam^{\text{(new)}} & \leftarrow  \left( \vC^{\text{(new)}} \right)^T   \vC^{\text{(new)}}
 \label{eq:our_update_sgd_full}
\end{align} where $\vC$ is a dense non-singular matrix,  we initialize $\vm$ by $0$, and 
we use the quadratic truncation of the matrix exponential as $\MExp(\vN)\approx \vI + \vN + \half \vN^2$. 
Thus, 
 \begin{equation} \MExp(  \vN )  \vC  \approx 
 \half \left( \vI + (\vI+\vN)(\vI+\vN) \right) \vC . \end{equation}
Note that when $\vN$ is a symmetric matrix, we have $ \vI + \vN + \half \vN^2 = \half \left(\vI + (\vI+\vN)(\vI+\vN)^T \right) \succ 0$. Since $\lparam_1$ is a symmetric matrix, we know that the updated $\gparam$ is SPD even when we use the truncation. The statement about the truncation also holds when $\vN$ is a triangular matrix arising from a new GNC using a Cholesky factor  $\vC$. 
 
We can recover the update of \citet{lin2021tractable} on a SPD manifold
by setting $\alpha=0$ in Eq.~\eqref{eq:our_update_sgd_full}.

For a Gaussian submanifold $\gparam=\begin{bmatrix}
    \vSigma + \vmu \vmu^T & \vmu \\
    \vmu^T & 1
\end{bmatrix}$ considered in Sec.~\ref{sec:sngd_special}, the update of \citet{lin2021tractable} on this submanifold is shown below, where we can use the Gaussian gradient identities in Eq.~\eqref{eq:gauss_grad_eq_matrix} and $\vSigma=\vU^T \vU$:
\begin{align}
    \vmu  & \leftarrow \vmu - \frac{\stepsize}{2} \vSigma \left( \nabla_\mu \ell \right) \nonumber \\
    \vU & \leftarrow \MExp\left(-\frac{\stepsize}{2} \vU\left(  \nabla_\Sigma \ell \right) \vU^T \right) \vU
\end{align} where we use a new GNC and  the  quadratic truncation for the matrix exponential. 
\end{appendices}

\end{document}